\RequirePackage{snapshot}
\documentclass[english,10pt,twocolumn,letterpaper]{article}
\usepackage[T1]{fontenc}
\usepackage[latin9]{inputenc}
\usepackage{color}
\usepackage{array}
\usepackage{float}
\usepackage{multirow}
\usepackage{amsmath}
\usepackage{amssymb}
\usepackage{stmaryrd}
\usepackage{graphicx}
\usepackage[numbers]{natbib}
\usepackage{pbox}
\usepackage{wrapfig}
\usepackage{etoolbox}
\usepackage{fancyhdr}
\usepackage[nomessages]{fp}
\patchcmd{\chapter}{plain}{fancy}{}{}

\definecolor{gray}{rgb}{0.5, 0.5, 0.5}

\newtoggle{arxiv}
\toggletrue{arxiv}

\newtoggle{arxiv_flatten}
\toggletrue{arxiv_flatten}

\usepackage{pifont}
\usepackage{etoc}
\usepackage{pdflscape} 

\usepackage{tocloft}
\usepackage{longtable}

\providecommand{\tabularnewline}{\\}

\usepackage{titlesec}

\iftoggle{arxiv}{
    \newcommand{\optvspace}[1]{}
    \newcommand{\arxvspace}[1]{\vspace{#1}}
}{
    \newcommand{\optvspace}[1]{\vspace*{#1}}
    \newcommand{\arxvspace}[1]{}
}

\iftoggle{arxiv}{\usepackage{cvpr}
\titlespacing*{\paragraph}{0pt}{0.2ex plus 0.05ex}{1.5ex}
}{\usepackage{cvpr-compact}
}

\usepackage{times}
\usepackage{epsfig}
\usepackage{graphicx}
\usepackage{amsmath}
\usepackage{amssymb}
\usepackage{subfig}

\usepackage[dvipsnames]{colortbl}

\usepackage{setspace}

\usepackage{enumitem}

\setlist{noitemsep, nolistsep}

\usepackage[pagebackref=true,breaklinks=true,letterpaper=true,colorlinks,bookmarks=false]{hyperref}

\cvprfinalcopy 
\def\cvprPaperID{932} 

\iftoggle{arxiv}{    \fancypagestyle{firststyle}{
    \fancyfoot[L]{\footnotesize 
      \textsf{This paper is published at CVPR~2018.}
    }
  }
}{  \fancypagestyle{firststyle}{}
\ifcvprfinal\fancyfoot{}\pagestyle{empty}\fi
}

\newcommand{\lowreswarn}{\vspace*{1em} \textsf{\textcolor{gray}{Due to the file size limit of arXiv, images in this file are significantly downsampled. 
The higher-quality PDF with larger file size is available at}}\\
\url{http://ytzhang.net/files/publications/2018-lmdis-rep.pdf} 
}
\renewcommand{\lowreswarn}{}

\iftoggle{arxiv_flatten}{
\fancyfoot[C]{\thepage \\ \footnotesize \lowreswarn}
}
{}

\definecolor{cellhl}{rgb}{0.95, 0.95, 0.96}

\usepackage[font=small,labelfont=bf]{caption}

\newcommand{\filluptopage}[1]{  \clearpage
  \loop\ifnum\value{page}<#1\relax
    \null\clearpage
  \repeat
}

\iftrue  
            \newcommand{\todo}[1]{}
        \newcommand{\outline}[1]{}
        \newcommand{\textgray}[1]{}
        \newcommand{\commenttext}[1]{}
        \newcommand{\commentfoot}[1]{}
        \newcommand{\commentselfoot}[2]{}
        \newcommand{\commentselrep}[2]{}
        \newcommand{\topic}[1]{}
        \newcommand{\commenthl}[1]{}
\else 
            \newcommand{\todo}[1]{{\textcolor{red}{[[TODO: {#1}]]}}}
        \newcommand{\outline}[1]{{\textcolor{blue}{[[{#1}]]}}}
        \newcommand{\textgray}[1]{\textcolor{gray}{[[{#1}]]}}
        \newcommand{\commenttext}[1]{\textcolor{red}{[[{#1}]]}}
        \newcommand{\commentfoot}[1]{\footnote{\textcolor{red}{\textit{#1}}}}
        \newcommand{\commentselfoot}[2]{{\textcolor{blue}{#1}}\commenttext{#2}}
        \newcommand{\commentselrep}[2] {{\textcolor{blue}{#1}} {\textcolor{green}{[[\textit{#2}]]}}}
        \newcommand{\topic}[1]{\textcolor{gray}{\textbf{(#1.)}}}
        \newcommand{\commenthl}[1]{\textcolor{blue}{[HL: #1]}}
               \fi

\definecolor{darkgreen}{rgb}{0.0,0.70,0}
\definecolor{darkyellow}{rgb}{0.9,0.60,0}

\newcommand{\supp}[1]{Appendix~{#1}}

\newcommand{\identitymat}{\mathbb{I}}

\usepackage{caption}

\iftoggle{arxiv}{

}{

}

\usepackage{babel}
\begin{document}

\global\long\def\op#1{\operatorname{#1}}
\global\long\def\mc{\text{\textcolor{blue}{[??]}}}
\global\long\def\mr{\text{\textcolor{blue}{??}}}
\global\long\def\normalized#1{\overline{#1}}

\newcommand{\mytitle}{Unsupervised Discovery of Object Landmarks as Structural Representations}

\newcommand{\myauthor}{Yuting Zhang$^1$, Yijie Guo$^1$, Yixin Jin$^1$, Yijun Luo$^1$, Zhiyuan He$^1$, Honglak Lee$^{2,1}$}
\newcommand{\myemail}{\{yutingzh, guoyijie, jinyixin, lyjtour, zhiyuan, honglak\}@umich.edu \quad honglak@google.com}

\newcommand{\myaffliation}{$^1$University of Michigan, Ann Arbor \quad \quad $^2$Google Brain \vspace{-0.2em}}

\title{\mytitle}

\author{
\myauthor
\\
\myaffliation
\\
{\small{} \texttt{\myemail}}
}

\twocolumn[
  \begin{@twocolumnfalse}
    \maketitle
       \end{@twocolumnfalse}
]

\begin{abstract}
\vspace*{-0.1in}
Deep neural networks can model images with rich latent representations, but they cannot naturally conceptualize structures of object categories in a human-perceptible way. 
This paper addresses the problem of learning object structures in an image modeling process without supervision.
We propose an autoencoding formulation to discover landmarks as explicit structural representations. 
The encoding module outputs landmark coordinates, whose validity is ensured by constraints that reflect the necessary properties for landmarks. 
The decoding module takes the landmarks as a part of the learnable input representations in an end-to-end differentiable framework. 
Our discovered landmarks are semantically meaningful and more predictive of manually annotated landmarks than those discovered by previous methods. 
The coordinates of our landmarks are also complementary features to pretrained deep-neural-network representations in recognizing visual attributes. 
In addition, the proposed method naturally creates an unsupervised, perceptible interface to manipulate object shapes and decode images with controllable structures.
\iftoggle{arxiv}{
The project web page: 
\begingroup
\normalfont{}\small{}
\url{http://ytzhang.net/projects/lmdis-rep} 
\endgroup
}{}

\vspace*{-0.1in}
\optvspace{-0.5em}
\end{abstract}

\etocdepthtag.toc{mtchapter}
\etocsettagdepth{mtchapter}{subsection}
\etocsettagdepth{mtappendix}{none}

\allowdisplaybreaks

\thispagestyle{firststyle}

\section{Introduction}
Computer vision seeks to understand object structures that reflect the physical states of objects and show invariance to individual appearance changes. 
Such intrinsic structures can serve as intermediate representations for high-level visual understanding. 
However, manual annotations or designs of object structures (e.g., skeleton, semantic parts) are costly and barely available for most object categories, making the automatic representation learning of object structure an attractive solution to this challenge. 

Modern neural networks can learn latent representations to effectively solve various vision problems, including image classification~\citep{alexnet,vggnet,googlenet,resnet}, segmentation~\citep{fcnn,deconv-seg,mask-rcnn}, object detection~\citep{rcnn,fgs-struct,faster-rcnn}, human pose estimation~\citep{hourglass}, 3D reconstruction~\citep{deep-depth,3d-interpreter,im2point}, and image generation~\citep{vae,gan,dc-gan}. 
Several existing studies~\citep{rcnn,zfnet,net-dissection} observe that these representations naturally encode massive templates of particular visual patterns.
However, little evidence suggests that deep neural networks can naturally conceptualize the intrinsic structures of an object category compactly and perceptibly. 

We aim at learning the physical parameters of conceptualized object structures without supervision. 
As a typical representation of intrinsic structures, landmarks represent the spatial configuration of stable local semantics across different object instances of the same category.
\citet{unsupervised-landmark} proposed an unsupervised method to locate landmarks at the places where a convolutional neural network can detect stable visual patterns with high spatial equivariance to image transformations. 
However, this method did not explicitly encourage the landmarks to appear at critical locations for image modeling.

This paper addresses the problem of discovering landmarks in a generic image modeling process. 
In particular, we take landmark discovery as an intermediate step for image autoencoding. 
To leverage the training signals from the landmark-based image decoder, gradients need to go through the landmark coordinates, which makes \citet{unsupervised-landmark}'s non-differentiable formulation infeasible. 
With a different way to calculate landmark coordinates, the image decoding module can make the landmark configuration informative regarding image reconstruction.
We also introduce additional regularization terms to enforce the desirable properties of the detected landmarks and to prevent the landmark coordinates from encoding irrelevant or redundant latent information.

Our contributions in this paper are as follows.
\begin{enumerate}
\item We develop a differentiable autoencoder framework for object landmark discovery, which allows the image decoder to propagate training signals back to the landmark detection module. 
We introduce several soft constraints to reflect the properties of landmarks, forcing the discovered representations to be valid landmarks. 
\item The proposed method discovers visually meaningful landmarks without supervision for a variety of objects. 
It outperforms the state-of-the-art method regarding the accuracy of predicting manually-annotated landmarks using discovered landmarks, and it performs comparably to fully supervised landmark detectors trained with a significant amount of labeled data.
\item The discovered landmarks show strong discriminative performance in recognizing visual attributes. 
\item Our landmark-based image decoder is useful for controllable image decoding, such as object shape manipulation and structure-conditioned image generation. 
\end{enumerate}

\section{Related work}

\paragraph{Discriminative part learning.}

Parts are commonly used object structures in computer vision. 
The deformable part-based
model~\citep{dpm} learns object part configurations to optimize
the object detection accuracy, where similar ideas are rooted in earlier
constellation approaches~\citep{recognition-scale-invariance-learning,obj-categ-discovery,recognition-local-global}.
A recent method~\citep{deformable-parts-pose} based on the deep
neural network performs end-to-end learning of deformable mixture
of parts for pose estimation. The recurrent architecture~\citep{rnn}
and spatial transformer network~\citep{stn} are also used to discover
and refine object parts for fine-grained image classification~\citep{HSnet}.
In addition, discriminative mid-level patches can be also discovered
without explicit supervision~\citep{discriminative-patch}. 
Object-part discovery based on subspace analysis and clustering techniques is also shown to improve neural-network-based image recognition~\citep{unsupervised-part-learning}.
Unlike the approaches specific to discriminative tasks, our work focuses on learning landmarks for generic image modeling. 

\paragraph{Learning structural representations.}

To capture the intrinsic structures of objects, existing studies~\citep{disentangling-face,visual-analogy,adversarial-disentangling} disentangle visual content into multiple factors of variations, like the camera viewpoint, motion, and identity. 
The physical parameters of these factors are, however, still embedded in non-perceptible latent representations. 
Methods based on multi-task learning~\citep{mtcnn,mask-rcnn,style-struct-gan,face-landmark-multitask} can take conceptualized structures (e.g., landmarks, masks, depth) as additional outputs. 
These structures in this setting are designed by humans and require supervision to learn. 

\paragraph{Learning explicit structures for image correspondence. }

Object structures create correspondence among object instances. Colocalization~\citep{colocalization-in-real,object-discovery-in-the-wild}
realizes the coarsest level of object correspondence. In a finer granularity,
AnchorNet~\citep{anchor-net} learns object parts and their correspondence
across different objects and categories. WarpNet~\citep{warp-net}
corresponds images in the same class by estimating the parameter of
a thin plate spline (TPS) transformation~\citep{TPS}, and it can
roughly reconstruct 3D point cloud using a single-view image. The
3D interpreter network~\citep{3d-interpreter} utilizes 2D landmark
annotations to discover 3D skeletons as the explicit structures of
objects. 
Our discovered landmarks are denser than object parts and sparser than 3D points. 
These landmark representations are also more sensitive to precise locations and obtained without supervision.

\paragraph{Landmark discovery with equivariance.}

Object structures like landmarks should be equivariant to image transformation, including object and camera motions. 
Using this property in 2D image domain, \citet{conv-geo-matching} proposed to discover TPS control points to match pairs of object images densely. 
\citet{dense-eqv} tried to densely map different objects to a canonical coordinate that reflects object structures.  
Instead of learning dense correspondence, \citet{unsupervised-landmark}
took the same equivariance property as the guidance to train deep
neural networks for object landmark discovery without manual supervision.
A similar idea was formulated differently using hand-crafted features
in early work~\citep{covariant-features}. In comparison, our method
not only takes the equivariance as a constraint to ensure the validity
of the landmarks, but also use a differentiable formulation to incorporate
the landmark coordinates into a generic image modeling process. Moreover,
our discovered landmarks are more predictive of manually annotated
landmarks than those obtained by \citet{unsupervised-landmark}, and
our method works on a broader range of object categories.

\begin{figure*}
\vspace*{-0.1in}
\begin{centering}
\includegraphics[width=0.9\textwidth]{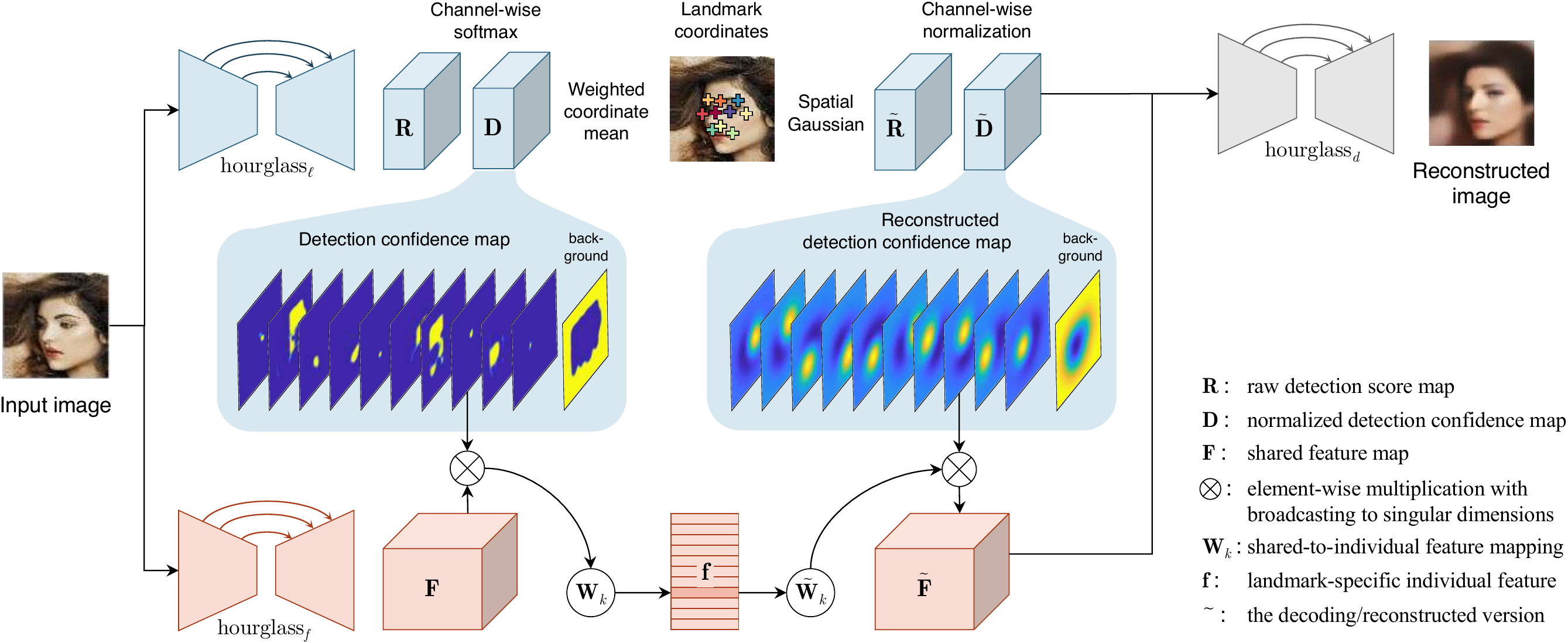}
\par\end{centering}
\caption{\label{fig:arch} 
Neural network architectures of our autoencoding framework for unsupervised landmark discovery. See text for the details.
}
\optvspace{5pt}
\end{figure*}
 
\paragraph{Image modeling with landmarks. }

Many unsupervised deep learning techniques exist to model visual content,
including stacked autoencoders (SAE)~\citep{SAE,scae}, variational
autoencoders~\citep{vae}, generative adversarial networks
(GAN)~\citep{gan,dc-gan}, and auto-regressive networks~\citep{pixel-rnn}
(e.g., PixelCNN~\citep{pixel-cnn}).  
The GAN- and PixelCNN-based image generators conditioned on given object landmarks are proposed in \citep{wwgan,par-pixel-cnn}. 
In contrast, our method uses the SAE framework to automatically discover landmarks that are informative for unsupervised image modeling.

\paragraph{Landmark detection. }

A vast amount of supervised landmark detection methods exist in the literature. 
For human faces, there are active appearance models~\citep{aam,aam-revisit,aam-clm},
template-based methods~\citep{deformable-field-face-landmark,face-det-pose-landmark},
regression-based methods\,\citep{face-boosted-regress,face-regress-forest,face-shape-regress,face-regress-lbp},
and more recent methods based on deep neural networks~\citep{face-deep-cascade,face-coarse-to-fine,face-landmark-multitask,face-landmark-aux-attr,point-transformer-network,face-landmark-recurrent,face-landmark-occlusion,face-landmark-2stages,face-landmark-2d3d-duel}.
Landmark detection methods are also available for human bodies~\citep{human-pose-yiyang-thesis,deep-pose,hourglass},
and birds~\citep{point-transformer-network}. We use our discovered
landmarks to predict manually annotated landmarks and compare our method with some recent supervised models. 
 
\section{Autoencoding-based landmark discovery}
\label{sec:method}
We aim at automatically discovering landmarks as an explicit representation of visual content. 
We propose an autoencoder that encodes landmark coordinates as (a part of) the encoder outputs (Section~\ref{sec:landmark-detector}).
Without supervision from hand-crafted labels, we introduce several constraints to encourage the discovered landmark coordinates to reflect the visual concept that agrees with human perception (Section~\ref{sec:landmark-concept}). 
The proposed constraints prevent landmark-based representations from degenerating to non-perceptible latent representations.
Another pathway of the encoder extracts the local latent descriptor for each discovered landmark (Section~\ref{sec:local-descriptor}).
We use both the landmarks and the latent descriptors to reconstruct the input image (Section~\ref{sec:decoder}).
This section presents the fully differentiable neural network architecture (Figure~\ref{fig:arch}) and training objectives (Section~\ref{sec:objective}) for landmark discovery and unsupervised image modeling. 

\subsection{Architecture of landmark detector}
\label{sec:landmark-detector}

We formulate landmark localization as the problem of detecting particular keypoints in the image~\citep{hourglass}.
Specifically, each landmark has a corresponding detector, which convolutionally outputs a detection score map with the detected landmark located at the maximum. 
In this framework, we use a deep neural network to transform an image $\mathbf{I}$ to a $(K+1)$-channel detection
confidence map 
$\mathbf{D}\in[0,1]^{W\times H\times(K+1)}$.
This map detects $K$ landmarks, and the $(K+1)$-th channel represents background. 
$\mathbf{D}$'s resolution $W\times H$ can be either equal to or less than that of $\mathbf{I}$, but they should have the same aspect ratio. 
Inspired by the success of the stacked hourglass network in human pose estimation~\citep{hourglass}, we propose a light-weighted hourglass-style network to get the raw detection score map
\begin{equation}
\mathbf{R}=\op{hourglass}_{\ell}(\mathbf{I};\theta_{\mathrm{\ell}})\in\mathbb{R}^{W\times H\times(K+1)},\label{eq:im2rawdet}
\end{equation}
where $\theta_{\mathrm{\ell}}$ denotes the parameters. 
The hourglass-style architecture (\supp{\ref{supp:arch-details}}) allows detectors to focus on the critical local patterns at landmark locations while utilizing higher-level context. 
Then, we transform the unbounded raw scores to probabilities and encourage each channel to detect a different pattern. 
To this end, we normalize $\mathbf{R}$ across the channels (including the background) using softmax and obtain the detection confidence map
\begin{equation}
\mathbf{D}_{k}(u,v)=\frac{\exp(\mathbf{R}_{k}(u,v))}{\sum_{k'=1}^{K+1}\exp\left(\mathbf{R}_{k'}(u,v)\right)},\label{eq:enc-norm-det}
\end{equation}
where the matrix $\mathbf{D}_{k}$ is the $k$-th channel of $\mathbf{D}$, and the scalar $\mathbf{D}_{k}(u,v)$  is the value of $\mathbf{D}_{k}$ at the pixel $(u,v)$. 
Later, we also use the vector $\mathbf{D}(u,v) \in [0,1]^{K+1}$ to denote the multi-channel values of $\mathbf{D}$ at $(u,v)$. 
The same notation convention applies to other tensors of three axes. 

Taking $\mathbf{D}_{k}$ as a weighting map, we use
the weighted mean coordinate as the location of the $k$-th landmark, i.e., 
\begin{equation}
(x_{k},y_{k})=\frac{1}{\zeta_k} \sum_{v=1}^{H}\sum_{u=1}^{W}(u,v) \cdot \mathbf{D}_{k}(u,v), \label{eq:det2lm}
\end{equation}
where $\zeta_k=\sum_{v=1}^{H}\sum_{u=1}^{W}\mathbf{D}_{k}(u,v)$ is the spatial normalization factor.
This formulation enables back-propagating
the gradient from the downstream neural network through the landmark
coordinates unless $\mathbf{D}_{k}$'s mass is totally concentrated
in a single pixel or totally uniformly distributed, which rarely happens in practice. 
As a shorthand notation, we write the landmarks and landmark detector
as 
\begin{equation}
\boldsymbol{\ell}=[x_{1},y_{1},\ldots,x_{K},y_{K}]^{\top}=\op{landmark}(\mathbf{I};\theta_{\ell}).\label{eq:lm-det}
\end{equation}
The left half of the blue pathway in Figure~\ref{fig:arch} illustrates
the landmark detector.

\subsection{Visual concept of landmarks}
\label{sec:landmark-concept}

The elements in $\boldsymbol{\ell}$ are supposed to be the discovered landmark coordinates, but so far, there is no guarantee to prevent
them from being arbitrary latent representations. 
Therefore, we propose the following soft constraints as regularizers to enforce the desirable properties for landmarks.

\paragraph{Concentration constraint}

As a detection confidence map for a single location, the mass of $\mathbf{D}_{k}$ need to be concentrated in a local region. 
Taking $\mathbf{D}_{k}/\zeta_k$ (spatially normalized as in \eqref{eq:det2lm}) as the density of a bivariate distribution on the image coordinate, we compute its variance $\sigma^2_{\mathrm{det},u}$ and $\sigma^2_{\mathrm{det},v}$ along the two axes. 
We define the concentration constraint loss as follows to encourage both variances to be small:  
\begin{equation}
L_{\mathrm{conc}}=2 \pi e \left(\sigma^2_{\mathrm{det},u}+\sigma^2_{\mathrm{det},v}\right)^2. 
\end{equation}
This equation makes $L_{\mathrm{conc}}$ the exponential of the entropy of the isotropic Gaussian distribution 
$\mathcal{N}( (x_k, y_k), \sigma^2_{\mathrm{det}} \identitymat )$, where $\sigma^2_{\mathrm{det}} = (\sigma^2_{\mathrm{det},u}+\sigma^2_{\mathrm{det},v})/2$, and $\identitymat$ is the identity matrix. 
This Gaussian distribution is an approximation of $\mathbf{D}_k/\zeta_k$, and lower entropy means a more peaked distribution. 
Note that, formally, this approximation is  
\begin{equation}
\normalized{\mathbf{D}}_{k}(u,v)=(1/WH) \mathcal{N}\left( (u,v) ; (x_k, y_k), \sigma^2_{\mathrm{det}}\identitymat \right). \label{eq:regular-map}
\end{equation}

\paragraph{Separation constraint}

Ideally, the autoencoder training objective can automatically encourage the $K$ landmarks to be distributed at \emph{different} local regions so that the whole image can be reconstructed. 
However, the initial randomness can make the landmarks, defined as the mean coordinates weighted by $\mathbf{D}$ as in \eqref{eq:det2lm}, all around the image center in the beginning of the training. 
This can lead to local optima from which the gradient descent may not escape (see \supp{\ref{supp:heatmap-evolution}}).
To circumvent this difficulty, we introduce an explicit loss to spatially separate the landmarks: 
\begin{equation}
L_{\mathrm{sep}}=\sum_{k\neq k'}^{1,\ldots,K}\exp\left(-\frac{\left\Vert (x_{k'},y_{k'})-(x_{k},y_{k})\right\Vert _{2}^{2}}{2\sigma_{\mathrm{sep}}^{2}}\right).\label{eq:sep-loss}
\end{equation}

\paragraph{Equivariance constraint}

A landmark should locate a stable local pattern (with definite semantics). 
This requires landmarks to show equivariance to image transformations. More specifically, a landmark should move according to the transformation (e.g., camera and object motion) applied to the image if the corresponding visual semantics still exist in the transformed image. 
Let $g(\cdot,\cdot)$ be a coordinate transformation that map image $\mathbf{I}$ to $\mathbf{I}'(u,v)=\mathbf{I}(g(u,v))$, and $\boldsymbol{\ell}'=[x_{1}',y_{1}',\ldots,x_{K}',y_{K}']^{\top}=\op{landmark}(\mathbf{I}')$.
We ideally have $g(x_{k}',y_{k}')=(x_{k},y_{k})$, inducing the soft constraint 
\begin{equation}
L_{\mathrm{eqv}}=\sum_{k=1}^{K}\left\Vert g(x_{k}',y_{k}')-(x_{k},y_{k})\right\Vert _{2}^{2},\label{eq:eqv-loss}
\end{equation}
This loss function is well-defined when $g$ is known. 
Inspired by \citet{unsupervised-landmark}, we simulate $g$ by a thin plate spline (TPS)~\citep{TPS} with random parameters. 
We use random translation, rotation, and scaling to determine the global affine component of the TPS; and, we spatially perturb a set of control points to determine the local TPS component. 
Besides the conventional way of selecting TPS control points at a predefined uniform grid (as used in \citep{unsupervised-landmark}), we also take the landmarks detected by the current model as the control points to improve simulated transformation's focus on key image patterns.
The two sets of control points are alternatively used in each optimization iteration (see \supp{\ref{supp:tps-control}} for details). 
Moreover, when training sample appear in the form of video, we can also take the dense motion flow as $g$ and the actual next frame as $\mathbf{I}'$. 

\paragraph{Cross-object correspondence}

Our model does not explicitly ensure the semantic correspondence among the landmarks discovered on different object instances. 
The cross-object semantic stability of the landmarks mainly relies on the fact that visual patterns activating the same convolutional filter are likely to share semantic similarities. 

\subsection{Local latent descriptors}
\label{sec:local-descriptor}

For simple images, like in MNIST~\citep{mnist} (see results for MNIST in \supp{\ref{supp:mnist}}), multiple landmarks can be enough to describe the object shapes. 
For most natural images, however, landmarks are insufficient to represent all visual content, so extra latent representations are needed to encode complementary information. 
Though necessary, the latent representations should not encode too much holistic information that can overwhelm the image structures reflected by the landmarks. 
Otherwise, the autoencoder would not provide enough driving force to localize landmarks at meaningful locations. 
To achieve this trade-off, we attach a low-dimensional local descriptor to each landmark. 

An hourglass-style neural network (see \supp{\ref{supp:arch-details}}) is introduced to obtain a feature map $\mathbf{F}$, which has the same size as the detection confidence map $\mathbf{D}$:
\begin{equation}
\mathbf{F}=\op{hourglass}_{f}(\mathbf{I};\theta_{f})\in\mathbb{R}^{W\times H\times S}.\label{eq:im2feat}
\end{equation}
Note that $\mathbf{F}$ is in a feature space shared among all landmarks and has $S$ channels.

For each landmark, we use an average pooling weighted by a soft mask centered at the landmark to extract the local feature in the shared space. 
In particular, we take $\normalized{\mathbf{D}}_{k}$, which is the Gaussian approximation of the detection confidence map defined in \eqref{eq:regular-map}, as the soft mask.
Then, a learnable linear operator is introduced for each landmarks to map the feature representation into a lower-dimensional individual space. 
Thus, the latent descriptor for the $k$-th landmark is
\begin{equation}
\mathbf{f}_{k}=\mathbf{W}_{k}\sum_{v=1}^{H}\sum_{u=1}^{W}\left(\normalized{\mathbf{D}}_{k}(u,v)\cdot\mathbf{F}(u,v)\right)\in\mathbb{R}^{C},\label{eq:lm-feat}
\end{equation}
where $C<S$. 
The landmark-specific linear operator enables each landmark descriptor to encode a particular pattern in limited bits. 
We can also use (\ref{eq:lm-feat}) to extract a low-dimensional background descriptor. 
Since it is unreasonable to approximate the background confidence map with a Gaussian distribution, we exactly set $\normalized{\mathbf{D}}_{K+1}=\mathbf{D}_{K+1}/\zeta_{K+1}$.
Note that $\mathbf{f}_{k}$ is differentiable regarding both the feature map and the detection confidence map. 

Putting all latent descriptors together, we have $\mathbf{f}=\op{vec}\left(\left[\mathbf{f}_{1},\mathbf{f}_{2},\ldots,\mathbf{f}_{K+1}\right]\in\mathbb{R}^{C\times(K+1)}\right)$.
The left half of the red pathway in Figure~\ref{fig:arch} illustrates the neural network architecture to extract the landmark descriptors.

\subsection{Landmark-based decoder}
\label{sec:decoder}

We approximately invert the landmark coordinates to the detection confidence map $\tilde{\mathbf{D}}\in\mathbb{R}^{W\times H\times(K+1)}$. 
Concretely, we use the probability density of an isotropic Gaussian distribution centered at each landmark to get raw score maps 
\begin{equation}
\tilde{\mathbf{R}}_{k}(u,v) \hspace{-1pt} = \hspace{-1pt} \mathcal{N}\left((u,v);(x_{k},y_{k}),\sigma_{\mathrm{dec}}^{2}\identitymat \right) \hspace{-1pt} , \,
\tilde{\mathbf{R}}_{K+1} \hspace{-1pt} = \hspace{-1pt} \mathbf{1}.\label{eq:lm2rawdet}
\end{equation}
and the background channel is set to $1$. 
$\tilde{\mathbf{R}}$ is then normalized across channels to obtain the reconstructed detection confidence map
\begin{equation}
\tilde{\mathbf{D}}(u,v)=\tilde{\mathbf{R}}_{k}(u,v)/\sum_{k=1}^{K+1}\tilde{\mathbf{R}}_{k}(u,v).\label{eq:dec-norm-det}
\end{equation}
Figure~\ref{fig:arch} (right half of the blue pathway) illustrates this.

For each landmark (including the background) descriptor $\mathbf{f}_{k}$, we transform it into a shared feature space by the landmark-specific operator $\tilde{\mathbf{W}}_{k}$ and an activation function (e.g., LeakyReLU~\citep{lrelu}). 
Using $\tilde{\mathbf{D}}$ as the soft switches for global unpooling, we recover the feature map
\begin{equation}
\tilde{\mathbf{F}}(u,v)=\sum_{k=1}^{K+1}\tilde{\mathbf{D}}_{k}(u,v)\cdot \tau(\tilde{\mathbf{W}}_{k}\mathbf{f}_{k}) \in\mathbb{R}^{W\times H\times S},\label{eq:feat2map}
\end{equation}
where $\tau(\cdot)$ is the non-linear activation function. 
This is illustrated by the right half of the red pathway in Figure~\ref{fig:arch}.

Though alternative neural network architectures are available (e.g., in \citep{wwgan,par-pixel-cnn}) for landmark-conditioned image decoding, our proposed architecture enables back-propagation through the landmark coordinates. 
The Gaussian variance $\sigma_{\mathrm{dec}}^{2}$ determines how much the neighboring pixels can contribute to the gradients for the landmark coordinates and how sharp the descriptor is localized in the recovered feature map. 
While it is important to include more pixels for back-propagation in the early stage of training, sharpness becomes more important as training goes on. 
To balance the two needs, we obtain multiple versions of $\tilde{\mathbf{D}},\tilde{\mathbf{F}}$ under different values of $\sigma_{\mathrm{dec}}$, say, $(\tilde{\mathbf{D}}^{1},\tilde{\mathbf{F}}^{1}),(\tilde{\mathbf{D}}^{2},\tilde{\mathbf{F}}^{2}),\ldots,(\tilde{\mathbf{D}}^{M},\tilde{\mathbf{F}}^{M})$. 

Let $\llbracket\cdots\rrbracket$ be the channel-wise concatenation.
We use another hourglass-style network to reconstruct the image
\begin{equation}
\tilde{\mathbf{I}}=\op{hourglass}_{d}(\llbracket\tilde{\mathbf{D}}^{1},\tilde{\mathbf{F}}^{1},\ldots,\tilde{\mathbf{D}}^{M},\tilde{\mathbf{F}}^{M}\rrbracket;\theta_{d}) \label{eq:map2im}
\end{equation}
The gray pathway in Figure~\ref{fig:arch} illustrates the image decoder.

\subsection{Overall training objective}

\label{sec:objective}

The image reconstruction loss $L_{\mathrm{recon}}$ drives the training of the entire autoencoder. 
We define $L_{\mathrm{recon}}$ as $\Vert \mathbf{I} - \tilde{\mathbf{I}} \Vert_F ^2$, and $\mathbf{I}$ is normalized to $[0,1]$.
The full loss is $L_{\mathrm{AE}}=$
\begin{equation}
\lambda_{\mathrm{recon}}L_{\mathrm{recon}}+\lambda_{\mathrm{conc}}L_{\mathrm{conc}}+\lambda_{\mathrm{sep}}L_{\mathrm{sep}}+\lambda_{\mathrm{eqv}}L_{\mathrm{eqv}}.\label{eq:ae-loss}
\end{equation}

\section{Experiments}

We evaluate our method on a variety of datasets, including CelebA~\citep{celeba} and AFLW~\citep{aflw} for human faces, the cat head dataset~\citep{cat-head}, a car dataset built from PASCAL 3D~\citep{pascal-3d}, shoe images from UT Zappos50k~\citep{shoe-dataset}, human pose images from Human3.6M~\citep{human36m,human80k}, MNIST (\supp{\ref{supp:mnist}}), and animal images from AwA~\citep{awa} (\supp{\ref{supp:animals}}).  

Section~\ref{subsec:exp-lm-discovery} describes the datasets and shows the qualitative results of landmark discovery. 
In Section~\ref{subsec:exp-gtlm-prediction}, we use the discovered landmarks to predict human-annotated landmarks, and we take the landmark detection accuracy as an indicator of the quality of discovered landmark. 
Section~\ref{subsec:exp-attr-recognition} demonstrates that our discovered landmarks can serve as effective image representations to predict shape-related facial attributes on CelebA. 
In Section~\ref{subsec:exp-attr-recognition}, we show that our decoding module and the automatically discovered landmarks can be used to manipulate the object shapes. 

\subsection{Landmark discovery on multiple datasets}

\label{subsec:exp-lm-discovery}

We train and evaluate landmark discovery models on a variety of objects.
The detailed architectures of the neural network modules (i.e., $\op{hourglass}_{\ell|f|d}$) depend on the image sizes on different datasets. 
\supp{\ref{supp:implementation}} describes implementation details, including data preprocessing, network architectures, model parameters, and optimization methods. 

\begin{figure}
\begin{centering}
\includegraphics[width=1\columnwidth]{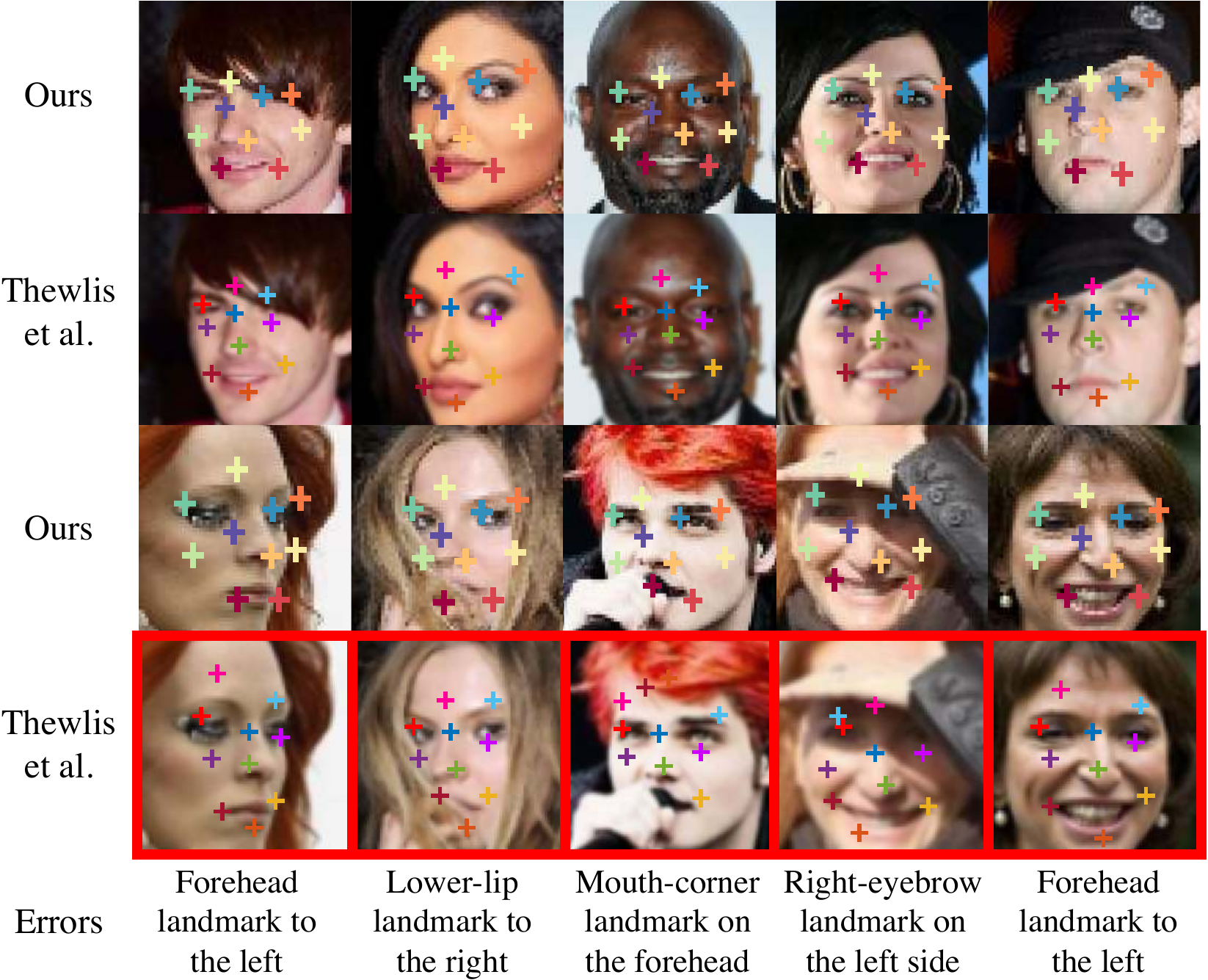}
\par\end{centering}
\caption{\label{fig:celeba-kp10}Discovering 10 landmarks on CelebA images.
All figures for \citet{unsupervised-landmark}'s come from their paper. 
The last row shows unsuccessful cases from \citep{unsupervised-landmark} with error descriptions below.}
\optvspace{0.01in}
\end{figure}

\begin{figure}
\begin{centering}
\includegraphics[width=.22\columnwidth]{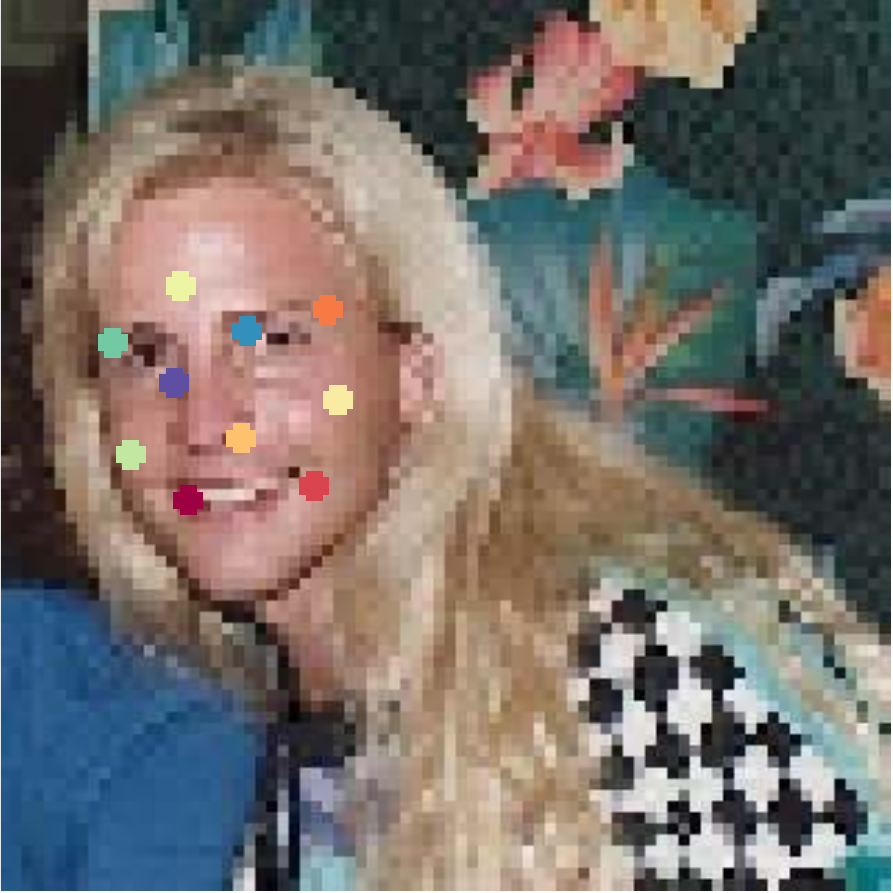}
\includegraphics[width=.22\columnwidth]{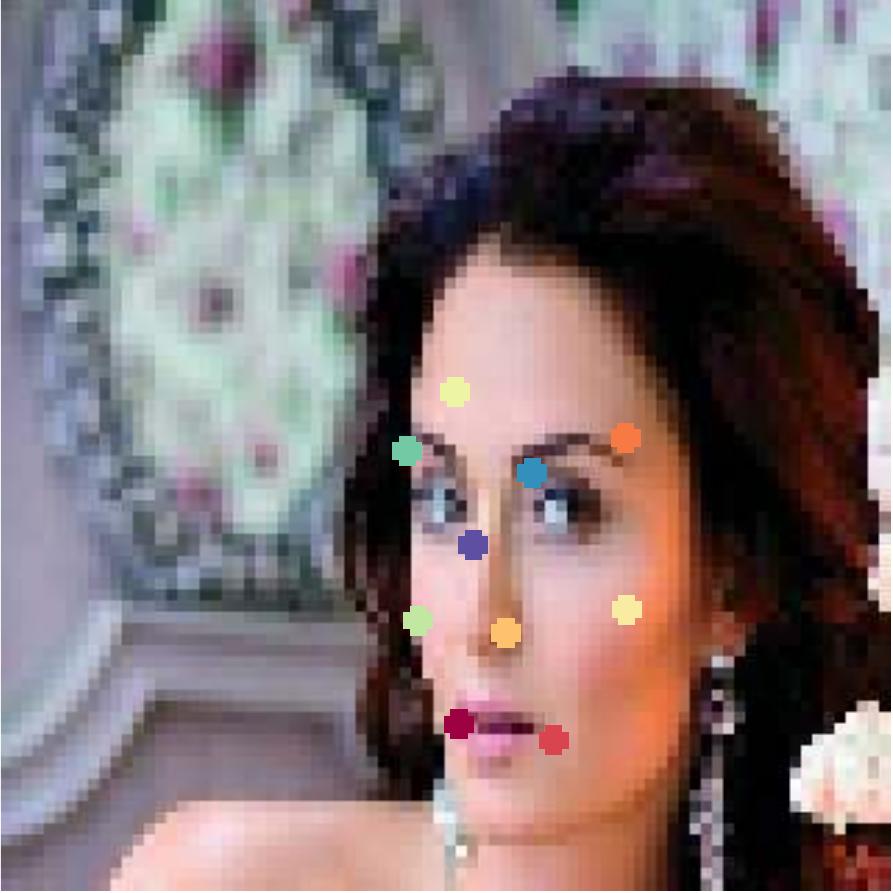}
\includegraphics[width=.22\columnwidth]{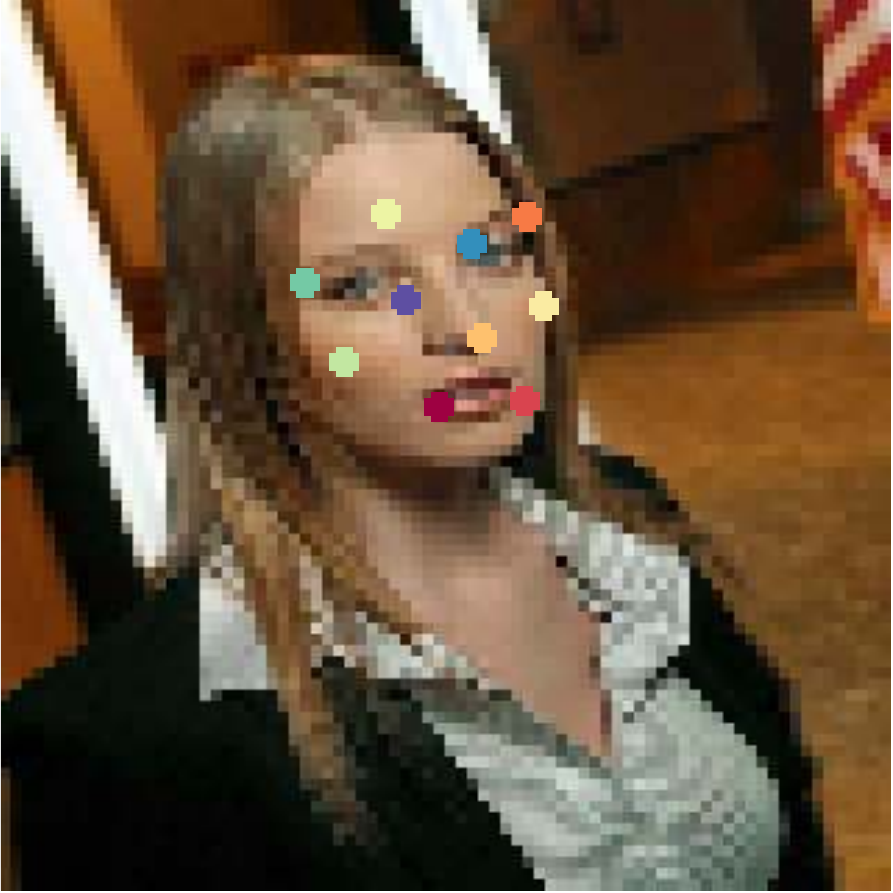}
\includegraphics[width=.22\columnwidth]{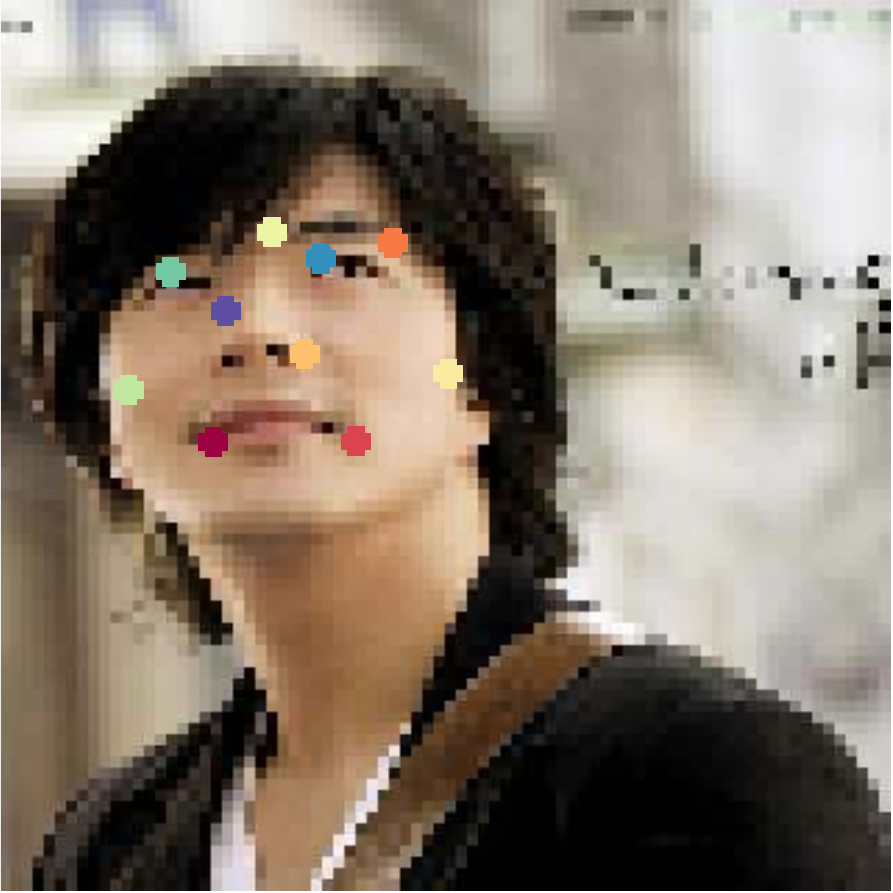}
\par\end{centering}
\optvspace{0.01in}
\caption{\label{fig:celeba-kp10-head-shoulder}
Discovering 10 landmarks on unaligned head-shoulder images using our model trained on aligned facial images. }
\end{figure}

\begin{figure}
\begin{centering}
\includegraphics[width=1\columnwidth]{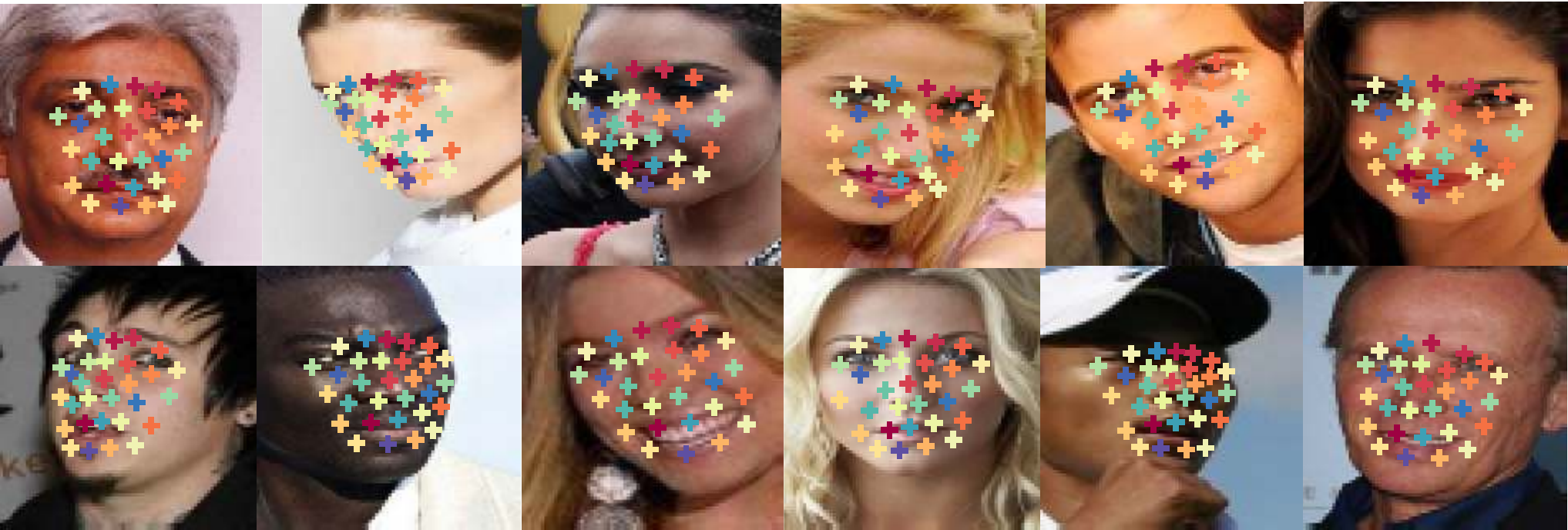}
\par\end{centering}
\optvspace{0.01in}
\caption{\label{fig:celeba-kp30-unaligned}Discovering 30 landmarks on unaligned CelebA images using our method. }
\end{figure}

\paragraph{CelebA}

Following \citep{unsupervised-landmark}, we use all facial images in the CelebA training set excluding those also appearing in the MAFL the test set\footnote{The MAFL dataset~\citep{face-landmark-multitask} is a subset of CelebA.} 
(then 16,1962 images in total) to train models for landmark discovery.
We use the MAFL testing set (1000 images) for all testing cases and reserve the MAFL training set (19,000 images) to train prediction models for manually-annotated landmarks. 
By default, we use the cropped and aligned images provided in the dataset. 

As shown in Figure~\ref{fig:celeba-kp10}, our method can automatically discover facial landmarks at semantically meaningful and stable locations, such as the forehead center, eyes, eyebrows, nose, and mouth corners. 
Compared to \citet{unsupervised-landmark}'s method, which results in a few significant errors, our method can locate landmarks more robustly against pose variations and occlusions. 
Interestingly, our method can work out-of-the-box on head-shoulder portraits without training on exactly the same type of images (Figure~\ref{fig:celeba-kp10-head-shoulder}).
Figure~\ref{fig:celeba-kp30-unaligned} shows that our method can also learn and detect a larger number (e.g., 30) of high-quality landmarks on unaligned facial images. 
\supp{\ref{supp:celeba}} shows more results.

\paragraph{AFLW}

Face images in AFLW are cropped differently from CelebA. 
The landmark discovery models (both ours and \citet{unsupervised-landmark}'s) are pretrained on CelebA and finetuned on the AFLW training set (10,122 images) for adaptation. 
Sampled results on the AFLW testing set (2,991 images) are in \supp{\ref{supp:aflw}}. 

\begin{figure}
\begin{centering}
\includegraphics[width=1\columnwidth]{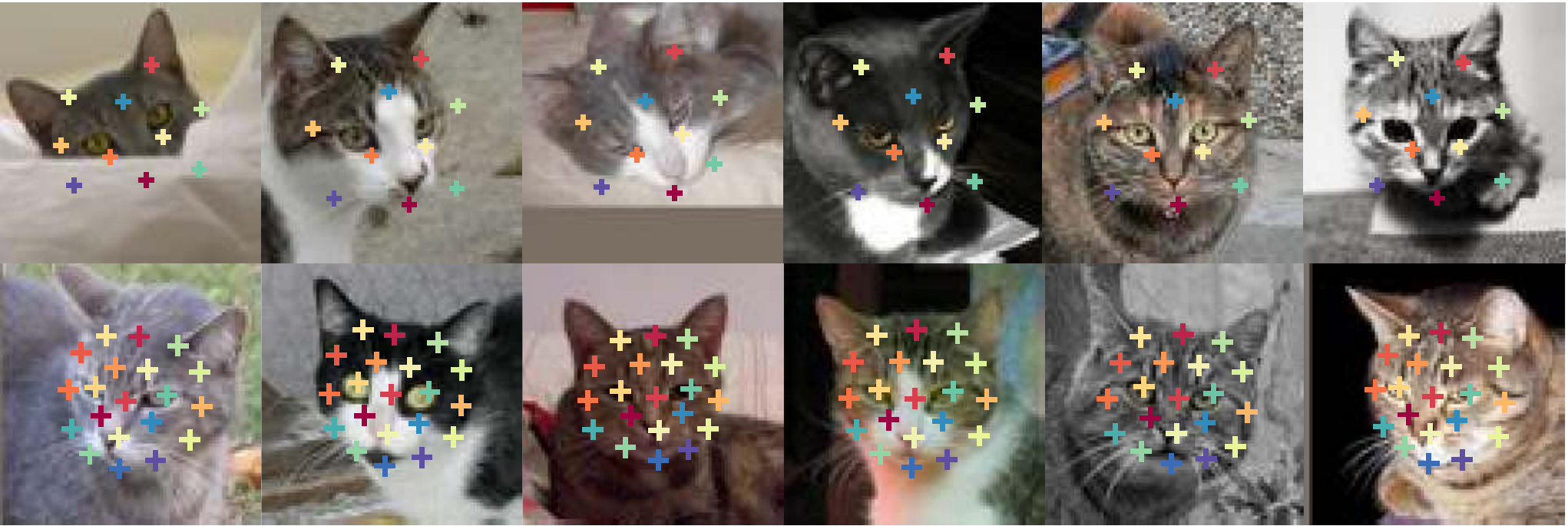}\\
\par\end{centering}
\caption{\label{fig:cat-kp-10and20}
Discovering landmarks on cat head images using our method. 
Top row: 10 landmarks; Bottom row: 20 landmarks.}
\optvspace{2pt}
\end{figure}

\begin{figure}
\begin{centering}
\includegraphics[width=1\columnwidth]{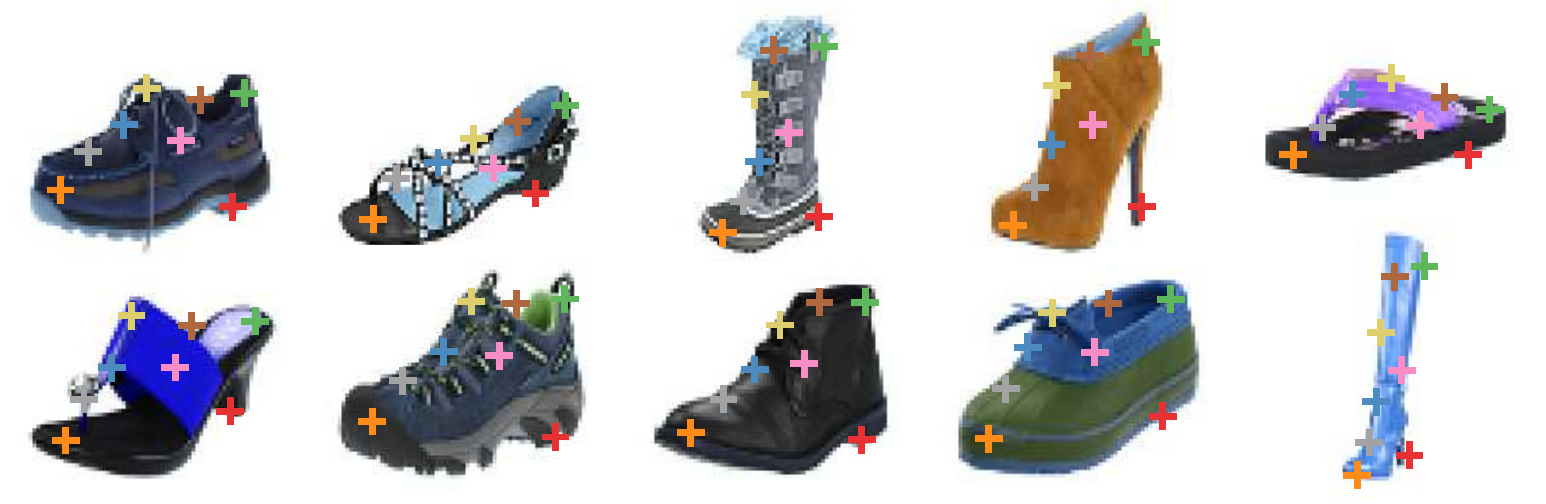}\\
\par\end{centering}
\caption{\label{fig:shoe-kp8}Discovering 8 landmarks on shoes.}
\optvspace{2pt}
\end{figure}

\paragraph{Cat heads}

Our model is trained on 7,747 cat head images and tested on 1,257 images. 
Compared to human faces, cat heads show more holistic appearance variations. 
As shown in Figure~\ref{fig:cat-kp-10and20}, our model can discover consistent landmarks (e.g., ears, nose, mouth) across different cat species and interestingly predict landmark locations under significant occlusion (the first image). 
\supp{\ref{supp:cat}} shows more results.

\paragraph{Cars}

\begin{figure}
\begin{centering}
\includegraphics[width=1\columnwidth]{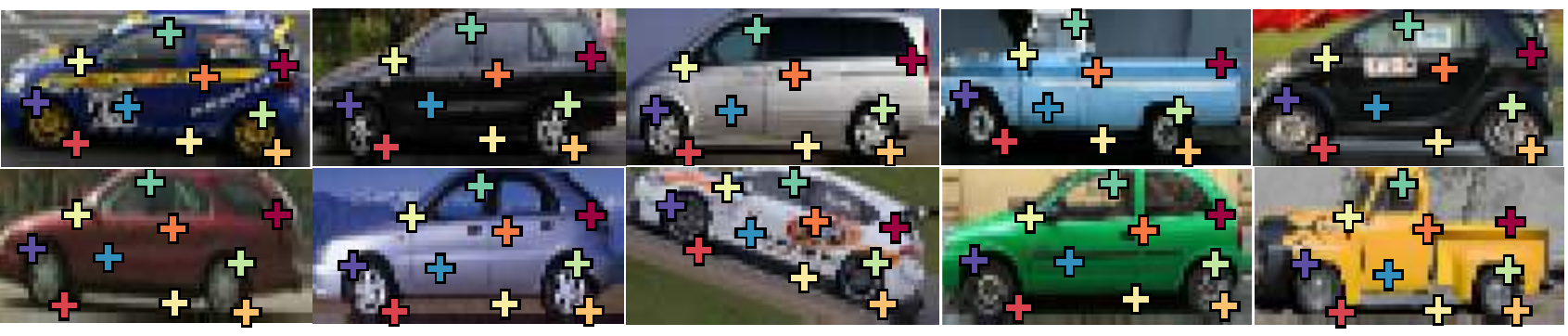}\\
\par\end{centering}
\optvspace{2pt}
\caption{\label{fig:car-kp10}
Discovering 10 landmarks on the profile images of cars.}
\end{figure}

We build the profile-view car dataset by cropping the car images from the PASCAL 3D dataset. 
This dataset has a limited number of samples (567 images for training and 63 images for testing). 
As shown in Figure~\ref{fig:car-kp10}, our method can still learn meaningful landmarks (e.g., the windshield, driver-side door, wheels, rear) using a relatively small training set. 
Note that we transform the 3D annotations of the cars to 2D landmarks, so this dataset is ready for quantitative evaluation. 
\supp{\ref{supp:cars}} shows more results.

\paragraph{Shoes}

We use the same setting as in \citep{unsupervised-landmark} (49,525, training images and 500 testing images). 
As shown in Figure~\ref{fig:shoe-kp8}, landmarks are detected at semantically stable locations for different types of shoes.
\supp{\ref{supp:shoes}} shows more results.

\paragraph{Human3.6M}

\begin{figure}
\begin{centering}
\includegraphics[width=0.8\columnwidth]{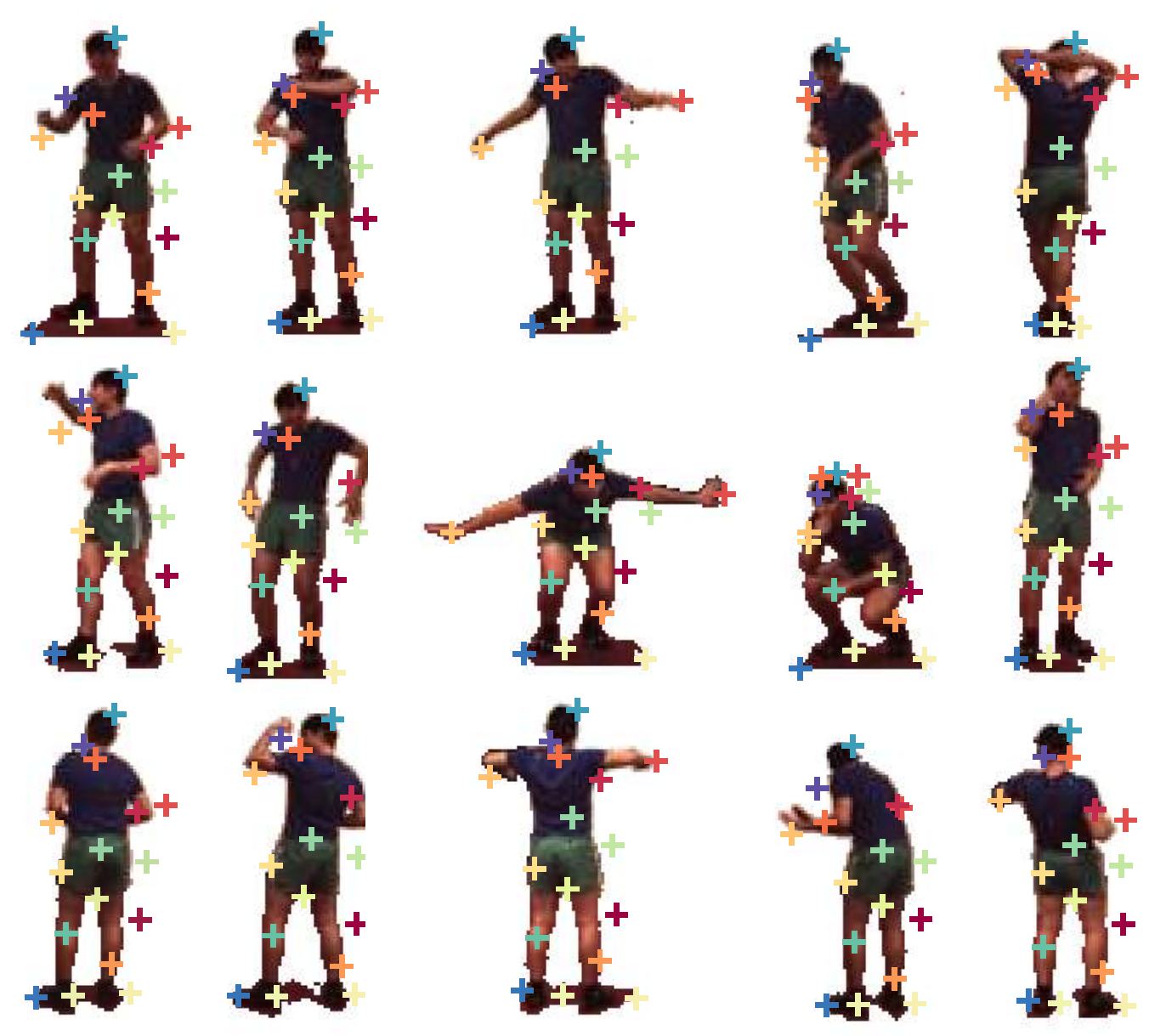}
\par\end{centering}
\caption{\label{fig:huam-kp16-of5}Discovering 16 landmarks on Human3.6M dataset.
\todo{maybe to add some figures for Oxford paper}}
\end{figure}

Human3.6M contains human activity videos in stable backgrounds.
We use all 7 subjects in Human3.6M training set for our evaluation (6 for training and 1 for validation)\footnote{Training subject IDs: S1,S5,S6,S7,S8,S9; Validation subject IDs: S11.}.
We consider six activities (direction, discussion, posing, waiting, greeting, walking), in which human bodies are in the upright direction most of the time, resulting in 796,648 image frames for training and 87,975 image frames for testing. 
We removed the background using the off-the-shelf unsupervised background subtraction method provided in the dataset. 
The human bodies are cropped and roughly aligned regarding the foot location so that the excessive background regions are removed. 

Compared to previously mentioned object types, human bodies have much more shape variations. \todo{talk about optical flow if the experimental results support that}
As shown in Figure~\ref{fig:huam-kp16-of5}, our method can discover roughly consistent landmarks across a range of poses. 
In particular, the landmarks at the head, back, waist, and legs are stable across images. 
The landmarks at the arms are relatively less consistent across different poses, but they are still at semantically meaningful locations.
Since the human body appearances in the frontal and back views are similar, we do not expect our discovered landmarks to distinguish the left and right sides of the human body, which means that a landmark at the left leg in the frontal view can locate the right leg in the back view. 
Since the training data is in the video format, optical flows are used as a short-term self-supervision for the eqvuivariance constraint in \eqref{eq:eqv-loss}. 
\supp{\ref{supp:human}} describes more details and results for Human3.6M experiments.

\subsection{Prediction of ground truth landmarks}
\vspace{-0.02in}

\label{subsec:exp-gtlm-prediction}

Unsupervised landmark learning is useful because of its potential to discover object structures that are coherent with the human's perception.
We evaluate discovered landmarks' quality by predicting manually-annotated landmarks.
Specifically, we use a linear model without a bias term to regress from the discovered landmarks to the human-annotated landmarks. 
Ground truth landmark annotations are needed to train this linear regressor. 
\citet{unsupervised-landmark} extensively used random TPS to augment both discovered and labeled landmarks for training (on CelebA and ALFW). 
However, we do not use data augmentation for our method to minimize the complexity of training. 
Even in this case, our method shows stronger performance. 

\begin{table}[t]
\begin{centering}
\subfloat[\small{}\label{tab:gtlm-celeba-unsup} 
Comparison with unsupervised landmark learning methods on the MAFL testing set. ]{
\begingroup
\small
\setlength\tabcolsep{2pt}
\begin{centering}
\begin{tabular}{>{\centering}m{0.22\columnwidth}|>{\centering}m{0.21\columnwidth}|>{\centering}m{0.19\columnwidth}>{\centering}m{0.19\columnwidth}}
\hline 
\# discovered landmarks & Regressor training set & \citet{unsupervised-landmark} & Ours\tabularnewline
\hline 
10 & CelebA & 6.32 & 3.46\tabularnewline
30 & CelebA & 5.76 & \textbf{3.15}\tabularnewline
50 & CelebA & 5.33 & -\tabularnewline
\hline 
10 & MAFL & 7.95 & 3.46\tabularnewline
30 & MAFL & 7.15 & \textbf{3.16}\tabularnewline
50 & MAFL & 6.67 & -\tabularnewline
\hline 
\end{tabular}
\par\end{centering}
\vspace*{-0.1in}
\endgroup
}
\par\end{centering}
\optvspace{-0.7em}
\begin{centering}
\subfloat[\small{} \label{tab:gtlm-celeba-sup}
Comparison with supervised methods on the MAFL and ALFW testing sets. ]{
\begingroup
\small
\setlength\tabcolsep{2pt}
\begin{centering}
\hspace*{-1ex}\begin{tabular}{c|c|cc}
\hline 
\multicolumn{2}{c|}{Method} & MAFL & ALFW\tabularnewline
\hline 
 & RCPR~\citep{face-landmark-robust-occulsion} & - & 11.60\tabularnewline
 & CFAN~\citep{face-coarse-to-fine} & 15.84 & 10.94\tabularnewline
Fully & TCDCN~\citep{face-landmark-aux-attr} & \textcolor{white}{0}7.95 & \textcolor{white}{0}7.65\tabularnewline
supervised & Cascaded CNN~\citep{face-deep-cascade} & \textcolor{white}{0}9.73 & \textcolor{white}{0}8.97 \tabularnewline
 & RAR~\citep{face-landmark-recurrent} & - & \textcolor{white}{0}7.23\tabularnewline
 & MTCNN~\citep{face-landmark-multitask} & \textbf{\textcolor{white}{0}}\textbf{5.39}  & \textcolor{white}{0}\textbf{6.90}\tabularnewline
\hline 
 & \citet{unsupervised-landmark} (50 landmarks) & \textcolor{white}{0}6.67 & 10.53\tabularnewline
Unsupervised & \citet{dense-eqv} (dense frames) & \textcolor{white}{0}5.83 & \textcolor{white}{0}8.80\tabularnewline
discovery & Ours (10 landmarks) & \textcolor{white}{0}3.46 & \textcolor{white}{0}7.01\tabularnewline
 & Ours (30 landmarks) & \textbf{\textcolor{white}{0}}\textbf{3.15} & \textbf{\textcolor{white}{0}}\textbf{6.58}\tabularnewline
\hline 
\end{tabular}\hspace*{-1ex}
\par\end{centering}
\endgroup
}
\par\end{centering}
\optvspace{-0.7em}
\begin{centering}
\subfloat[\small{} \label{tab:gtlm-celeba-ablation}
Using ablative training losses of our method. Refer to (\ref{eq:ae-loss}) for each loss terms.
Results are obtained on the MAFL testing set using 10 discovered landmarks. ]{
\begingroup
\small
\setlength\tabcolsep{2pt}
\begin{centering}\hspace*{-1ex}
\begin{tabular}{>{\centering}p{0.18\columnwidth}|>{\centering}p{0.18\columnwidth}>{\centering}p{0.18\columnwidth}>{\centering}p{0.18\columnwidth}>{\centering}p{0.18\columnwidth}}
\hline 
Full $L$ & w/o $L_{\mathrm{recon}}$ & w/o $L_{\mathrm{conc}}$ & w/o $L_{\mathrm{sep}}$ & w/o $L_{\mathrm{eqv}}$\tabularnewline
\hline 
3.15 & 3.45 & 3.91 & 16.56 & 8.42\tabularnewline
\hline 
\end{tabular}\hspace*{-1ex}
\par\end{centering}
\endgroup
}
\par\end{centering}
\begin{centering}
\par\end{centering}
\caption{
Mean errors of the annotated landmark prediction on human face datasets. Errors are in \% regarding the biocular distance.
}
\end{table}

\paragraph{Stronger relevance to human-designed landmarks. }

\begin{figure}
\begin{centering}
\includegraphics[width=1\columnwidth]{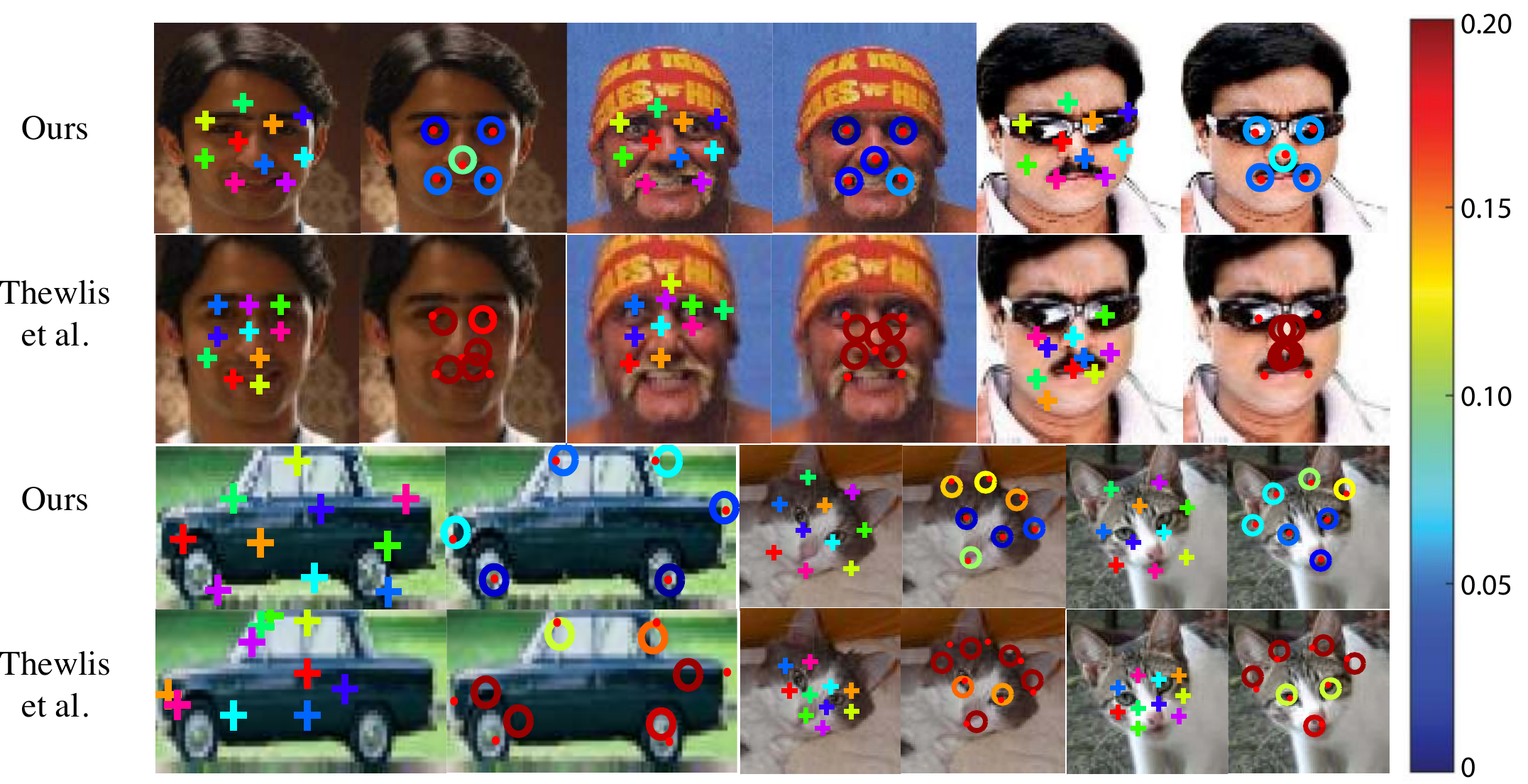}
\par\end{centering}
\caption{\label{fig:gtlm-pred-errors}
Prediction of annotated landmarks. 
Colorful cross: discovered landmark; Red dot: annotated landmark; Circle: regressed landmark, whose color represent its distance to the annotated landmarks. 
See the color bar for the distance (i.e., prediction error).}
\vspace*{0.03in}
\end{figure}

In Table~\ref{tab:gtlm-celeba-unsup}, we regress the landmarks discovered using the models trained on the CelebA training set to the 5 annotated landmarks.
The landmark labels in either the CelebA training set or the much smaller MAFL training set are used to train the regressor.
Our method is not sensitive to the decreased size of the labeled training set.
It outperforms \citet{unsupervised-landmark}'s by 55\% decrease of the landmark detection error and \citet{dense-eqv}'s by 45\%. 
Notably, we achieve this with 30 discovered landmarks while theirs uses 50 landmarks or dense object frames. 
Additionally, Table~\ref{tab:gtlm-celeba-car-cat} demonstrates the consistent superiority of our method on the cat head dataset (7 target landmarks\footnote{9 annotated landmarks in total. We do not use the 2 at the ears.}), the car dataset (6 target landmarks), and Human3.6M\footnote{See \supp{\ref{supp:human}} for details} (32 target landmarks). 
Figure~\ref{fig:gtlm-pred-errors} illustrates the landmark regression results.

\paragraph{Competitive performance compared to fully supervised methods. }

Putting the landmark discovery model together with the linear regressor, we obtain a detector of human-designed landmarks. Unlike fully supervised methods, our model is trainable with a huge amount of unlabeled data, and the linear regressor can be trained using a relatively small amount of labeled data within a few minutes. 
Table~\ref{tab:gtlm-celeba-sup} demonstrates that our model outperforms previous unsupervised methods and off-the-shelf pretrained fully-supervised models on the MAFL and AFLW testing sets. 
On AFLW, we take the 5 always-visible landmarks as the regression target. 
All models reported are either trained on the MAFL training set or publicly available. 

\paragraph{Landmark detection with few labeled samples. }

Taking our model as a detector of manually annotated landmarks, we find that less than 200 samples are enough for our model to achieve less than 4\% mean error on the MAFL testing set, which is better than the performance of TCDCN and MTCNN.
Learning curves are provided in \supp{\ref{supp:number-training-samples}}.

\paragraph{Effectiveness of different loss terms. }
Our method combines several loss terms in the training objective (\ref{eq:ae-loss}).
Table~\ref{tab:gtlm-celeba-ablation} shows that the removal of any term can cause performance drop of our model. 
In particular, the removal of the separation loss can devastate the model, and more detailed discussion about this loss term is in \supp{\ref{supp:heatmap-evolution}}.
Our new differentiable formulation of the landmark validity constraints can already lead to a lower landmark detection error than \citet{unsupervised-landmark}'s.
Adding the reconstruction loss can further improve the accuracy.

\begin{table}
\begingroup
\small
\setlength\tabcolsep{2pt}
\begin{centering}
\begin{tabular}{c|cc|cc|c}
\hline 
Dataset & \multicolumn{2}{c|}{Car} & \multicolumn{2}{c|}{Cat head} & Human3.6M\tabularnewline
\hline 
\# discovered landmarks & 10 & 24 & 10 & 20 & 16\tabularnewline
\hline 
\citet{unsupervised-landmark} & 11.42 & 11.11 & 26.76 & 26.94 & 7.51 \tabularnewline
Ours & \textbf{\textcolor{white}{0}5.87} & \textbf{\textcolor{white}{0}5.80} & \textbf{15.35} & \textbf{14.84} & \textbf{4.14} \tabularnewline
\hline 
\end{tabular}
\par\end{centering}
\endgroup
\vspace*{0.02in}
\caption{\label{tab:gtlm-celeba-car-cat}
Mean errors of the annotated landmark prediction on the cat heads, cars, and human bodies. 
Errors are in \% regarding the biocular distance, bi-wheel distance, and image size, respectively.}
\end{table}

\subsection{Visual attribute recognition}
\label{subsec:exp-attr-recognition}

\begin{table*}
\optvspace{-0.03in}
\begingroup
\footnotesize{}
\setlength\tabcolsep{2pt}
\begin{centering}
\begin{tabular}{l|c|ccccccccccccc|c}
\hline 
Methods & \footnotesize{}\begin{tabular}{@{}l@{}}
Feature\tabularnewline
Dimen-\tabularnewline
sion\tabularnewline
\end{tabular} & \footnotesize{}\begin{tabular}{@{}l@{}}
Arched\tabularnewline
Eyebrows\tabularnewline
\end{tabular} & \footnotesize{}\begin{tabular}{@{}l@{}}
Bags \tabularnewline
Under\tabularnewline
Eyes\tabularnewline
\end{tabular} & \footnotesize{}\begin{tabular}{@{}l@{}}
Big\tabularnewline
Lips\tabularnewline
\end{tabular} & \footnotesize{}\begin{tabular}{@{}l@{}}
Big\tabularnewline
Nose\tabularnewline
\end{tabular} & \footnotesize{}\begin{tabular}{@{}l@{}}
Double\tabularnewline
Chin\tabularnewline
\end{tabular} & \footnotesize{}\begin{tabular}{@{}l@{}}
High\tabularnewline
Cheek-\tabularnewline
bones\tabularnewline
\end{tabular} & \footnotesize{}\begin{tabular}{@{}l@{}}
Male\tabularnewline
\end{tabular}& \footnotesize{}\begin{tabular}{@{}l@{}}
Mouth \tabularnewline
Slightly\tabularnewline
Open\tabularnewline
\end{tabular} & \footnotesize{}\begin{tabular}{@{}l@{}}
Narrow\tabularnewline
Eyes\tabularnewline
\end{tabular} & \footnotesize{}\begin{tabular}{@{}l@{}}
Oval\tabularnewline
Face\tabularnewline
\end{tabular} & \footnotesize{}\begin{tabular}{@{}l@{}}
Pointy\tabularnewline
Nose\tabularnewline
\end{tabular} & \footnotesize{}\begin{tabular}{@{}l@{}}
Receding\tabularnewline
Hairline\tabularnewline
\end{tabular} & \footnotesize{}Smiling  & \footnotesize{}Average\tabularnewline
\hline 
Ours (discovered landmarks) & 60  & 79.4 & 80.9 & 76.9 & 82.3 & 94.5 & 82.5 & 88.4 & 81.3 & 88.0 & 73.2 & 73.7 & 92.1 & 88.8 & 83.2 \tabularnewline
\hline 
FaceNet~\citep{facenet} (top-layer)  & 128  & 76.4 & 80.3 & 76.8 & 80.4 & \textbf{94.5} & 72.6 & 82.7 & 74.4 & 87.9 & 72.7 & 73.1 & 92.2 & 76.2 & 80.0\tabularnewline
FaceNet (top-layer) + Ours & 188  & \textbf{81.3} & \textbf{81.3} & \textbf{77.5} & \textbf{82.6} & \textbf{94.5} & \textbf{83.5} & \textbf{91.2} & \textbf{83.8} & \textbf{88.4} & \textbf{73.7} & \textbf{75.0} & \textbf{92.7} & \textbf{89.9} & \textbf{84.3} \tabularnewline
\hline 
FaceNet~\citep{facenet} (conv-layer)  & 1792 & 78.8 & 81.5 & \textbf{77.4} & 80.5 & 94.6 & 77.3 & 90.0 & 80.9 & 88.4 & 74.2 & 73.6 & 92.4 & 81.5 & 82.4 \tabularnewline
FaceNet (conv-layer) + Ours & 1852 & \textbf{80.1} & \textbf{81.8} & 77.2 & \textbf{82.3} & \textbf{94.7} & \textbf{82.1} & \textbf{90.8} & \textbf{85.0} & \textbf{88.6} & \textbf{74.5} & \textbf{73.6} & \textbf{92.4} & \textbf{90.5} & \textbf{84.1}\tabularnewline
\hline 
\end{tabular}
\par\end{centering}
\endgroup
\arxvspace{-3pt}
\caption{\label{tab:attr-celeba}Visual attribute recognition using pretrained FaceNet features and our discovered 30 landmarks on the MAFL test set. }
\optvspace{3pt}
\arxvspace{-3pt}
\end{table*}

Landmarks reflect object shapes. We use our discovered landmarks as a feature representation to recognize the shape-related binary facial attributes (13 labeled attributes are found) on CelebA. 
We still take the MAFL testing set for the quantitative evaluation. 
A linear SVM is trained for each attribute on the CelebA training set. 
We also compare our landmark coordinates with pretrained FaceNet~\citep{facenet} (InceptionV1) \emph{top-layer} (128-dim) and top \emph{conv-layer} (1792-dim) features for the attribute recognition task. 
As shown in Table~\ref{tab:attr-celeba}, our discovered landmarks (60-dim) outperforms the FaceNet top-layer features for most attributes. 
The conv-layer features outperform our landmarks slightly but have a much higher dimension. 
Combining the landmark coordinates and the FaceNet features, higher accuracy is achieved. 
This suggests that the discovered landmarks are complementary to image features pretrained on classification tasks.

\subsection{Image manipulation and generation}

\label{subsec:exp-manipulation}

Our jointly trained image decoding module conditioned its outputs on the input landmarks and their latent descriptors. 
If the two conditions are disentangled, we should be able to manipulate the object shape without changing other appearance factors by adjusting only the landmarks; 
or, \emph{vice versa}. 
Note that landmark-based image morphing is not a new topic, and landmark-based hierarchical image decoding has also been explored
recently~\citep{wwgan,long-video-predict,par-pixel-cnn}. 
However, these landmarks are all designed and annotated by humans. 
So far, little evidence has suggested that the automatically discovered landmarks are accurate and representative enough as a reliable condition for image generation. 

\begin{figure}
\begin{centering}
\includegraphics[width=1\columnwidth]{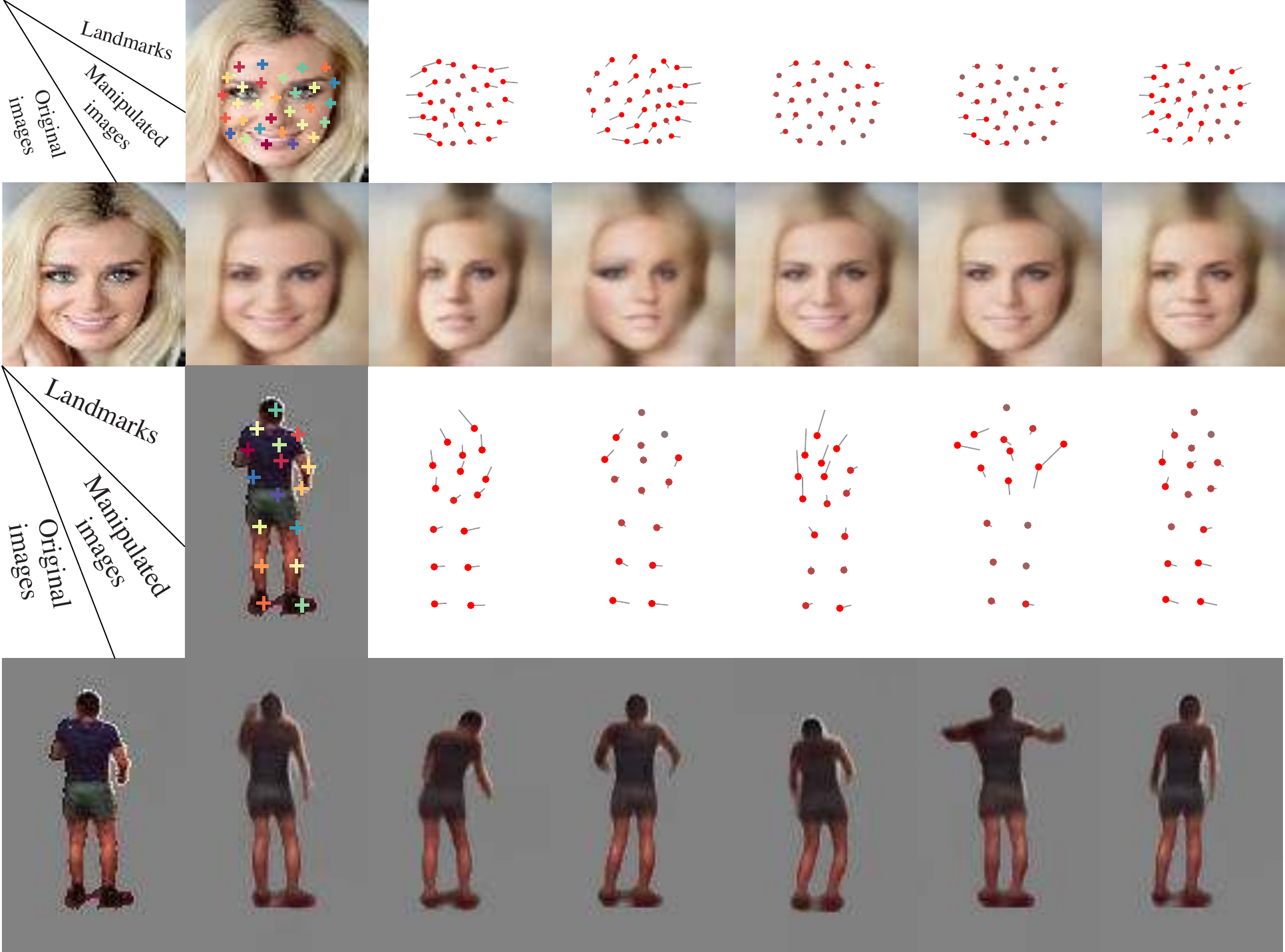}
\par\end{centering}
\caption{\label{fig:face-manipulate}
Image manipulation with our discovered landmarks and landmark-based decoder on the MAFL and Human3.6M testing set.
1st column: input images; 2nd column: discovered landmarks and reconstructed images; other columns: the red dots for new landmark locations, the gray lines for the synthetic adjustment of landmarks, and the images for the decoder outputs. 
}
\optvspace{0pt}
\end{figure}

In Figure~\ref{fig:face-manipulate}, we synthesize flows to adjust the discovered landmarks of an input image. 
Fixing the landmark latent descriptors, we obtain realistic facial and human-body images whose shapes agree with the new landmarks. 
Other than the facial and body shape, then appearance factors of the input image are not visually changed.
This result suggests that our image decoding module can synthesize realistic image using the landmarks learned without supervision, and it also suggests that our discovered landmarks have become an explicit representation disentangled from other factors of variations for image modeling. 
Implementation details and more results about unsupervised landmark-based face manipulation are available in \supp{\ref{supp:manipulation}}. 

\begin{figure}
\begin{centering}
\includegraphics[width=1\columnwidth]{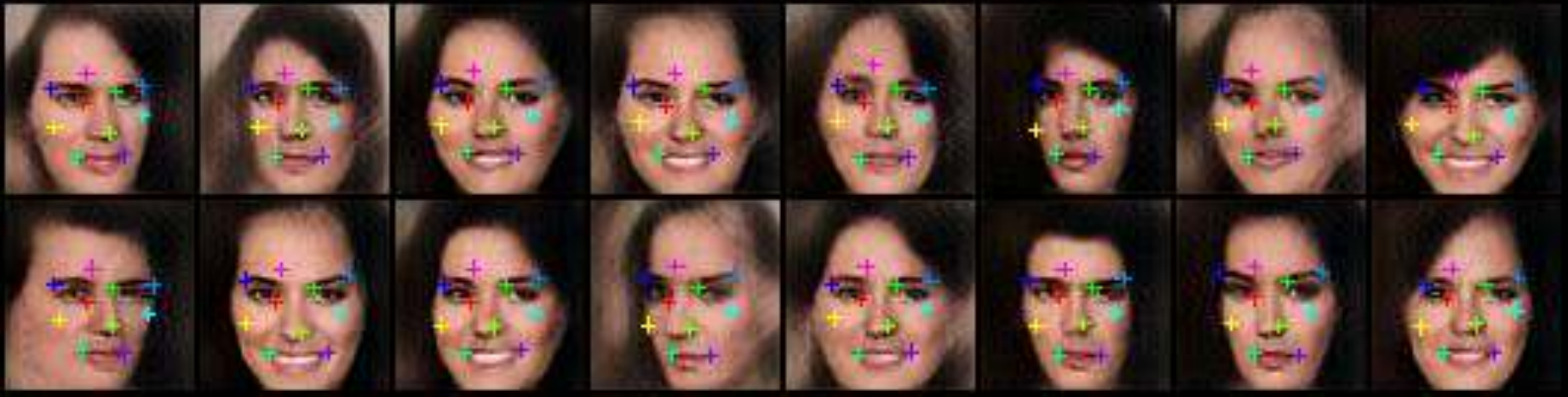}
\par\end{centering}
\caption{\label{fig:face-generation}
Face generation conditioned on discovered landmarks. 
}
\end{figure}

In Figure~\ref{fig:face-generation}, instead of adjusting the landmark coordinates, we use the discovered landmarks of a reference image as the control signal to generate new facial images. 
Following the GAN framework~\citep{gan}, the latent representation of the generated image is randomly drawn from a prior distribution. 
As in \citet{wwgan}, the landmark coordinates and latent representation are combined for image generation. 
We adopt BEGAN~\citep{began} for the discriminator and training objective.
In addition, we apply a cyclic loss for the landmark coordinates, which encourages the same landmarks to be detected on the generated images as on the reference image. 
Our results provide additional evidence on the usefulness of the discovered landmarks for image modeling. 
Implementation details are in \supp{\ref{supp:face-generation-details}}.

\section{Conclusion}

We address the problem of unsupervised object landmark discovery and
take it as an intermediate step of image representation learning.
In particular, a fully differentiable neural network architecture is proposed for determining the landmark coordinates, together with soft contraints to enforce the validity of the detected landmarks. 
The discovered landmarks are visually meaningful and quantitatively
more relevant to human-designed landmarks. In our framework, the discovered
landmarks are an explicit part of the learned image representations.
They are disentangled from the latent representations of the other
appearance factors. The landmark-based explicit representations not
only provide an interface for manipulating the image generation process
but also appear to be complementary to pretrained deep-neural-network
features for solving discriminative tasks.

\paragraph*{Acknowledgements}
This work was supported in part by ONR N00014-13-1-0762, NSF CAREER IIS-1453651, and Sloan Research Fellowship.

\begingroup

\small{}

\iftoggle{arxiv}{
\setstretch{1}
\setlength{\bibsep}{0pt plus 1pt}
\bibliographystyle{plainnat}
}{
\setstretch{0.95}
\setlength{\bibsep}{0pt plus 1pt}
\bibliographystyle{abbrvnat}
}

\bibliography{ref}

\endgroup
 
\clearpage

\iftoggle{arxiv}{  }{\pagestyle{plain}
\setcounter{page}{1}
}

\onecolumn

\appendix

\begin{center}
\textsf{\textbf{\Large{}Appendices}}
\par\end{center}

\vspace{0.5em}

\begin{center}
\textbf{\Large{}\mytitle}
\par\end{center}{\Large \par}

\vspace{1em}

\begin{center}
{\large{} \myauthor \par}
\vspace{4pt}
{\large{} \myaffliation \par}
\vspace{4pt}
{\small{} \texttt{\myemail}}
\vspace{0.5em}

\par\end{center}

\definecolor{filerefcolor}{RGB}{0, 150, 255}
\newcommand{\fileref}[1]{\textcolor{filerefcolor}{\texttt{#1}}}

\newcommand{\addstufftotoc}[2][toc]{  \addtocontents{#1}{#2}
}

\etocdepthtag.toc{mtappendix}
\etocsettagdepth{mtchapter}{none}
\etocsettagdepth{mtappendix}{subsection}

\vspace{1em}
\begin{center}

\iftoggle{arxiv}{
\begingroup
\large{}
The project web page (code and results): \url{http://ytzhang.net/projects/lmdis-rep} 
\endgroup
}{
\begingroup
\large{}
Code and results, including images and videos, are available at the project web page: 

\vspace{3pt}

\url{http://ytzhang.net/projects/lmdis-rep}
\endgroup
}

\vspace{1em}

\iftoggle{arxiv}{
Supplementary videos referred to in the appendices can be obtained at \linebreak
\url{http://ytzhang.net/files/lmdis-rep/supp-videos.tar.gz}
}{
Supplementary videos referred to in the appendices can be found in the zip file of the supplementary materials 
\linebreak
or obtained at \quad \url{http://ytzhang.net/files/lmdis-rep/supp-videos.tar.gz}
}
\end{center}
\vspace{2em}

\begingroup
\hypersetup{allcolors=black}

\renewcommand{\cftsecfont}{\large\mdseries}
\renewcommand{\cftsubsecfont}{\large\mdseries}
\renewcommand{\cftsecleader}{\large\mdseries\cftdotfill{\cftdotsep}}
\renewcommand{\cftsecpagefont}{\large\mdseries}
\renewcommand{\cftsubsecpagefont}{\large\mdseries}
\renewcommand{\cftsecafterpnum}{}
\setlength{\cftsecindent}{1 em}
\setlength{\cftsubsecindent}{2.5em}
\setlength{\cftbeforesecskip}{0.6em}
\setlength{\cftbeforesubsecskip}{0.3em}
\addtolength{\cftsecnumwidth}{6pt}
\addtolength{\cftsubsecnumwidth}{7pt}

\vspace{-1em}

\tableofcontents
\endgroup

\clearpage

\section{More details and results on face manipulation using unsupervised landmarks}
\label{supp:manipulation}

The discovered landmarks constitute the explicitly structural part of the image representation learned by our model. 
They provide an interface for humans to manipulate the image representation intuitively. 
Our decoding module can generate realistic facial images using the landmark descriptors extracted from a given image and different sets of landmarks. 

In addition to the results shown in the main paper, we provide more qualitative results for unsupervised landmark-based face manipulation in this section. 
We train models of 10, 20, and 30 landmarks. 
To evaluate our method on many target landmarks, we take the landmarks discovered from other images and the interpolation/extrapolation between the landmarks discovered on two images as the targets. 

In this section, we show results for our 30-landmark model (results for our 10,20-landmark model are available as supplementary videos). 
Figure~\ref{fig:face-morph-all} and \ref{fig:face-morph-all-2} show results for manipulating all 30 landmarks. 
Figure~\ref{fig:face-morph-mouth} and \ref{fig:face-morph-mouth-2} show results for manipulating the 3 landmarks at the mouth. 
Figure~\ref{fig:face-morph-mouthext} and \ref{fig:face-morph-mouthext-2} show results for manipulating the 5 landmarks at the mouth and jaw. 
Videos are available in the following folders for gradually morphing the landmarks from their original coordinates to the target by linear interpolation.  
\begin{itemize}
\item 
\begin{flushleft}
30-landmark models: \linebreak \fileref{\texttt{videos/face.30landmark-model.manipulate-\{all|mouth|mouthext\}-landmarks}} 
\end{flushleft}
\item 
\begin{flushleft}
10,20-landmark models: \linebreak \fileref{\texttt{videos/face.\{10|20\}landmark-model.manipulate-\{all|mouth\}-landmarks}} 
\end{flushleft}
\end{itemize}

\vfill{}
\begingroup
\begin{center}
\Large{} See next page for the figure.
\end{center}
\endgroup
\vfill{}

\clearpage

\newcommand{\insertmanvid}[2]{
\vspace{-10pt}\subfloat[Video:  \fileref{\texttt{videos/#1/#2.mp4}}]{\includegraphics[width=0.8\textwidth]{figures.reduced/#1/#2.pdf}}\par
}

\newcommand{\insertmancaptext}[2]{
#2. \textbf{1st column}: input images; 
\textbf{2nd column}: discovered landmarks and reconstructed images; 
\textbf{other columns}: the red dots denote the target landmark locations (gray dots means not too much offset regarding the original landmarks), the gray lines denote the synthetic adjustment of landmarks, and the facial images are the decoder outputs. 
Best viewed in zoom mode. Videos are available at \fileref{\texttt{videos/#1}} for the morphing process.
}

\begin{figure}[H]

\begin{center}

\insertmanvid{face.30landmark-model.manipulate-all-landmarks}{01}
\insertmanvid{face.30landmark-model.manipulate-all-landmarks}{02}
\insertmanvid{face.30landmark-model.manipulate-all-landmarks}{03}
\insertmanvid{face.30landmark-model.manipulate-all-landmarks}{04}
\insertmanvid{face.30landmark-model.manipulate-all-landmarks}{06}

\end{center}

\caption{
\label{fig:face-morph-all}
\insertmancaptext{face.30landmark-model.manipulate-all-landmarks}{Face manipulation by modifying all 30 discovered landmarks on the MAFL testing set}}

\end{figure}

\newpage

\begin{figure}[H]

\begin{center}

\insertmanvid{face.30landmark-model.manipulate-all-landmarks}{06}
\insertmanvid{face.30landmark-model.manipulate-all-landmarks}{07}
\insertmanvid{face.30landmark-model.manipulate-all-landmarks}{08}
\insertmanvid{face.30landmark-model.manipulate-all-landmarks}{09}
\insertmanvid{face.30landmark-model.manipulate-all-landmarks}{10}

\end{center}

\caption{
\label{fig:face-morph-all-2}
\insertmancaptext{face.30landmark-model.manipulate-all-landmarks}{Continued from Figure~\ref{fig:face-morph-all}. Face manipulation by modifying all 30 discovered landmarks on the MAFL testing set}}
\end{figure}

\newpage

\begin{figure}[p]

\begin{center}

\insertmanvid{face.30landmark-model.manipulate-mouth-landmarks}{01}
\insertmanvid{face.30landmark-model.manipulate-mouth-landmarks}{02}
\insertmanvid{face.30landmark-model.manipulate-mouth-landmarks}{03}
\insertmanvid{face.30landmark-model.manipulate-mouth-landmarks}{04}
\insertmanvid{face.30landmark-model.manipulate-mouth-landmarks}{05}

\end{center}

\caption{
\label{fig:face-morph-mouth}
\insertmancaptext{face.30landmark-model.manipulate-mouth-landmarks}{Face manipulation by modifying 3 discovered mouth landmarks on the MAFL testing set}}
\end{figure}

\clearpage

\begin{figure}[p]

\begin{center}

\insertmanvid{face.30landmark-model.manipulate-mouth-landmarks}{06}
\insertmanvid{face.30landmark-model.manipulate-mouth-landmarks}{07}
\insertmanvid{face.30landmark-model.manipulate-mouth-landmarks}{08}
\insertmanvid{face.30landmark-model.manipulate-mouth-landmarks}{09}
\insertmanvid{face.30landmark-model.manipulate-mouth-landmarks}{10}

\end{center}

\caption{
\label{fig:face-morph-mouth-2}
\insertmancaptext{face-manipulation-mouth-landmarks}{Continued from Figure~\ref{fig:face-morph-mouth}. Face manipulation by modifying mouth 3 discovered mouth landmarks on the MAFL testing set}}
\end{figure}

\clearpage

\begin{figure}[p]

\begin{center}

\insertmanvid{face.30landmark-model.manipulate-mouthext-landmarks}{01}
\insertmanvid{face.30landmark-model.manipulate-mouthext-landmarks}{02}
\insertmanvid{face.30landmark-model.manipulate-mouthext-landmarks}{03}
\insertmanvid{face.30landmark-model.manipulate-mouthext-landmarks}{04}
\insertmanvid{face.30landmark-model.manipulate-mouthext-landmarks}{05}

\end{center}

\caption{
\label{fig:face-morph-mouthext}
\insertmancaptext{face.30landmark-model.manipulate-mouthext-landmarks}{Face manipulation by modifying 6 discovered mouth and jaw landmarks on the MAFL testing set}}
\end{figure}

\clearpage

\begin{figure}[p]

\begin{center}

\insertmanvid{face.30landmark-model.manipulate-mouthext-landmarks}{06}
\insertmanvid{face.30landmark-model.manipulate-mouthext-landmarks}{07}
\insertmanvid{face.30landmark-model.manipulate-mouthext-landmarks}{08}
\insertmanvid{face.30landmark-model.manipulate-mouthext-landmarks}{09}
\insertmanvid{face.30landmark-model.manipulate-mouthext-landmarks}{10}

\end{center}

\caption{
\label{fig:face-morph-mouthext-2}
\insertmancaptext{face-manipulation-mouthext-landmarks}{Continued from Figure~\ref{fig:face-morph-mouth}. Face manipulation by modifying mouth 6 discovered mouth and jaw landmarks on the MAFL testing set}}
\end{figure}

\clearpage

\section{Our model without landmark descriptors on MNIST}
\label{supp:mnist}
We train our landmark discovery model without the landmark descriptor pathway on MNIST in two settings: one model for each digit (\supp{\ref{supp:mnist-individual}}) and one model for all ten digits (\supp{\ref{supp:mnist-all}}). 
Using the model for all digits, we can perform geometrically meaningful morphing between different digits (\supp{\ref{supp:mnist-morph}}), e.g., morphing 2 to 9. 
Videos for the morphing process are available in the folder \fileref{\texttt{videos/mnist-morphing}} . 

\subsection{Models for individual digits}
\label{supp:mnist-individual}
\begin{figure}[H]
\centering
  \begin{tabular}{c}
    \includegraphics[width=0.8\textwidth]{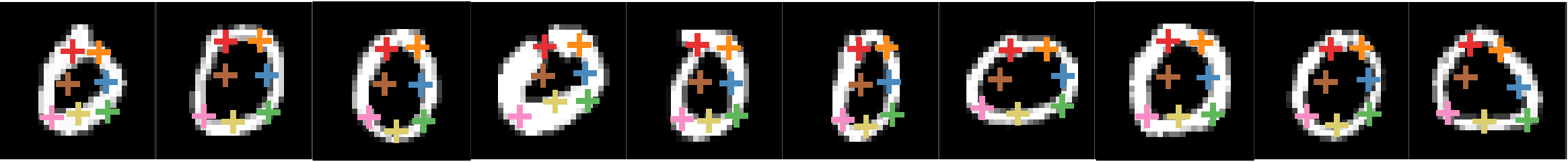} \tabularnewline
    \includegraphics[width=0.8\textwidth]{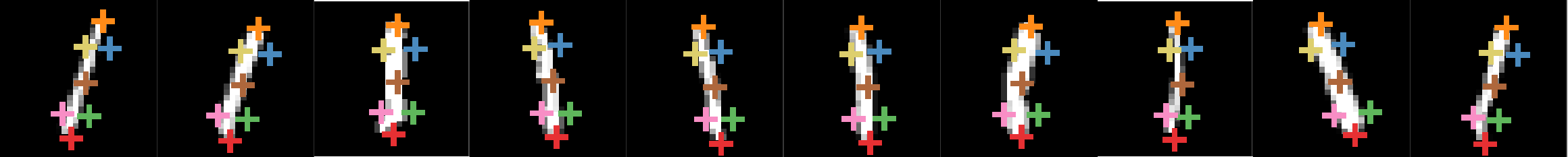} \tabularnewline
    \includegraphics[width=0.8\textwidth]{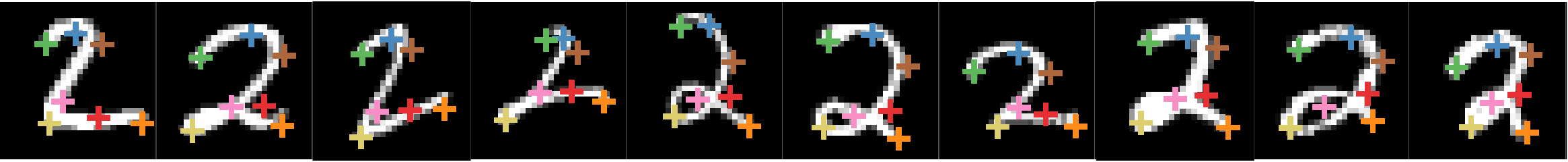} \tabularnewline 
    \includegraphics[width=0.8\textwidth]{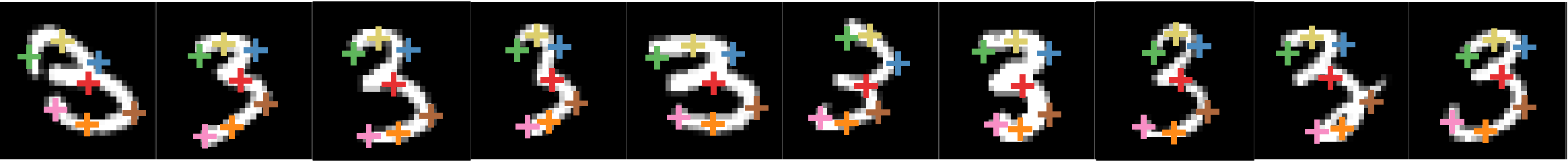} \tabularnewline 
    \includegraphics[width=0.8\textwidth]{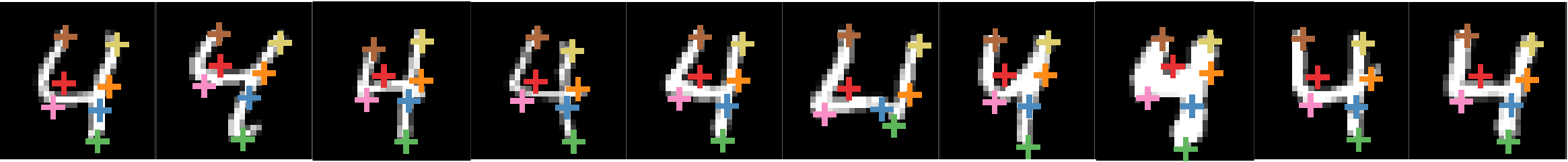} \tabularnewline 
    \includegraphics[width=0.8\textwidth]{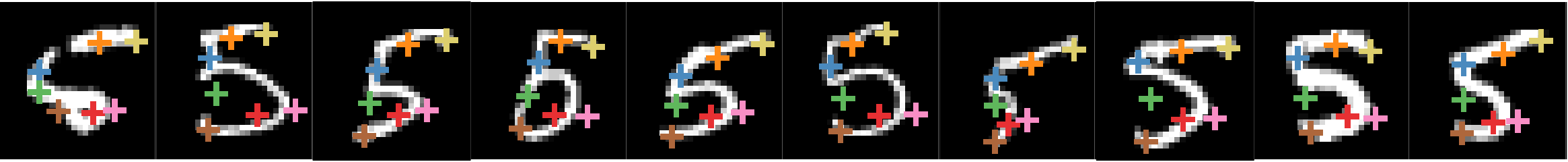} \tabularnewline 
    \includegraphics[width=0.8\textwidth]{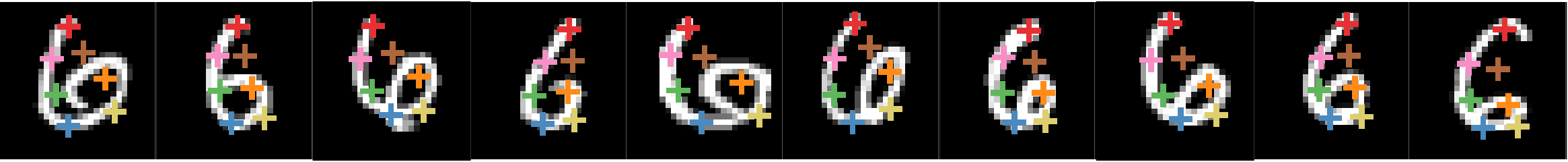} \tabularnewline 
    \includegraphics[width=0.8\textwidth]{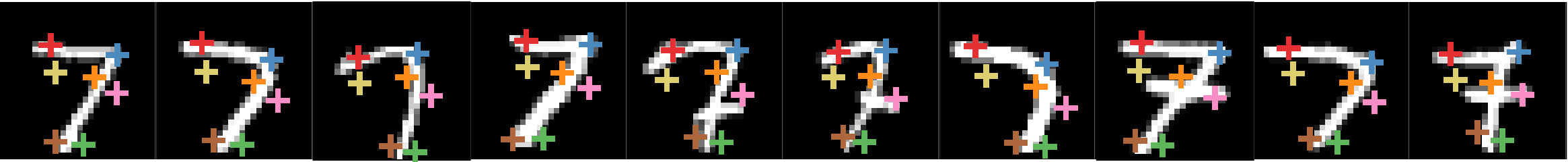} \tabularnewline 
    \includegraphics[width=0.8\textwidth]{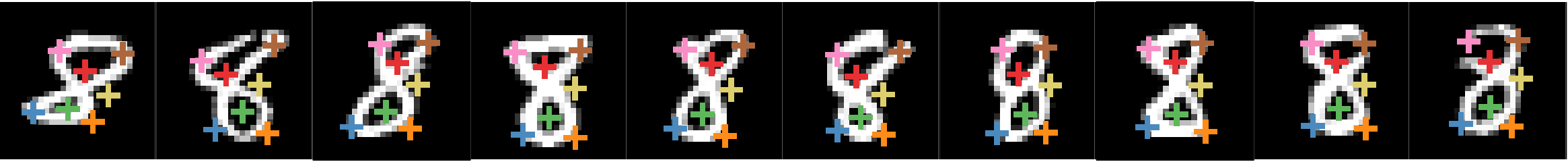} \tabularnewline 
    \includegraphics[width=0.8\textwidth]{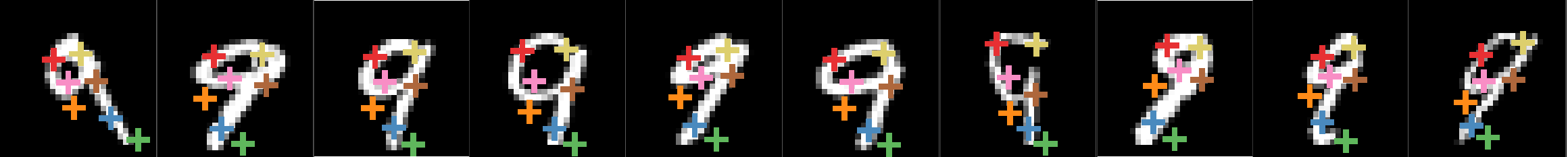} \tabularnewline 
    
  \end{tabular}
  \caption{\label{fig:sep-mnist}Discovering 7 landmarks on MNIST. Our model is trained independently for each digit.}
\end{figure}

In this section, our landmark discovery model is trained for each digit independently. As shown in Figure~\ref{fig:sep-mnist}, the discovered landmarks are consistent within each digit despite the shape variations.

\vfill

\vfill

\clearpage

\subsection{Models for all digits}
\label{supp:mnist-all}
\begin{figure}[H]
\begin{centering}
\includegraphics[width=0.8\textwidth,height=0.8\textheight,keepaspectratio]{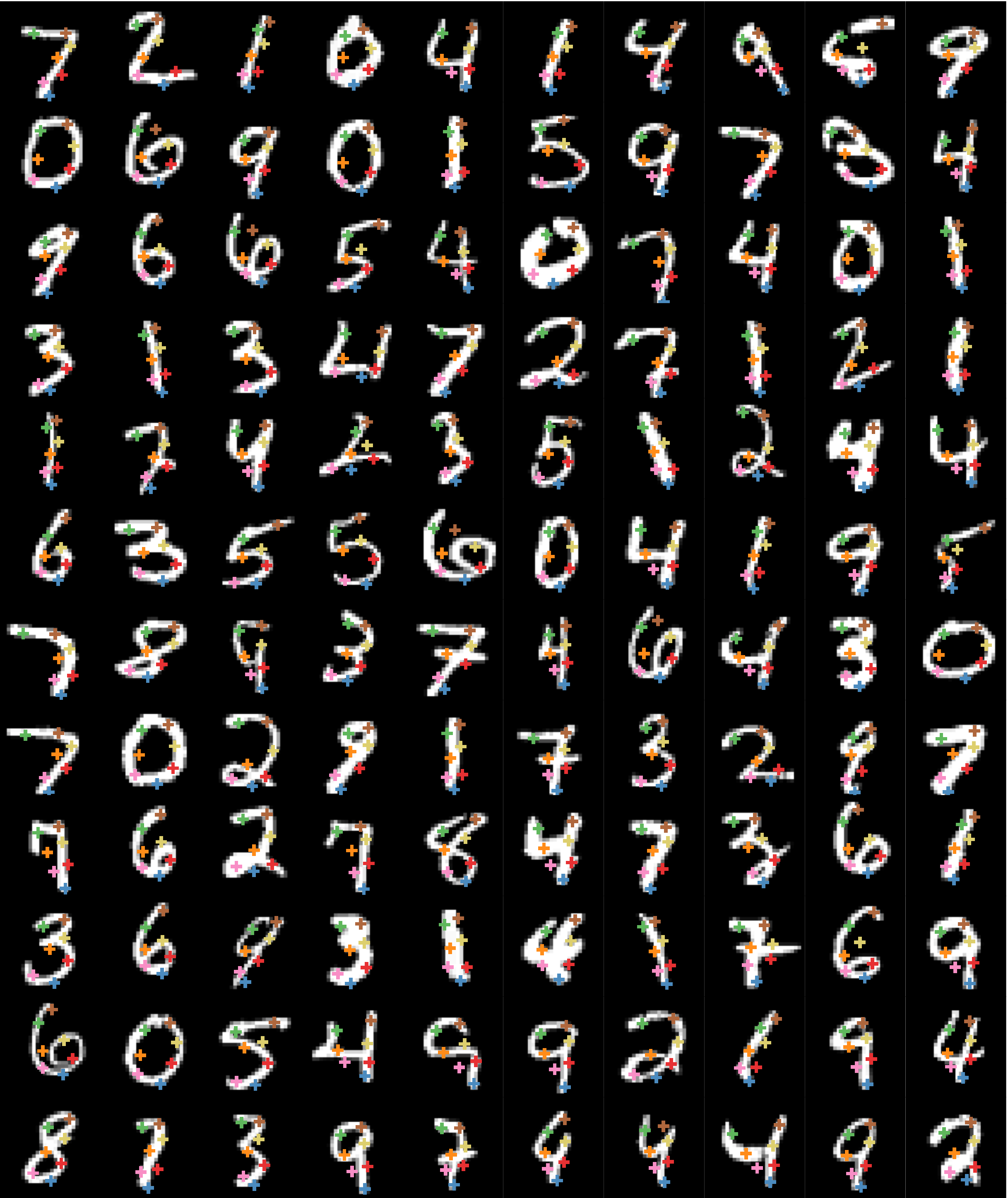}
\par\end{centering}
\caption{\label{fig:all-digits}Discovering 7 landmarks on MNIST for all digits. A single model is trained for all ten digits.  The corresponding landmarks for different images are in the same color.}
\end{figure}

In this section, we train a single model for all ten digits together. 
In Figure~\ref{fig:all-digits}, the corresponding landmarks are shown in the same color. 
The landmarks discovered on the same digits are consistent across different image. 
\textbf{More interesting, the corresponding landmarks across different digits are also semantically consistent.} 
For examples, the orange cross is always in the middle of a digits, the blue cross at the bottom, and the green one at the most left-top part of every digit.

\vfill

\vfill

\clearpage

\subsection{Digit morphing using discovered landmarks}
\label{supp:mnist-morph}

Our model trained on the mix of all ten digits can discover corresponding patterns among different digit categories. 
Using this model, we can perform geometrically meaningful cross-category morphing. 
Figure~\ref{fig:digit-morph} and \ref{fig:digit-morph-2} illustrate the morphing process. 
Note that the landmark coordinates constitute the full image representation for MNIST digits.    
Videos for the morphing process are available in the folder \fileref{\texttt{videos/mnist-morphing}} . 

\newcommand{\insertmnistmorphcaptext}{
Geometric digital morphing using our discovered landmarks and image decoding module.
The landmark coordinates are linearly interpolated between the source digit and the target digit.
\textbf{1st column}: input images; 
\textbf{2nd column}: discovered landmarks and reconstructed images of the source digit; 
\textbf{last column}: discovered landmarks and reconstructed images of the target digit; 
\textbf{other columns}: the red dots denote the target landmark locations (gray dots means not too much offset regarding the original landmarks), the gray lines denote the synthetic adjustment of landmarks, and the facial images are the decoder outputs. 
Best viewed in zoom mode. Videos are available at \fileref{\texttt{videos/mnist-morphing}} for the morphing process.
}

\vfill

\begin{figure}[H]
\begin{centering}
\includegraphics[width=\textwidth,height=0.8\textheight,keepaspectratio]{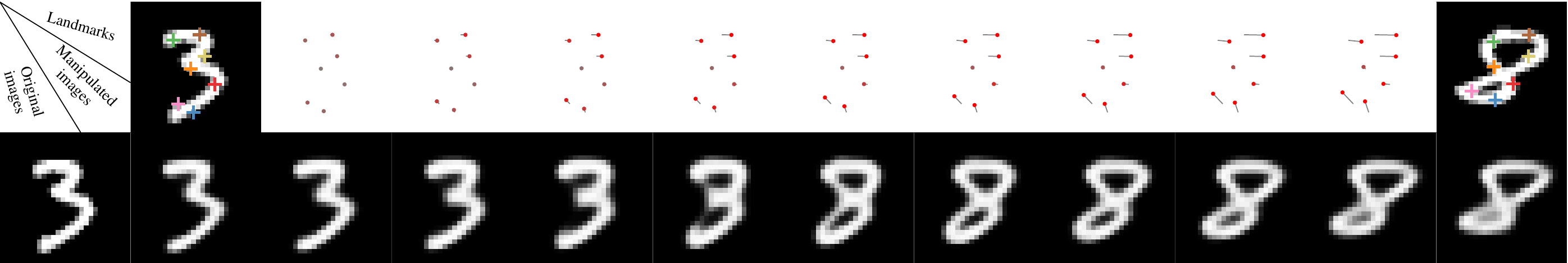}
\includegraphics[width=\textwidth,height=0.8\textheight,keepaspectratio]{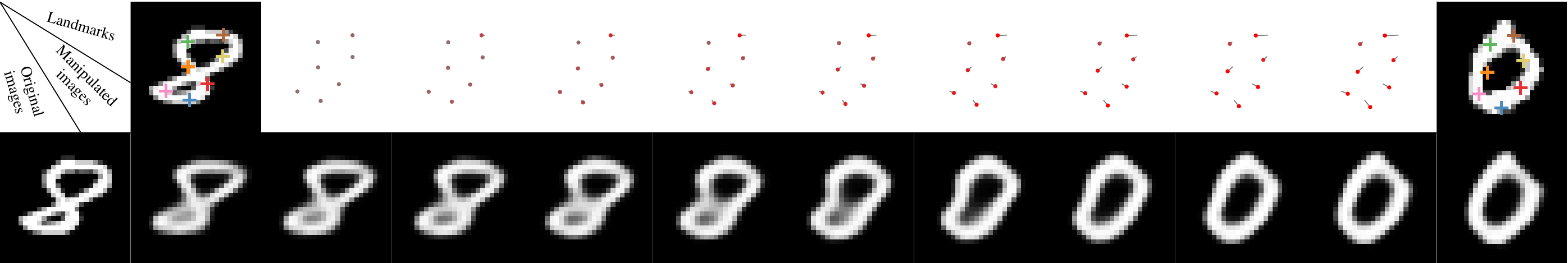}
\includegraphics[width=\textwidth,height=0.8\textheight,keepaspectratio]{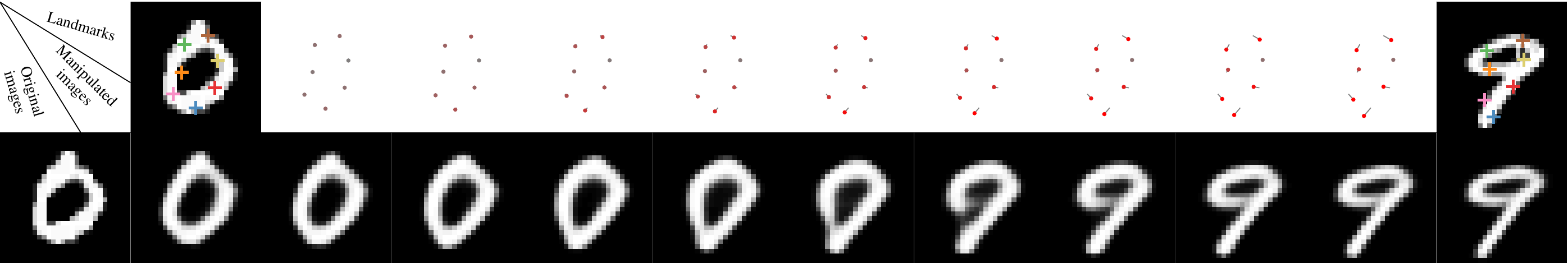}
\includegraphics[width=\textwidth,height=0.8\textheight,keepaspectratio]{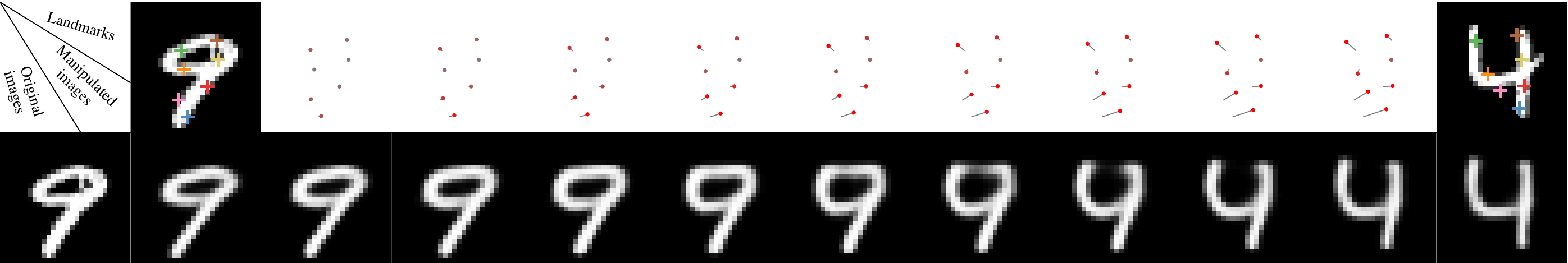}
\par\end{centering}
\caption{\label{fig:digit-morph}\insertmnistmorphcaptext}
\end{figure}

\vfill

\clearpage

\begin{figure}[p]
\begin{centering}
\includegraphics[width=\textwidth,height=0.8\textheight,keepaspectratio]{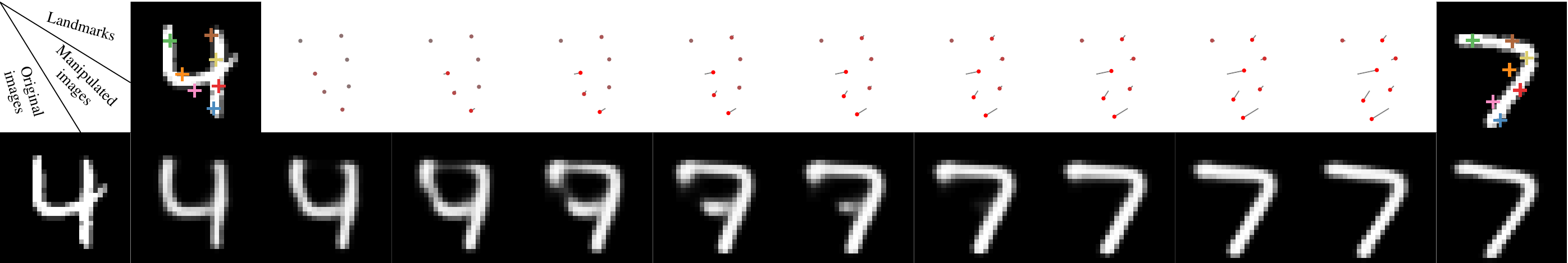}
\includegraphics[width=\textwidth,height=0.8\textheight,keepaspectratio]{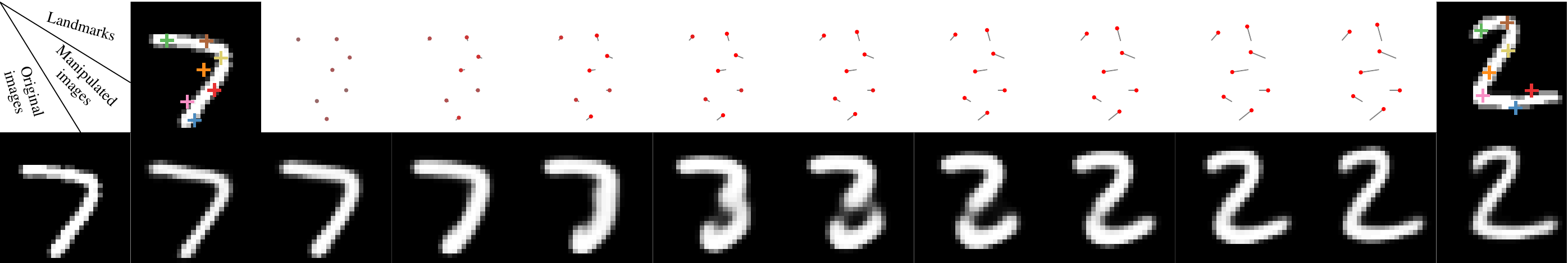}
\includegraphics[width=\textwidth,height=0.8\textheight,keepaspectratio]{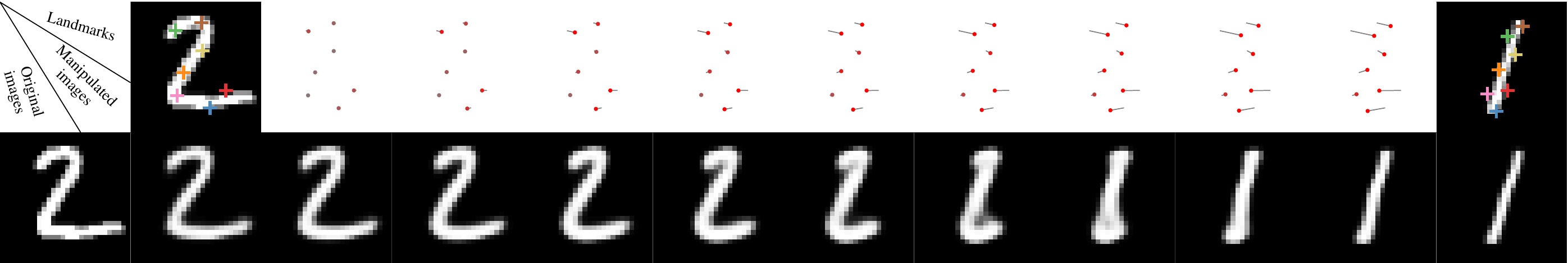}
\includegraphics[width=\textwidth,height=0.8\textheight,keepaspectratio]{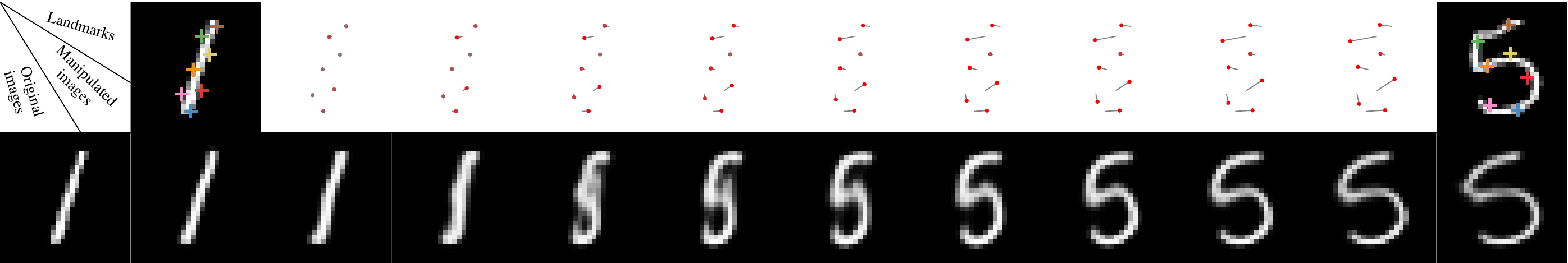}
\includegraphics[width=\textwidth,height=0.8\textheight,keepaspectratio]{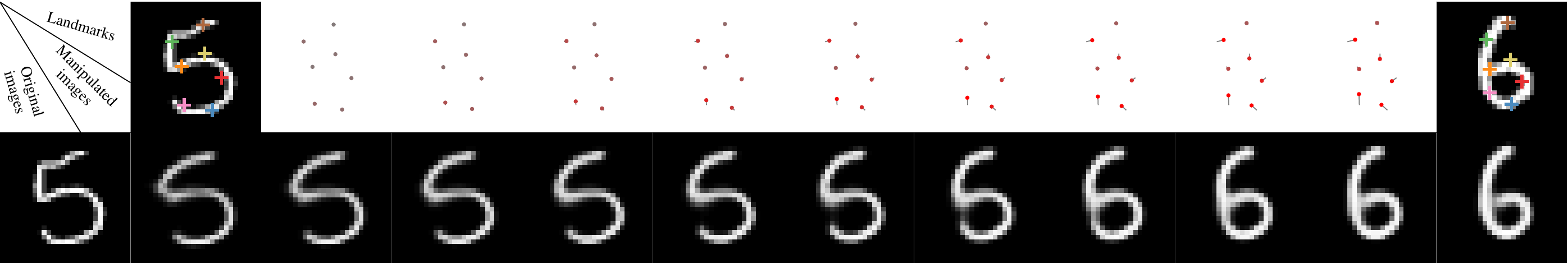}
\par\end{centering}
\caption{\label{fig:digit-morph-2}Continued from Figure~\ref{fig:digit-morph}. \insertmnistmorphcaptext}
\end{figure}

\clearpage

\section{Details and more results on Human3.6M}
\label{supp:human}
On the Human3.6M dataset, we train our landmark discovery model on six actions: waiting, posing, greeting, directions, discussions, and walking. 
We report quantitative results on predicting the 32 annotated landmarks\footnote{Some markers are close to each other (e.g., on each foot, there are two markers), so the effective locations annotated by the markers are less than 32 (around 16).} (acquired by wearable markers) using our models trained for the mix of all six actions and each independent action.
We also show qualitative results of the mixed-action model (trained for all six actions). 

\subsection{Optical flow as self-supervision for equivariance}

The Human3.6M training data is in the video format. 
We can calculate the optical flows between nearby frames and take them as self-supervision for the equivariance constraint defined in \eqref{eq:eqv-loss}.
Following the same notations, the two frames are $\mathbf{I}$ and  $\mathbf{I}'$, and the optical flows define the transformation $g(\cdot, \cdot)$. 
In particular, we use the Farneback method in OpenCV to compute the dense optical flows at 5-frame intervals.
We then accumulate two optical flow fields to calculate the optical flows at 10-frame intervals.
The 10-frame-interval optical flow fields in addition to the random TPS transform are used in the equivariance constraint.

Note that using optical flow as the self-supervision for the equivariance constraint can push the landmarks to the background region.
This phenomenon occurs because locating landmarks at image regions with weaker optical flows can result in low equivariance loss. 
To prevent trivial landmarks, we encourage the landmarks to be discovered at locations with strong optical flows. 
Let  $\mathbf{O}_{\mathrm{x}}$ and $\mathbf{O}_{\mathrm{y}}$ be the $x,y$ components of the optical flow map from the current image to another frame. 
The flow magnitude map is $\mathbf{O}_{\mathrm{n}}(u,v)=\left(\mathbf{O}_{\mathrm{x}}(u,v)\right)^{2}+\left(\mathbf{O}_{\mathrm{y}}(u,v)\right)^{2}$.
Recall that $\tilde{\mathbf{R}}$ is the multi-channel heatmap computed from the landmark locations. 
We encourage $\mathbf{O}_{\mathrm{n}}$ and $\tilde{\mathbf{R}}$ to have a significant correlation.
In particular, loss to encourage 
\begin{equation}
L_{\mathrm{flow\text{-}prefer}}=-\frac{\sum_{v=1}^{H}\sum_{u=1}^{W}\mathbf{O}_{\mathrm{n}}(u,v)\sum_{k=1}^{K}\tilde{\mathbf{R}}_{k}(u,v)}{\sum_{v=1}^{H}\sum_{u=1}^{W}\mathbf{O}_{\mathrm{n}}(u,v)} \label{eq:loss-flow-prefer}
\end{equation}
We use the same loss weight $\lambda_{\mathrm{eqv}}$ for $L_{\mathrm{flow\text{-}prefer}}$ as $L_{\mathrm{eqv}}$. 

Using the above formulation, we can reduce the ground landmark prediction error from $4.91$ (without optical flows) to $4.14$ (with optical flows). 
For all experiments on Human3.6M, we use the optical flow as self-supervision for equivariance as described above.

\subsection{Quantitative results}

We compare our model with \citet{unsupervised-landmark}'s unsupervised landmark discovery method regarding the annotated-landmark prediction accuracy. 
Both models discover 16 landmarks. The whole training set is used to train the linear mapping from the discovered landmarks to the annotated ones. 
As discussed in the main paper, we do not expect the two unsupervised methods to distinguish the frontal and back views. 
Thus, in the evaluation, we compute the errors against the original landmark annotations and its left-right-flipped counterpart\footnote{For examples, we swap the coordinates of the left-shoulder landmark and the right-shoulder landmark. }, and then we choose the minimum value as the final error. 
Note that, when flipping the landmark annotations, the landmarks for the whole body are flipped simultaneously. 
As to the linear regressor training, we propose the following training strategy. 
\begin{enumerate}
\item Figure out the rough orientations of the human body, heuristically. If more than 2/3 of the left-hand side annotated landmarks are to the right of the right-hand side annotated landmarks, the human is in the frontal view.
\item Train the regressor using the images with the landmark annotations in the frontal view. The other images are ignored in this step.
\item 
Use the aforementioned evaluation protocol to determine if the landmark annotations on other images should be flipped or not. The model is then retrained with all the training images.
\item Repeat the step 3 until the model is converged.
\end{enumerate}

As shown in Table~\ref{tab:human-quan}, our method outperforms \citet{unsupervised-landmark}'s method significantly. 
We also report the results obtained by \citet{hourglass}'s supervised stacked hourglass network using their off-the-shelf pretrained 16-landmark model. 
Both unsupervised methods perform worse than the supervised stacked hourglass network. 
However, our model is unsupervised, and our neural network architectures are also smaller. 
We believe that our results show the potential of unsupervised methods for discovering complicated object structures.

\begin{table*}[h]
\begingroup
\setlength\tabcolsep{2pt}
\begin{centering}
\begin{tabular}{c|c|c|cccccc}
\hline 
\multicolumn{2}{c|}{Methods} & \footnotesize{}\begin{tabular}{@{}l@{}}
Mixed\tabularnewline
Actions\tabularnewline
\end{tabular} & 
Waiting & Posing  & Greeting & Directions & Discussion & Walking \tabularnewline
\hline 
Unsupervised & \citet{unsupervised-landmark} & 7.51  & 7.54  & 8.56  & 7.26 & 6.47  & 7.93  &  5.40 \tabularnewline
 & Ours (discovered landmarks) & \textbf{4.14}  & \textbf{5.01}
  & \textbf{4.61}  & \textbf{4.76} & \textbf{4.45} & \textbf{4.91}  & \textbf{4.61} \tabularnewline \hline
 Supervised & Hourglass \citet{hourglass} & \textbf{2.16} & \textbf{1.88 } & \textbf{1.92} & \textbf{2.15} &  \textbf{1.62} &  \textbf{1.88} &  \textbf{2.21} \tabularnewline
\hline 
\end{tabular}
\par\end{centering}
\endgroup
\centering{}\caption{\label{tab:human-quan}Comparison with unsupervised and supervised methods for annotated landmark prediction on the Human 3.6M testing sets. The error is in \% regarding the edge length of the image.}
\vspace*{0.05in}
\end{table*}

\vfill
 
\subsection{Qualitative results}

\begin{figure}[H]
\centering
  \begin{tabular}{c}
   \includegraphics[width=\textwidth,height=0.9\textheight,keepaspectratio]{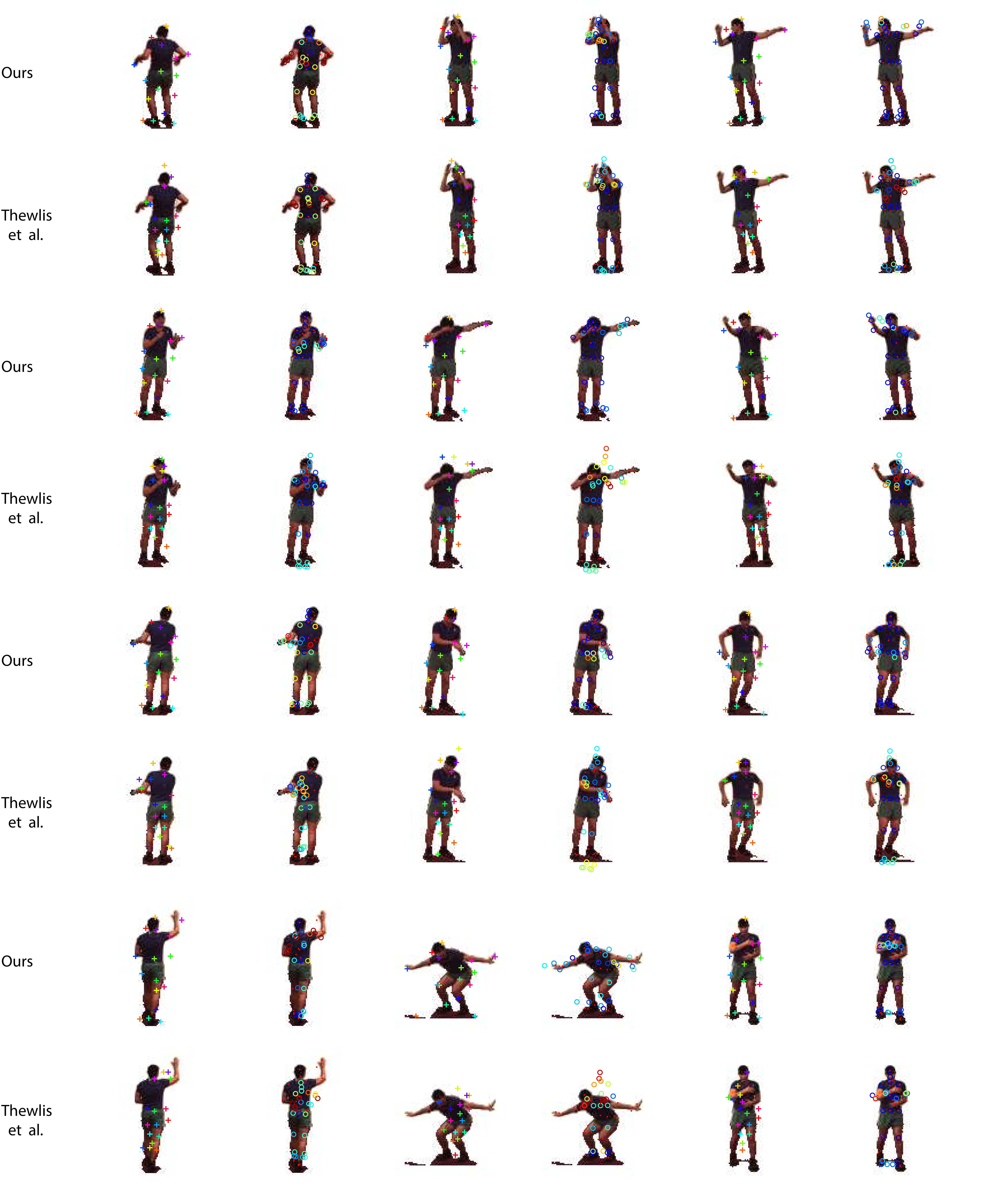} \tabularnewline
    \includegraphics[width=0.8\textwidth,height=0.8cm,keepaspectratio]{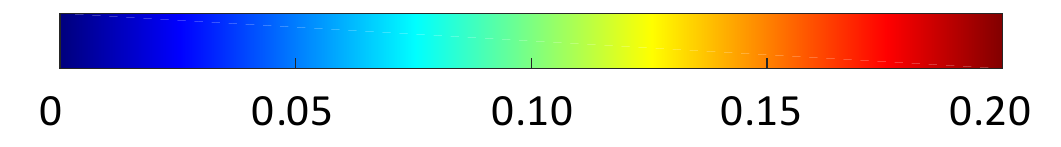} 
  \end{tabular}
  \caption{\label{fig:human-comp}Prediction of 32 annotated landmarks on Human 3.6M. Colorful cross: discovered landmark; Red dot: annotated landmark; Circle: regressed landmark, whose color represent its distance to the annotated land-marks. See the color bar for the distance (i.e., prediction error). Our method shows more deep blue circles (for example, the image in second row second column, the image in last row second column), which means more landmarks with low error compared with \citet{unsupervised-landmark}}
\end{figure}

\begin{figure}[p]
   \includegraphics[width=\textwidth,height=0.9\textheight,keepaspectratio]{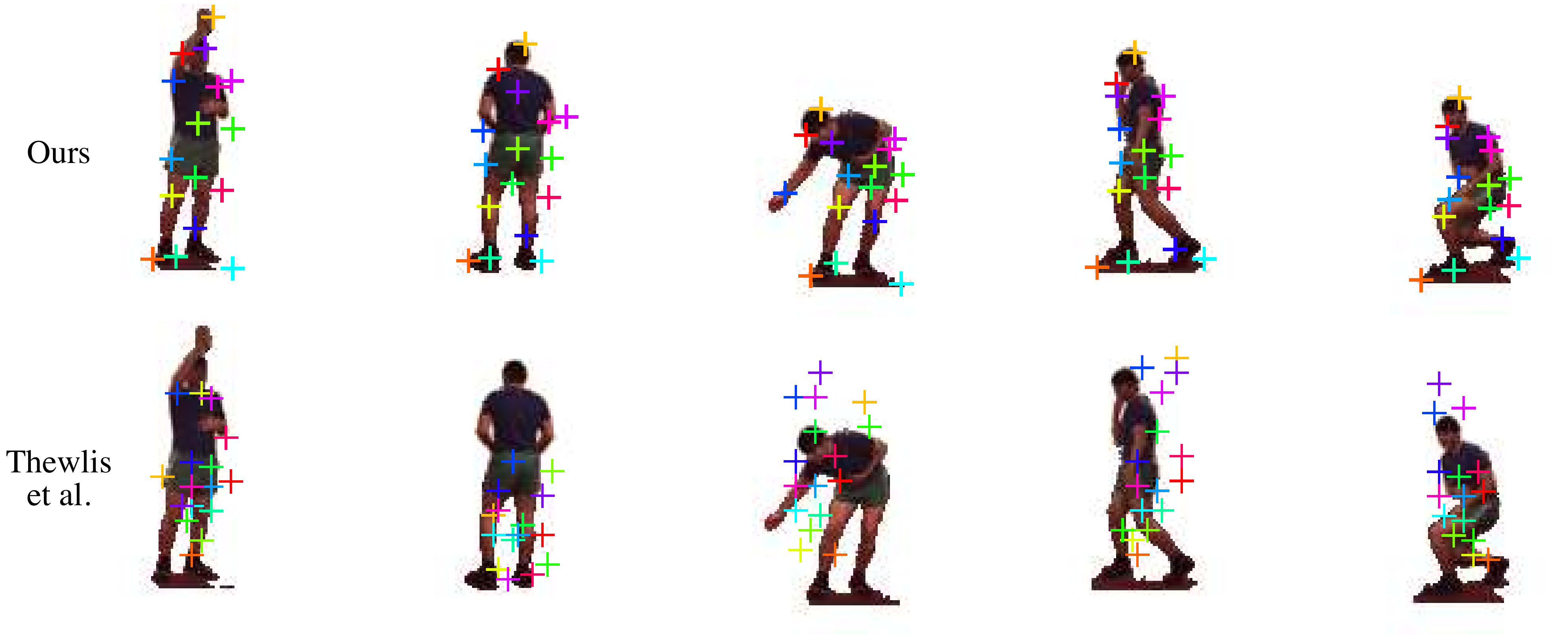}
   \includegraphics[width=\textwidth,height=0.9\textheight,keepaspectratio]{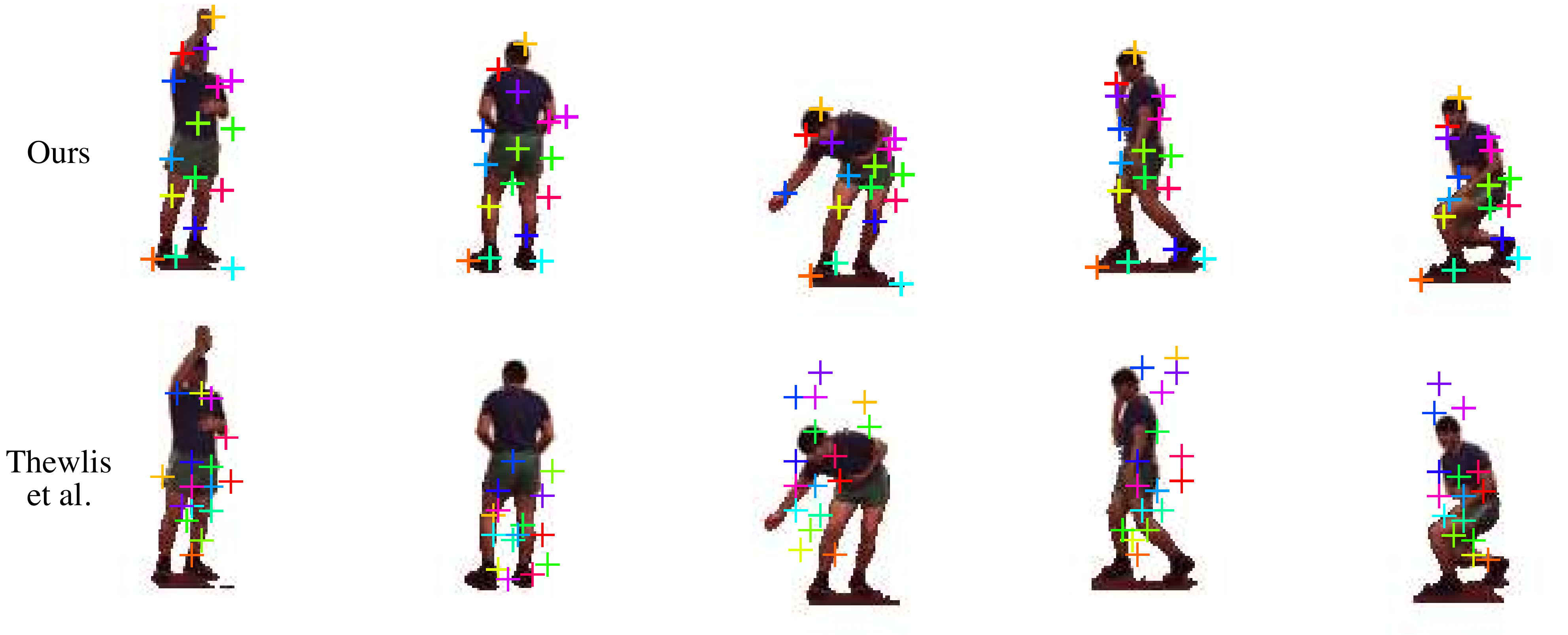} 
  \caption{\label{fig:human-lm-discovery} Discovering 16 landmarks on Human3.6M testing set. We illustrate some cases when our method outperforms \citet{unsupervised-landmark}'s very significantly.}
\end{figure}

We train our model and  \citet{unsupervised-landmark}'s model on all six chosen actions and perform the annotated-landmark prediction. 
Figure~\ref{fig:human-comp} shows the side-by-side comparison. In general, our method visually outperforms  \citet{unsupervised-landmark}'s.
Figure~\ref{fig:human-lm-discovery} shows landmark discovery examples, where our method outperforms \citet{unsupervised-landmark}'s method very significantly.

\vfill{}

\begingroup
\begin{center}
\Large{} See next page for the figure.
\end{center}
\endgroup

\vfill{}

\newpage

\pagebreak

\clearpage
 
\clearpage
\section{Results on animals of mixed species}
\label{supp:animals}

\begin{figure}[H]
\begin{centering}
\includegraphics[width=\textwidth,height=0.9\textheight,keepaspectratio]{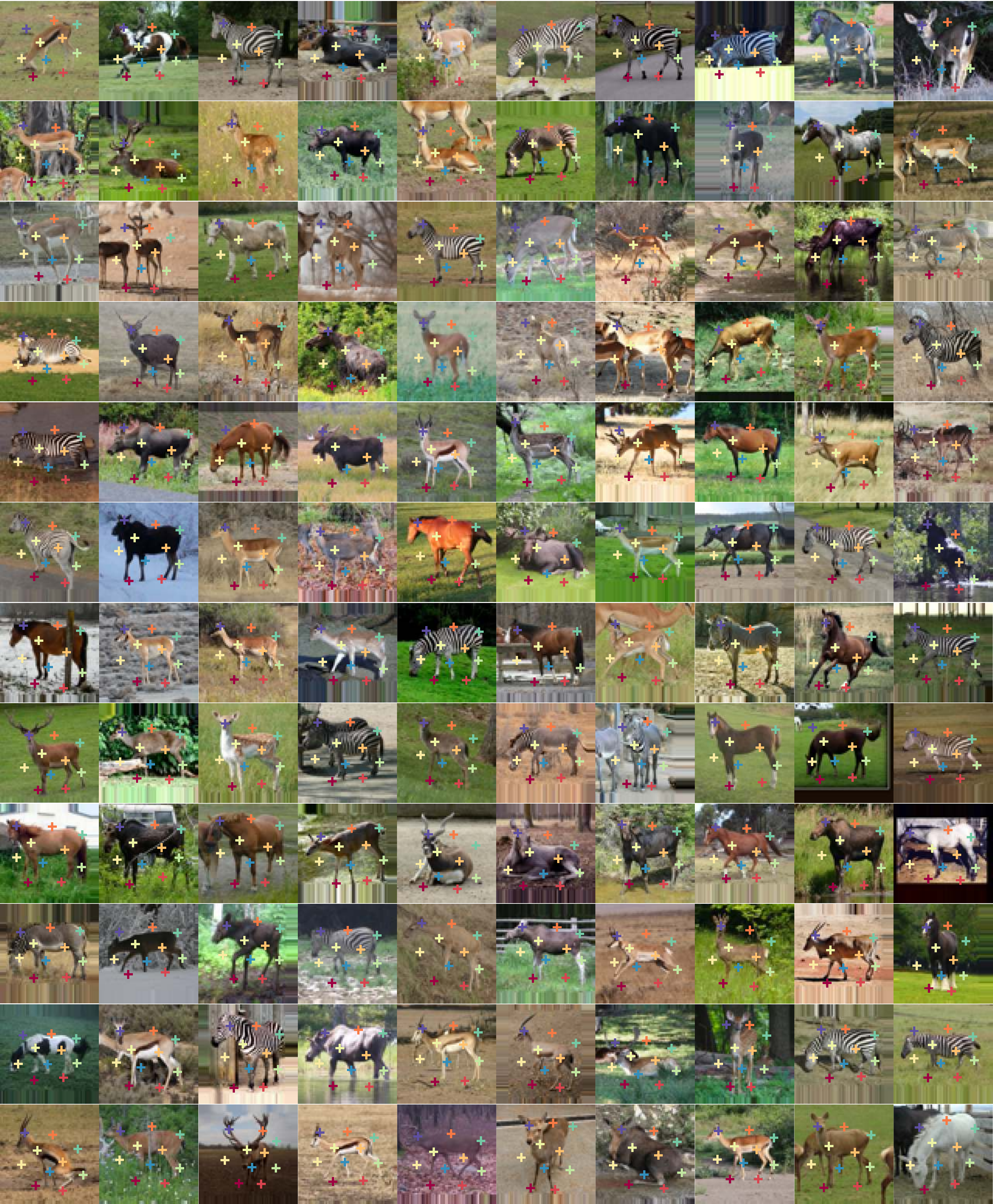}
\par\end{centering}
\caption{\label{fig:awa}Discovering 10 landmarks on mixed animal images of five different species: antelope, deer, moose, horse, zebra }
\end{figure}

On the animal-with-attributes (AwA) dataset~\citep{awa}, we choose the profile images from five animal categories (antelope, deer, moose, horse, zebra) and try to detect landmarks on these mixed species of animals. 
As shown in Figure~\ref{fig:awa}, even though multiple species of animals with different appearance are mixed, our method can still find several consistent landmarks. 
For example, the yellow cross is always on the hoof, the orange cross always above the back and the light green cross at the buttock. 
The landmarks are consistently detected despite the significant variations in species, pose and individual appearance.

\vfill{}

\begingroup
\begin{center}
\Large{} See next page for the figure.
\end{center}
\endgroup

\vfill{}

\newpage

\newpage
 
\clearpage
\section{More qualitative results on human faces, cat heads, cars, and shoes}
\label{supp:more-qual}
In this section, we show more result of landmark discovery and ground truth landmark prediction compared with  \citet{unsupervised-landmark}. All the shown images are randomly sampled from the test set.

\subsection{CelebA}
\label{supp:celeba}

\begin{figure}[H]
\begin{centering}
\includegraphics[width=\textwidth,height=0.8\textheight,keepaspectratio]{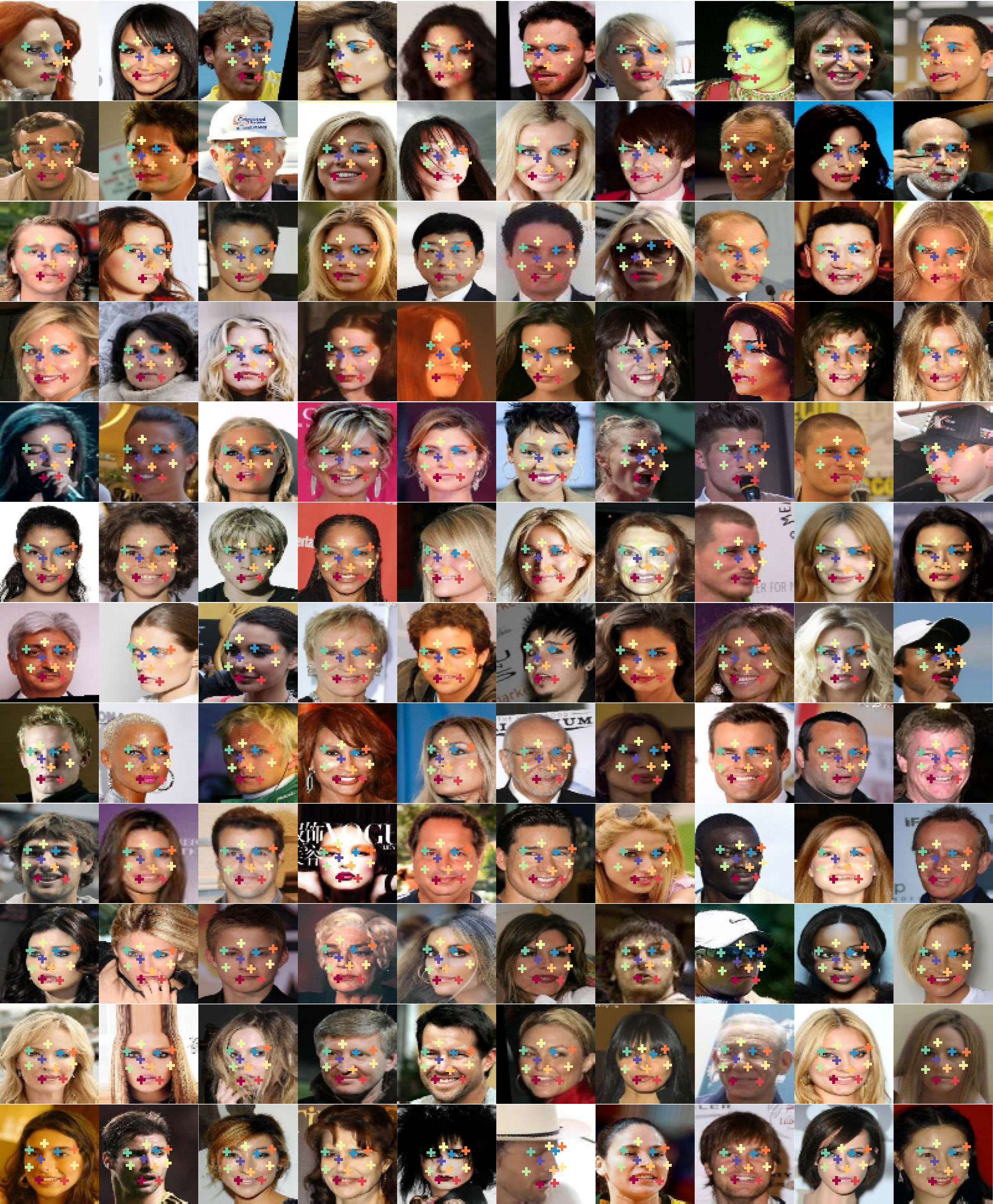}
\par\end{centering}
\caption{\label{fig:celeba-more-10}Discovering 10 landmarks on CelebA, the detected landmarks are highly aligned with facial features such as mouth corner, eyes corner and nose}
\end{figure}

\begin{figure}[H]
\begin{centering}
\includegraphics[width=\textwidth,height=0.9\textheight,keepaspectratio]{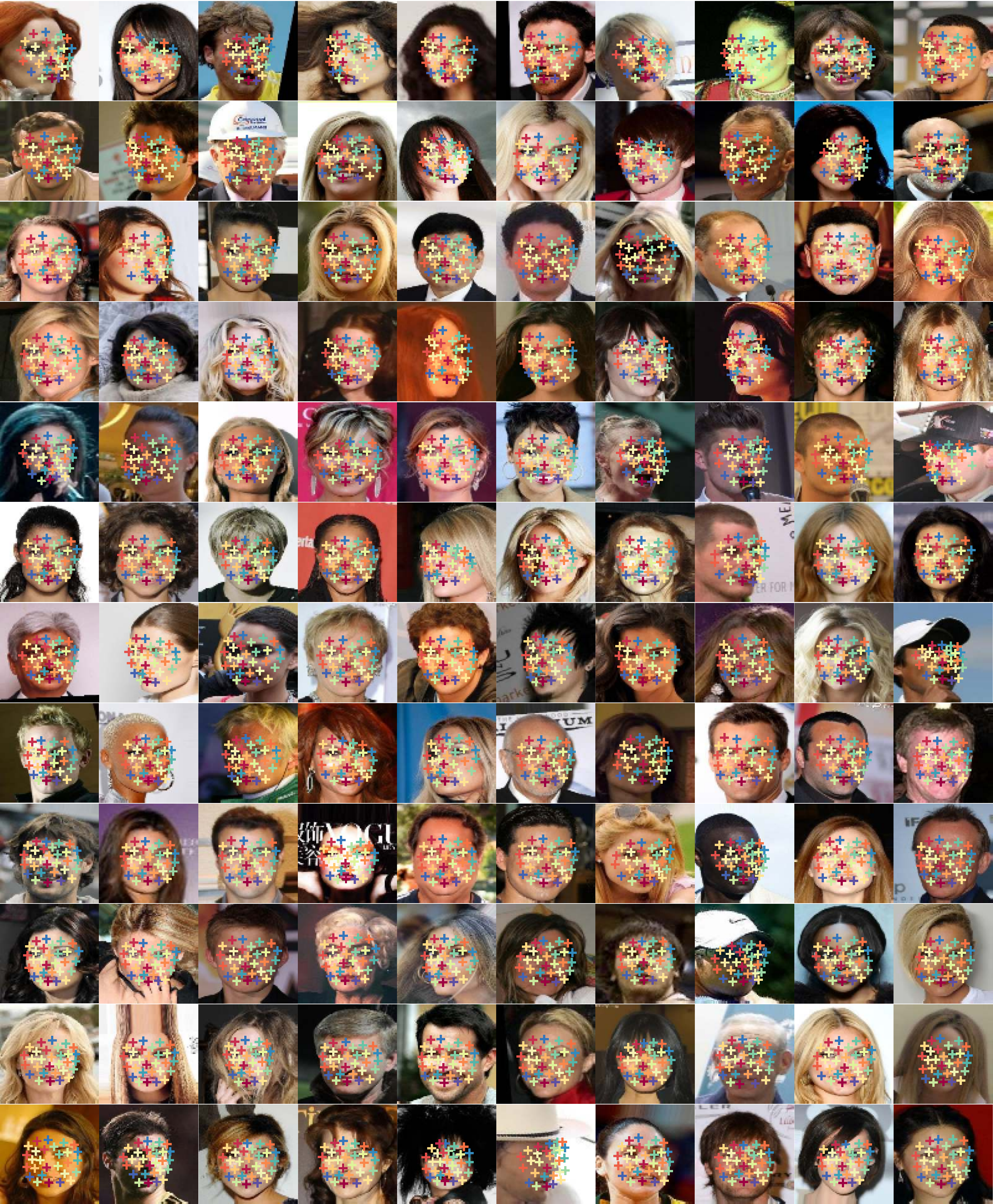}
\par\end{centering}
\caption{\label{fig:celeba-more-30}Discovering 30 landmarks on CelebA}
\end{figure}

\begin{figure}[H]
\centering
  \begin{tabular}{c}
   \includegraphics[width=\textwidth,height=0.85\textheight,keepaspectratio]{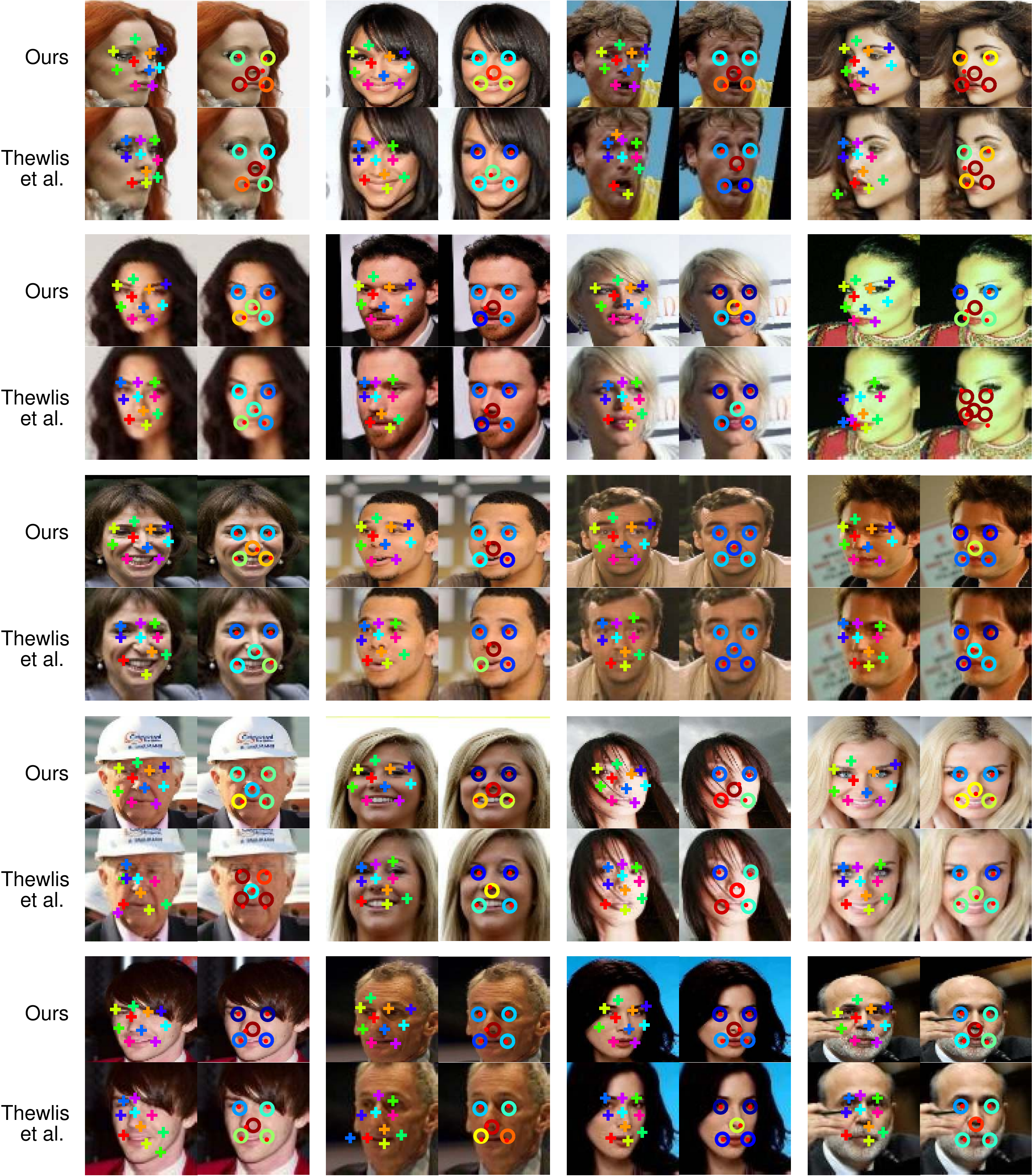} \tabularnewline
    \includegraphics[width=0.8\textwidth,height=1.2cm,keepaspectratio]{figures.reduced/supp_figures/colormap.pdf} 
  \end{tabular}
  \caption{\label{fig:celeba-comp}Prediction of 5 annotated landmarks on CelebA. Colorful cross: discovered landmark; Red dot: annotated landmark; Circle: regressed landmark, whose color represent its distance to the annotated land-marks. See the color bar for the distance (i.e., prediction error). Our method shows more deep blue circles (for example, the image in first row first column, the image in first row fourth column), which means more landmarks with low error compared with \citet{unsupervised-landmark}}
\end{figure}

\begin{figure}[H]
\begin{centering}
\includegraphics[width=\textwidth,height=0.9\textheight,keepaspectratio]{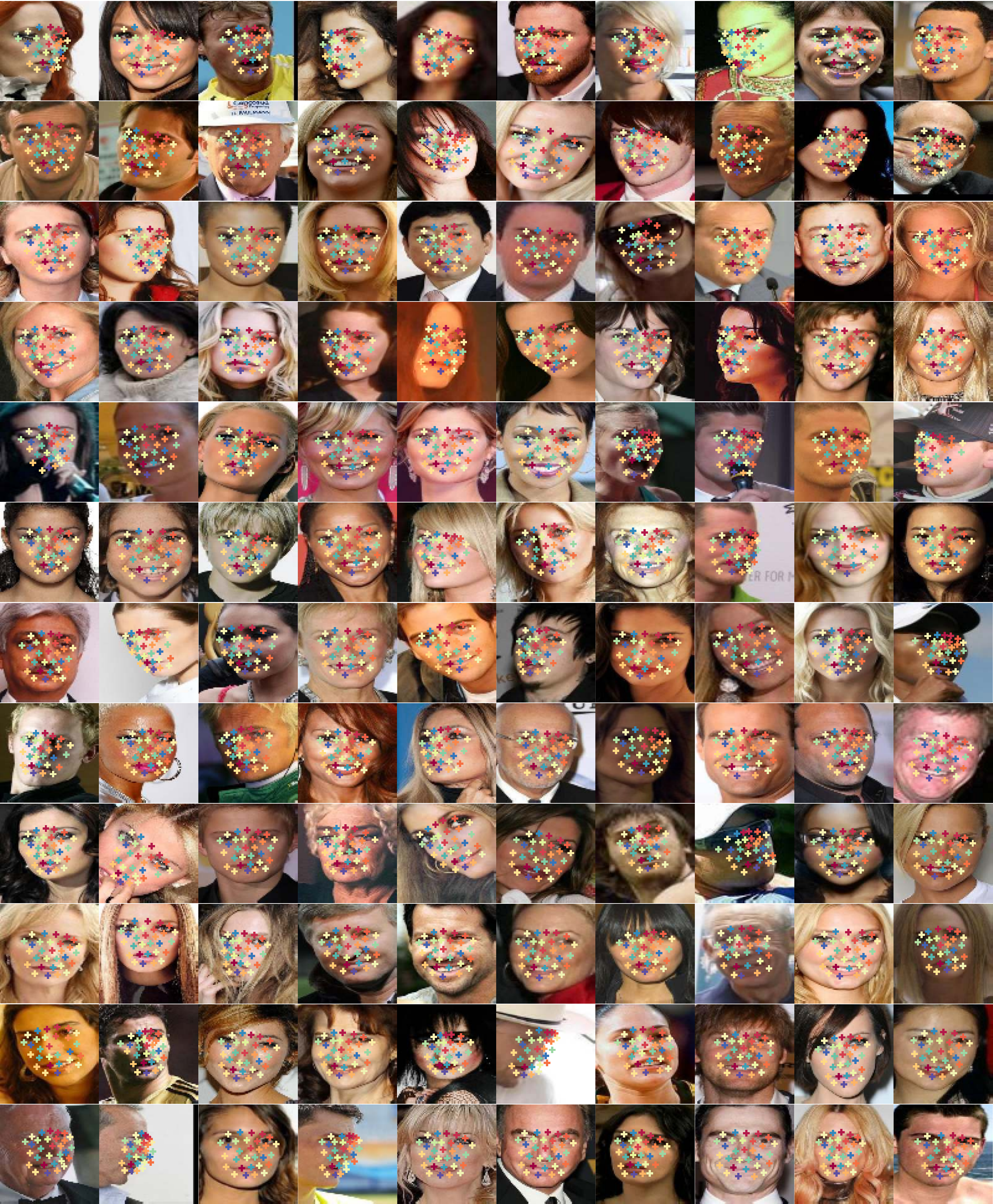}
\par\end{centering}
\caption{\label{fig:celeba-more-30-unaligned}Discovering 30 landmarks on unaligned CelebA images.}
\end{figure}

\subsection{AFLW}
\label{supp:aflw}

\begin{figure}[H]
\begin{centering}
\includegraphics[width=\textwidth,height=0.9\textheight,keepaspectratio]{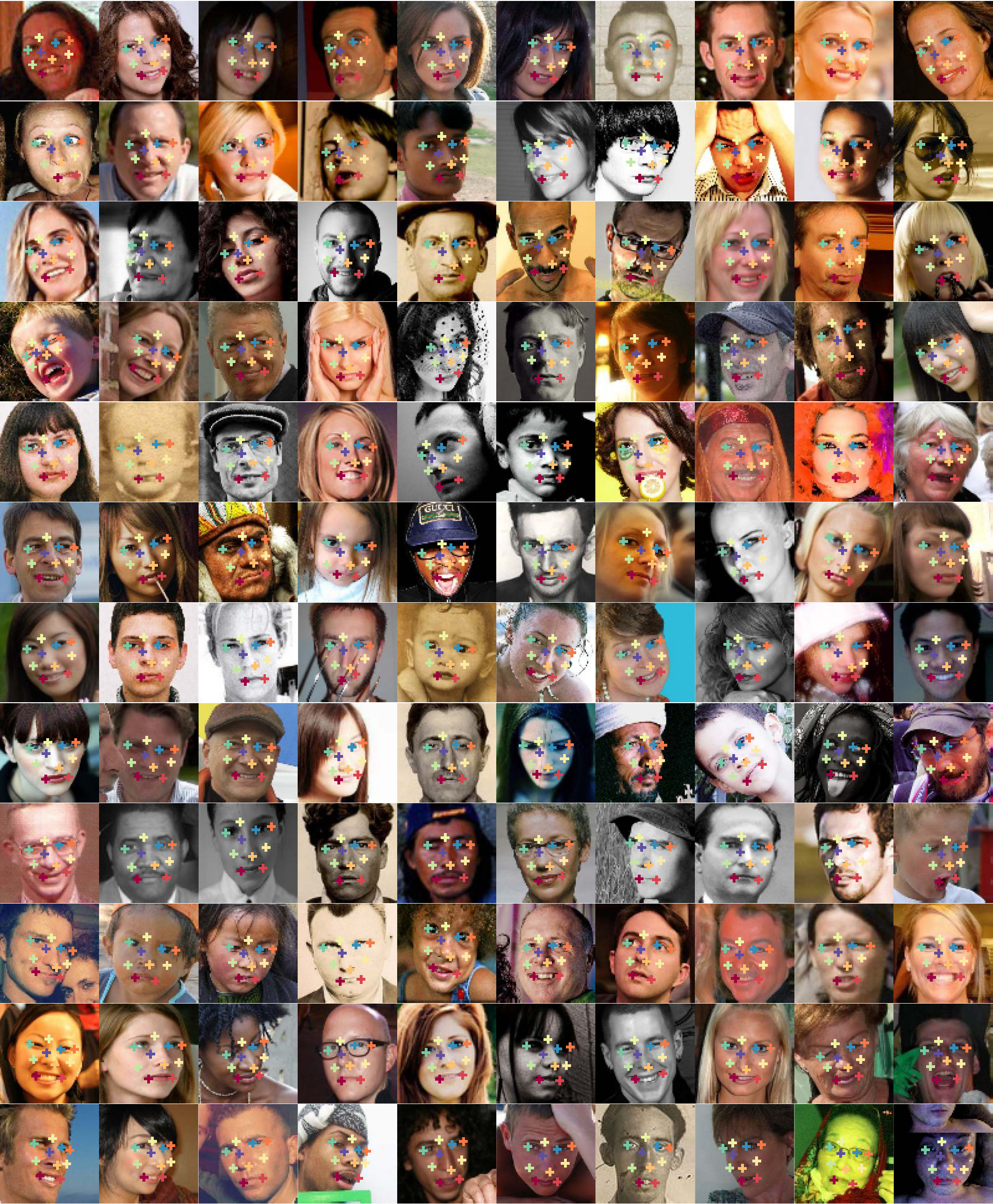}
\par\end{centering}
\caption{\label{fig:aflw-10}Discovering 10 landmarks on AFLW}
\end{figure}

\begin{figure}[H]
\begin{centering}
\includegraphics[width=\textwidth,height=0.9\textheight,keepaspectratio]{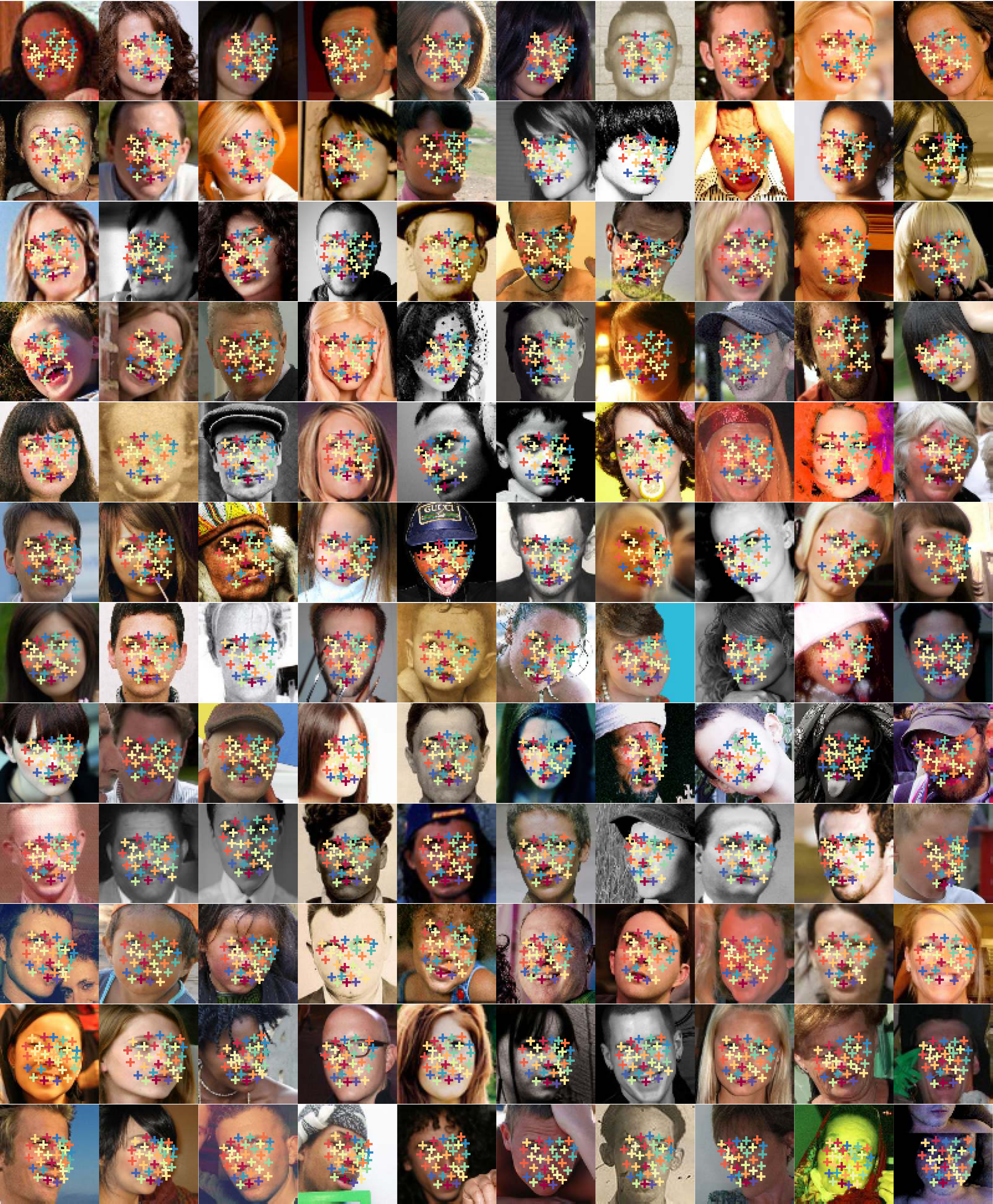}
\par\end{centering}
\caption{\label{fig:aflw-30}Discovering 30 landmarks on AFLW }
\end{figure}

\subsection{Cat heads}
\label{supp:cat}

\begin{figure}[H]
\begin{centering}
\includegraphics[width=\textwidth,height=0.9\textheight,keepaspectratio]{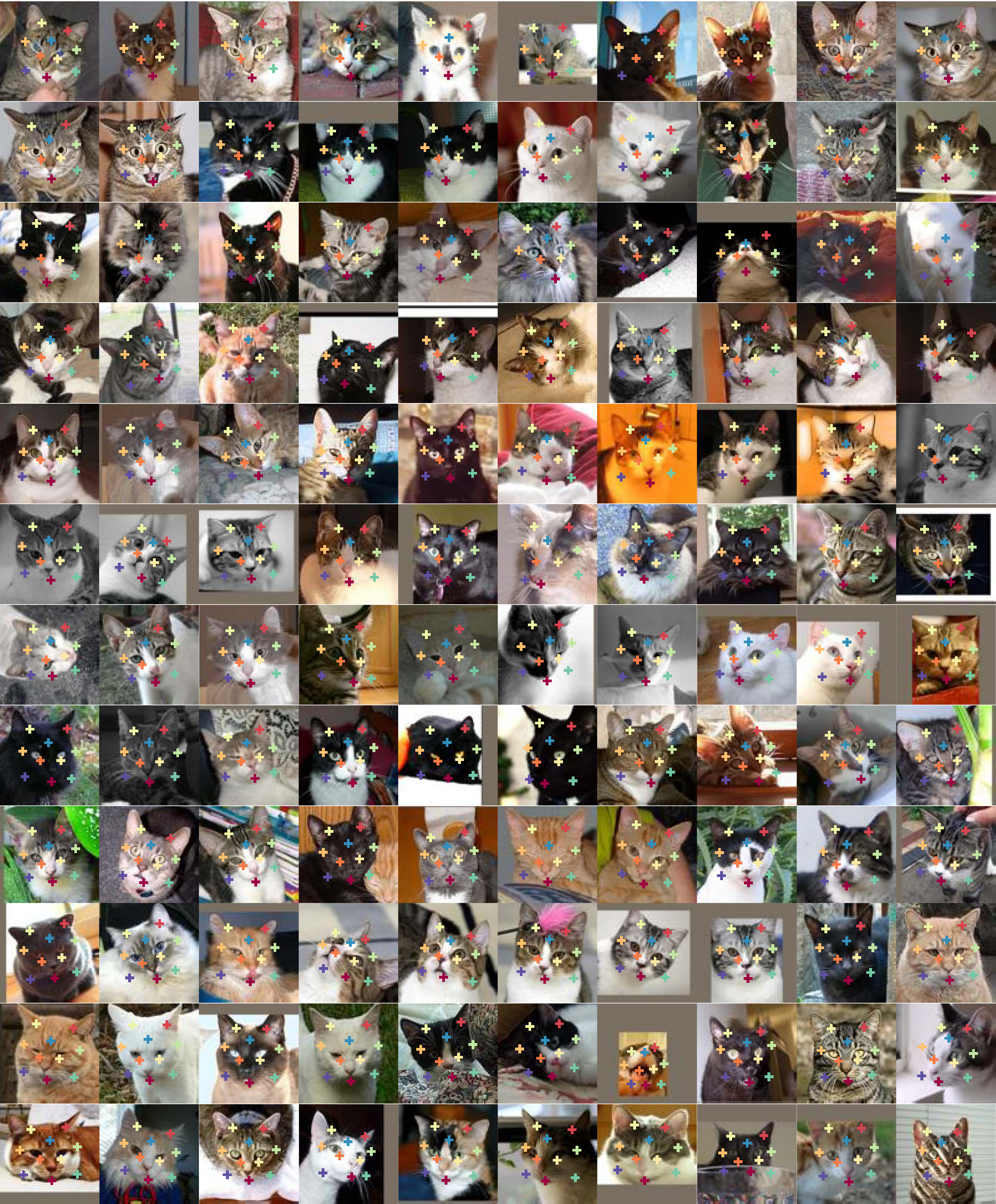}
\par\end{centering}
\caption{\label{fig:cat-more-10}Discovering 10 landmarks on Cat Head dataset, our method find some landmarks on the nose, eyes and base of earlobe}
\end{figure}

\begin{figure}[H]
\begin{centering}
\includegraphics[width=\textwidth,height=0.95\textheight,keepaspectratio]{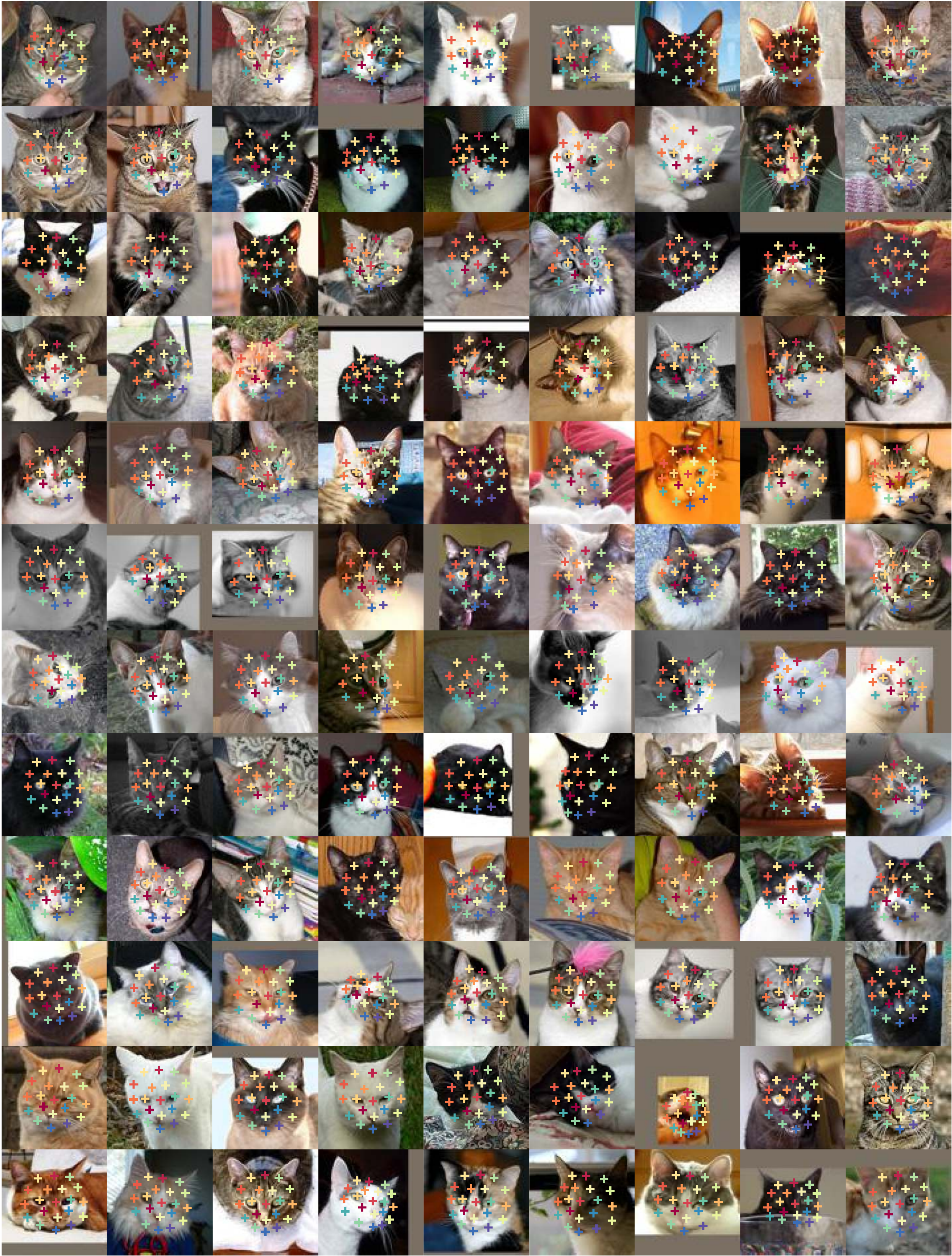}
\par\end{centering}
\caption{\label{fig:cat-more-20}Discovering 20 landmarks on Cat Head dataset}
\end{figure}

\begin{figure}[H]
\centering
  \begin{tabular}{c}
   \includegraphics[width=\textwidth,height=0.85\textheight,keepaspectratio]{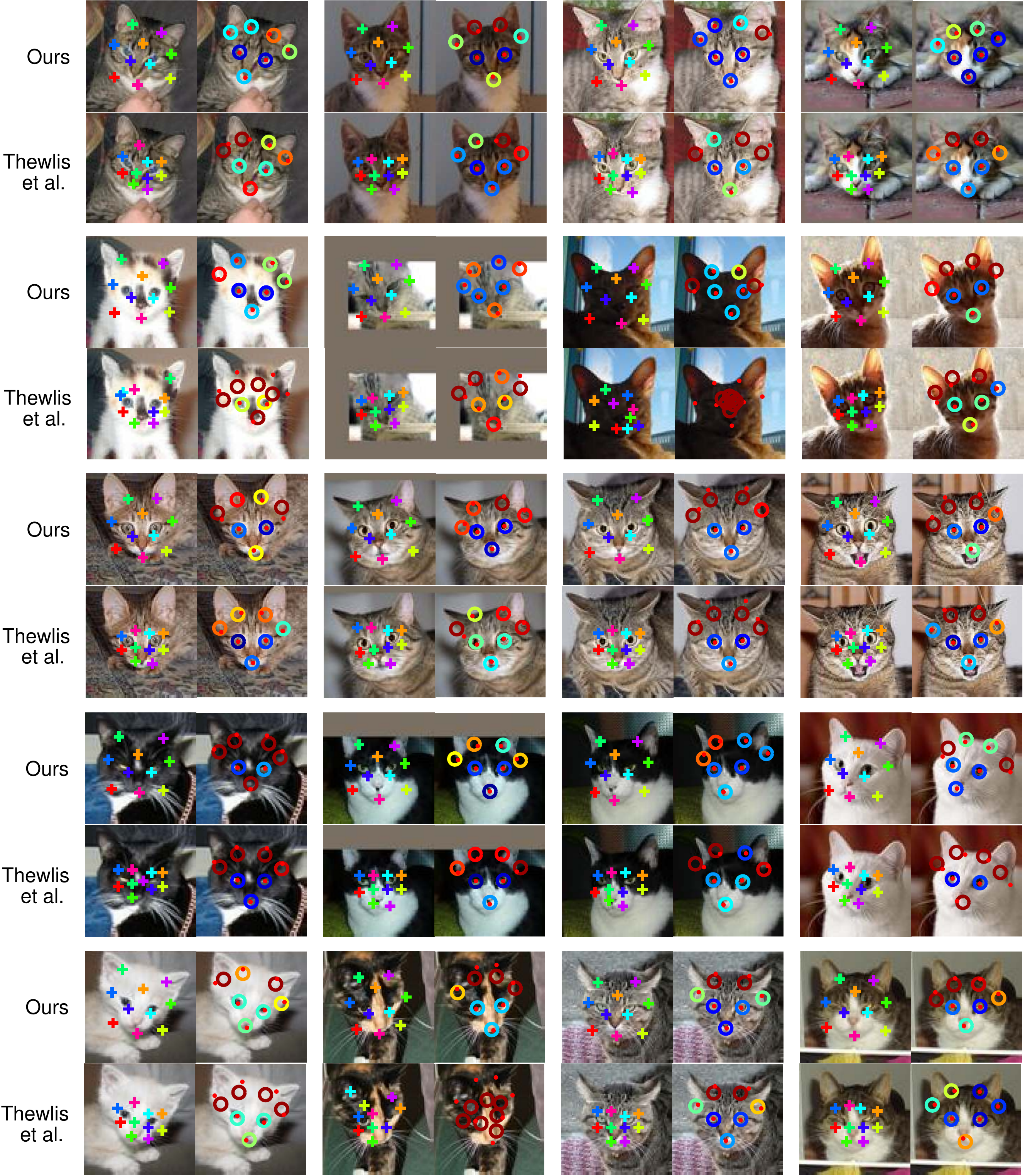} \tabularnewline
    \includegraphics[width=0.8\textwidth,height=1.2cm,keepaspectratio]{figures.reduced/supp_figures/colormap.pdf} 
  \end{tabular}
  \caption{\label{fig:cat-comp}Prediction of 7 annotated landmarks on cat head. Colorful cross: discovered landmark; Red dot: annotated landmark; Circle: regressed landmark, whose color represent its distance to the annotated land-marks. See the color bar for the distance (i.e., prediction error).Our method shows more deep blue circles (for example, the image in fourth row third column, the image in second row first column), which means more landmarks with low error compared with \citet{unsupervised-landmark}}
\end{figure}

\subsection{Cars}
\label{supp:cars}

\begin{figure}[H]
\begin{centering}
\includegraphics[width=\textwidth,height=0.35\textheight]{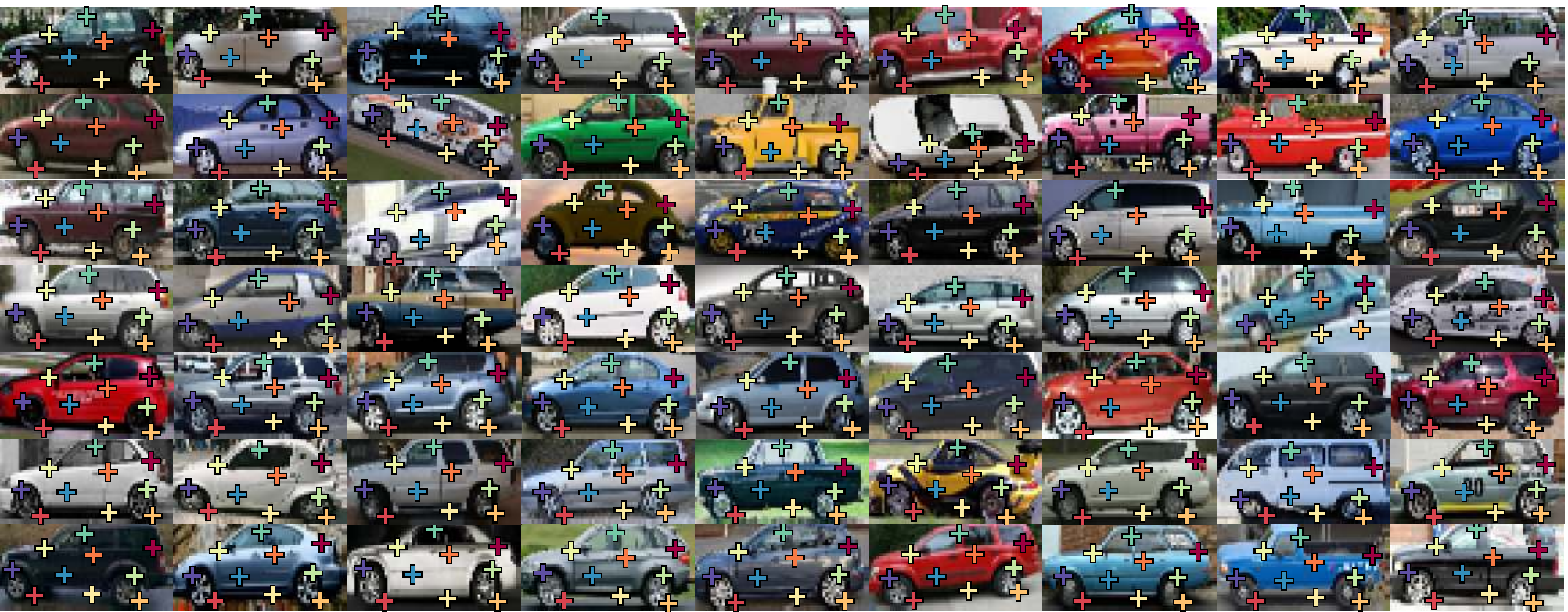}
\par\end{centering}
\caption{\label{fig:car-more-10}Discovering 10 landmarks on PASCAL-VOC 3D Car dataset}
\end{figure}

\begin{figure}[H]
\begin{centering}
\includegraphics[width=\textwidth,height=0.35\textheight]{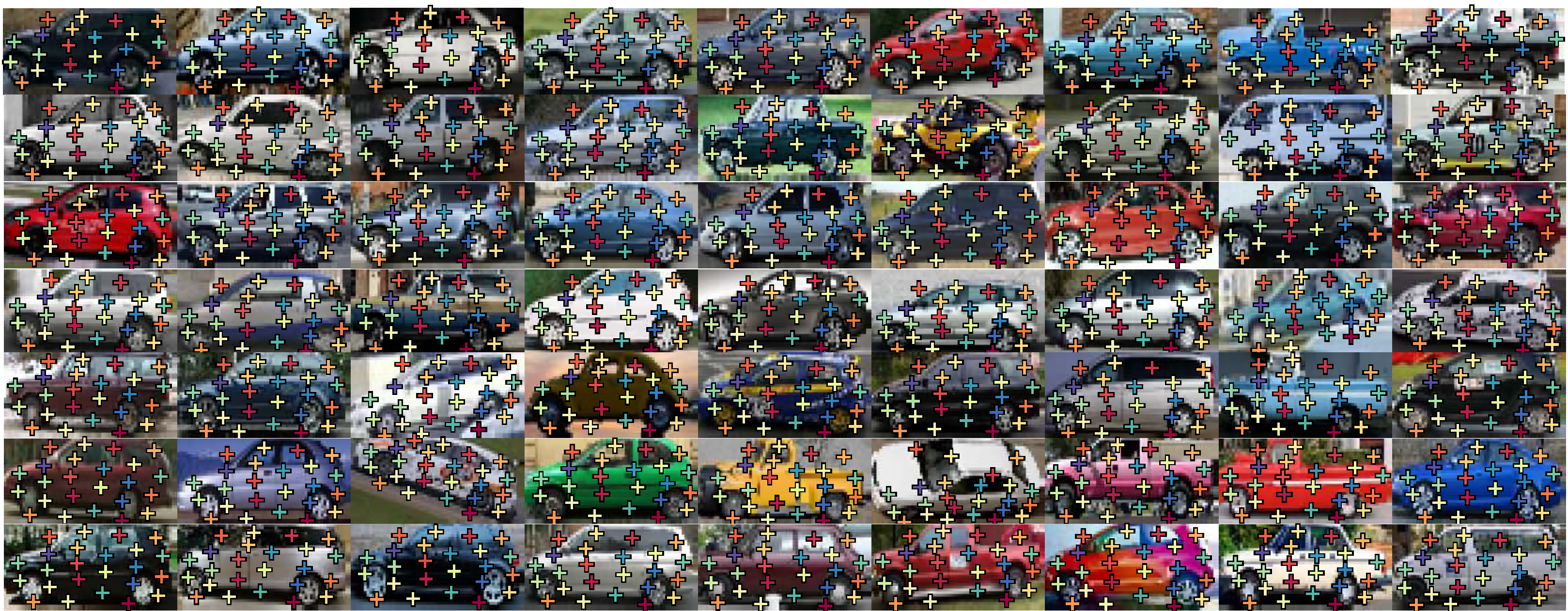}
\par\end{centering}
\caption{\label{fig:car-more-24}Discovering 24 landmarks on PASCAL-VOC 3D Car dataset}
\end{figure}

\begin{figure}[H]
\centering
  \begin{tabular}{c}
   \includegraphics[width=\textwidth]{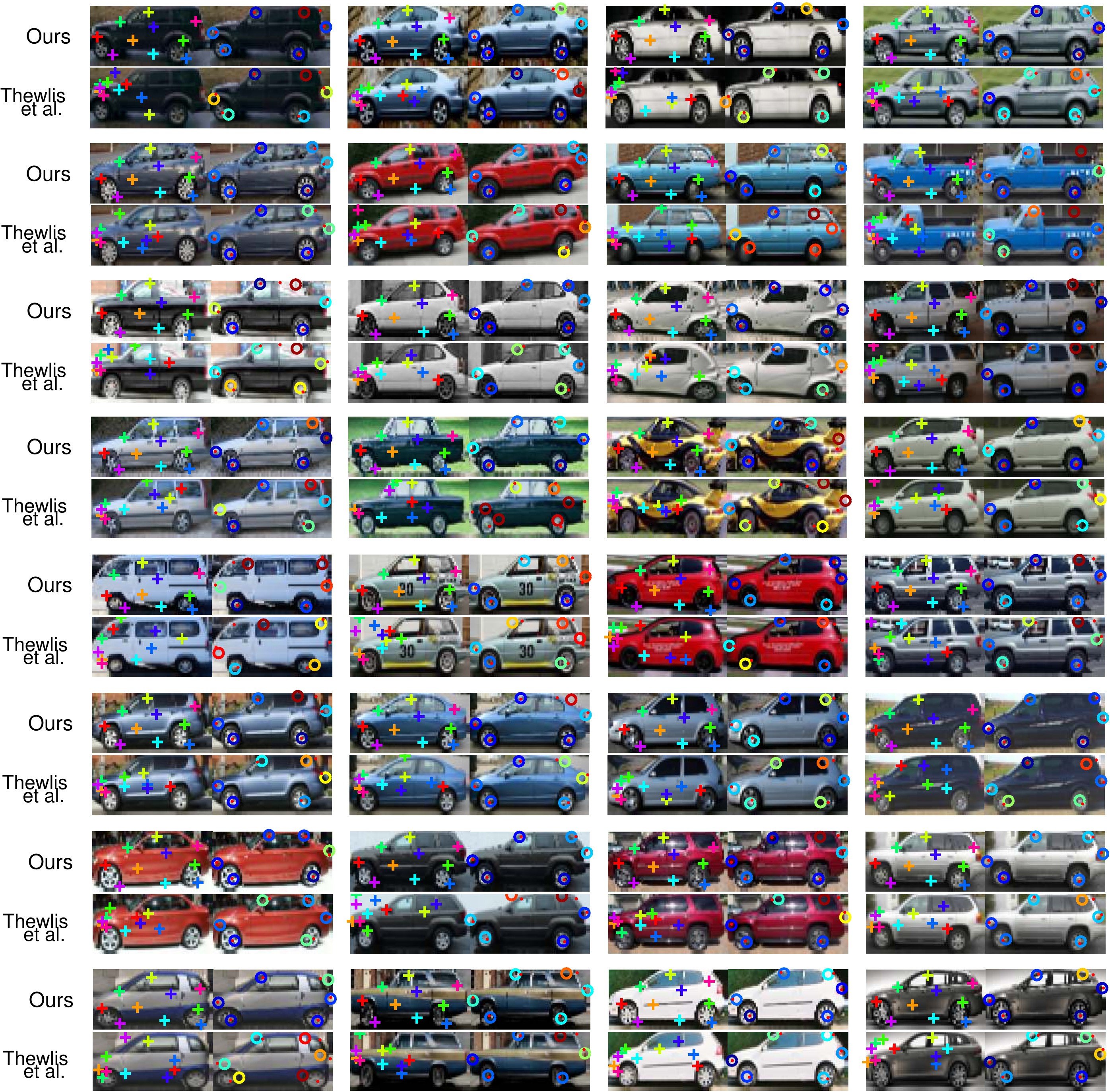} \tabularnewline
    \includegraphics[width=0.8\textwidth,height=1.2cm,keepaspectratio]{figures.reduced/supp_figures/colormap.pdf} 
  \end{tabular}
  \caption{\label{fig:car-comp}Prediction of 6 annotated landmarks on car. Colorful cross: discovered landmark; Red dot: annotated landmark; Circle: regressed landmark, whose color represent its distance to the annotated land-marks. See the color bar for the distance (i.e., prediction error).Our method shows more deep blue circles (for example, the image in last row first column, the image in last row last column), which means more landmarks with low error compared with \citet{unsupervised-landmark}}
\end{figure}

\subsection{Shoes}
\label{supp:shoes}

\begin{figure}[H]
\begin{centering}
\includegraphics[width=\textwidth,height=\textheight,keepaspectratio]{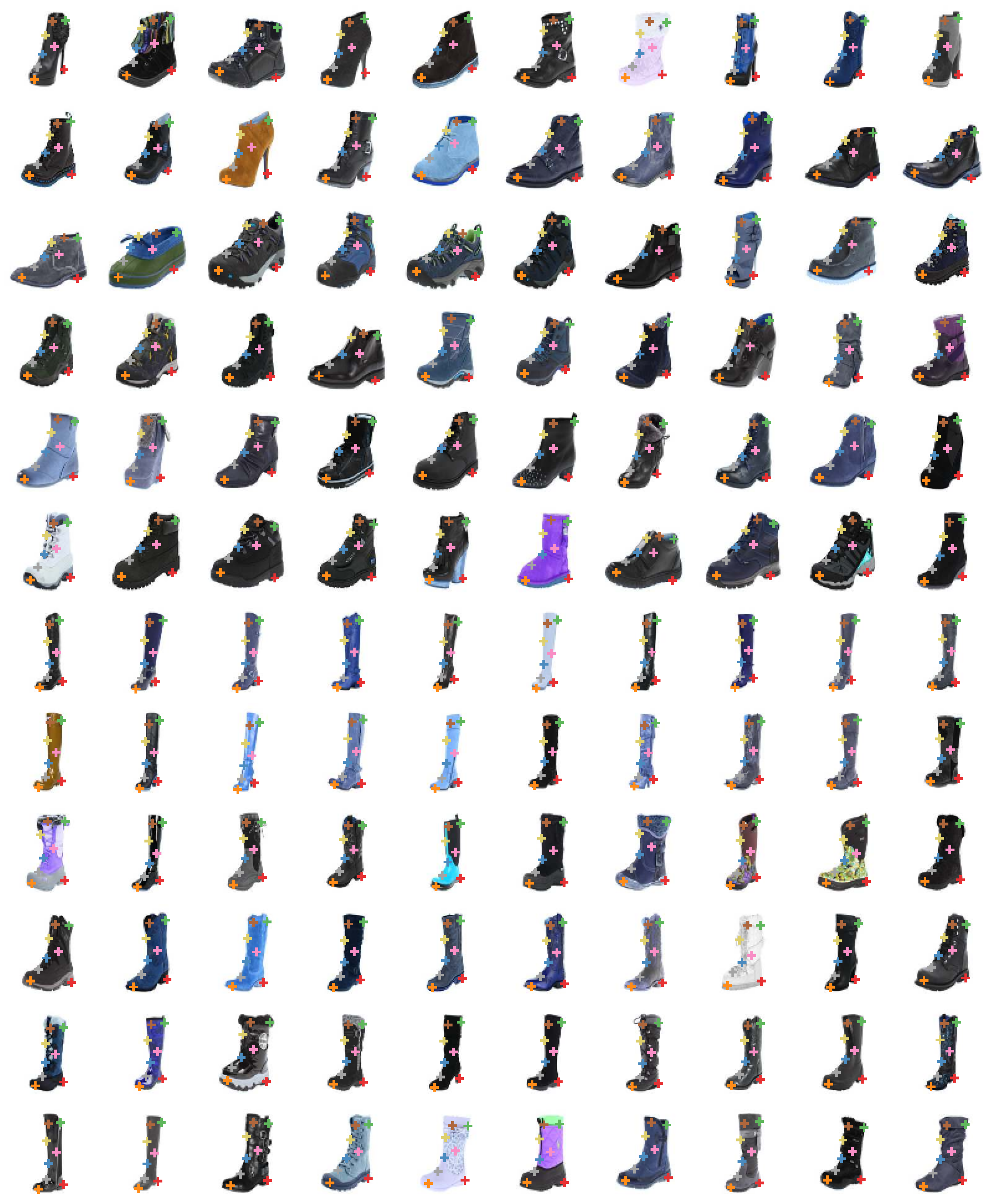}
\par\end{centering}
\caption{\label{fig:gtlm-num-samples}Landmark discovery results of our model on shoes using 8 landmarks }
\end{figure}

\clearpage
\section{Ablative study}
\label{supp:ablative}
\subsection{Number of labeled samples for annotated-landmark prediction}
\label{supp:number-training-samples}

\begin{figure}[H]
\begin{centering}
\includegraphics[width=0.6\columnwidth]{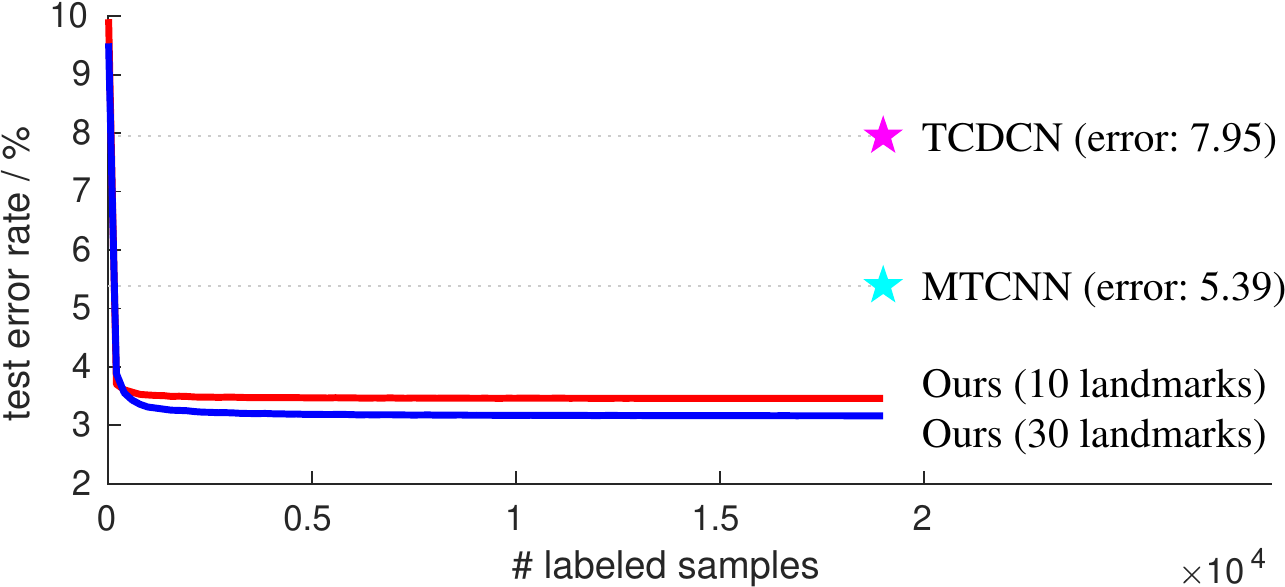}
\par\end{centering}
\caption{\label{fig:gtlm-num-samples}Prediction errors of manually-annotated
landmarks on the MAFL testing set when using different numbers of
labeled samples to train the linear regressor. }

\end{figure}

Taking our model as a detector of manually annotated landmarks, we find that less than 200 samples are enough for our model to achieve
less than 4\% mean error on the MAFL testing set, which is better than the performance of two popular off-the-shelf models.
This result suggests that it is possible to train a high-accurate landmark detector using only a few labeled sample when sufficient unlabeled samples are given to train our unsupervised model. 
We show its performance versus the number of labeled samples in Figure~\ref{fig:gtlm-num-samples}.

\subsection{Evolution of detection confidence map during training}
\label{supp:heatmap-evolution}

\begin{figure}[H]
\includegraphics[width=1\columnwidth]{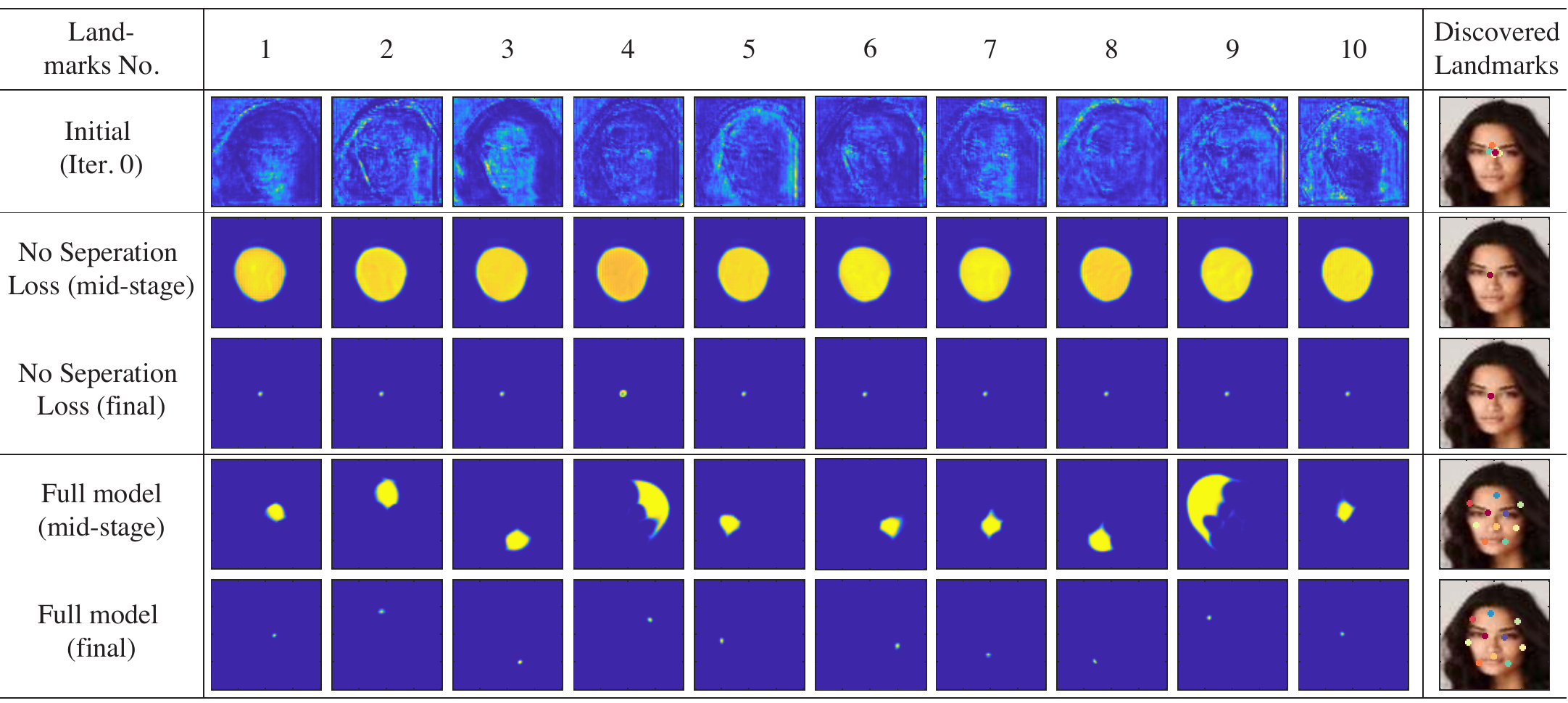}
\caption{The evolution of the detection confidence map $\mathbf{D}$ during training.  
The training of a 10-landmark CelebA model is monitored. 
}
\label{fig:sep-heatmap-evolution}
\end{figure}

Figure~\ref{fig:sep-heatmap-evolution} shows the detection confidence maps of an input image at different training stage of our model.
In the beginning, the heatmap shows random values over the whole image. 
As a result, the landmarks, defined as the mean coordinates weighted by the confidence maps, are all at the center of the image.
As the training goes on, the values gradually becomes spatially concentrated.
With the separation loss defined in \eqref{eq:sep-loss}, the peaked value of each channel of the confidence map can move to a different location. 
Without the separation loss, every channel can have a peaked value at the center of the images, resulting in degenerate landmarks. 

\subsection{TPS control points}
\label{supp:tps-control}

For the random TPS in our equivariance constraint (defined in \eqref{eq:eqv-loss}), we both use the regular-grid control points and take the discovered landmarks in the current iteration as the control points. 
The two sets of control points are alternatively used in each optimization iteration with 7$:$3 chance. 
We do not exhaustively tune the ratio and keep it the same in all experiments. 
As shown in Figure~\ref{tab:tps-control}, the performance of our model is fairly insensitive to this ratio when the other hyper-parameters are fixed.
However, introducing the discovered landmarks as the TPS control points does benefit. 

\begin{table}[H]
\begin{center}
\begin{tabular}{c|cccccc}
\hline 
grid$:$landmark & 10:0 & 1:9 & 3:7 & 5:5 & 7:3 & 9:1\tabularnewline
\hline 
GT prediction error / \% & 4.17 & 3.86 & 3.46 & 3.38 & 3.46 & 3.41\tabularnewline
\hline 
\end{tabular}
\end{center}
\par
\caption{\label{tab:tps-control}
Impact of the ratio between two types of TPS control points (i.e., regular grid and discovered landmarks). 
Our 10-landmark CelebA model is trained under different ratios of the TPS control points. 
The evaluation metric is the ground truth (GT) landmark prediction errors with the CelebA training set for linear regressor training.
}
\end{table}

Note that the discovered landmarks are clustered at the center of the image (see discussion about the separation constraint in Section~\ref{sec:landmark-concept}) and cannot serve as good TPS control points. 
As a result, we use only the regular-grid control points in the beginning and start to apply the previously mentioned ratio after training the model for several thousands of iterations. 
 
\section{Implementation details}
\label{supp:implementation}
\subsection{Data preprocessing}
\label{supp:data-preproc}

The main paper and this supplementary materials report results on several datasets. 
Table~\ref{tab:data-param} summarizes the image size we used for each each dataset. 
In our landmark discovery formulation, we need to perform random TPS to calculate the equivariance constraint in (\ref{eq:eqv-loss}). 
It requires the image to have large enough margins so that the foreground will not be out of image due to the random transformation. 
Table~\ref{tab:data-param} also summarizes the image size after padding with the edge values. 

\begin{table}[H]
\begin{centering}
\begin{tabular}{c|ccc}
\hline 
Dataset & Image size & Padded size & Remark\tabularnewline
\hline 
\hline 
CelebA, AFLW, Cat Heads, Shoes  & 80$\times$80 & 96$\times$96 & -\tabularnewline
\hline 
Profile Cars from PASCAL 3D & 64$\times$64 & 96$\times$96 & visualized as $W:H=2:1$\tabularnewline
\hline 
Animals from AwA & 64$\times$64 & 80$\times$80 & -\tabularnewline
\hline 
Human3.6M & 128$\times$128 & 192$\times$192 & -\tabularnewline
\hline 
MNIST & 28$\times$28 & 56$\times$56 & -\tabularnewline
\hline 
\end{tabular}
\par\end{centering}
\caption{\label{tab:data-param}Data processing parameters for different dataset.}
\end{table}

For different datasets, we crop the foreground images and prepare input images as follows.

\begin{description}
\item[CelebA] We started from the cropped-and-aligned images (218$\times$178) in CelebA dataset, scaled them to 100$\times$100 pixels and then cropped the 80$\times$80 center patches.

\item[AFLW] The dataset provides annotated bounding boxes. We enlarged the bounding boxes on each image with a margin on the top ($1/4$ height of the original bounding box), margins on the left and right ($1/8$ width of the original bounding box) so that the cropped facial images look similar to the CelebA data. The cropped images were also scaled to 80$\times$80. 

\item[Cat Heads] The dataset provides ground truth landmarks on the cat head. 
We figured out the bounding box for cropping each image according to the landmarks and then scaled the cropped images to 80$\times$80.

\item[Shoes] We scaled the shoe images (102$\times$136) to 64$\times$64, and padded the images to be 80$\times$80 with white margins. We linearly transformed the color value range from $[0,1]$ to $[0.1, 0.9]$ to avoid a huge amount of saturated responses for the output layer with the sigmoid activation. The color was scaled back for visualization.

\item[Cars] The PASCAL 3D dataset provides annotations for the orientation and landmarks of several objects. We cropped the profile car images according to the bounding box of the ground truth landmarks. Slight margins are added to the bounding box, and the cropped image is scaled to a square image without preserving the aspect ratio.  

\item[Animals from AwA] 
We manually annotated bounding boxes and orientation labels (i.e., left, right, frontal, back) for several types of animals. 
For each rectangular bounding box, we enlarged its shorter edge to make the box square, cropped the patch, and scaled it to 64$\times$64.
We ignored the images with frontal and back views, and we flipped the right-facing animals horizontally so that all animals in the image face to the left. 

\item[Human3.6M] 
The human body was cropped using a square bounding box from the original video frames. 
We use the 3D landmarks provided in this dataset, acquired by wearable markers, as side information to roughly align the scales and foot locations of the human bodies in different images. The cropped square images are scaled to 128$\times$128 pixels.
We use the provided segmentation masks, obtained by an off-the-shelf unsupervised background removal method, to mask out the background image with gray color. For visualization, we show the gray color as white for the printing clarity.

\end{description}

\clearpage

\begin{landscape}

\begin{figure}[p]

\begin{centering}

\includegraphics[width=0.92\columnwidth]{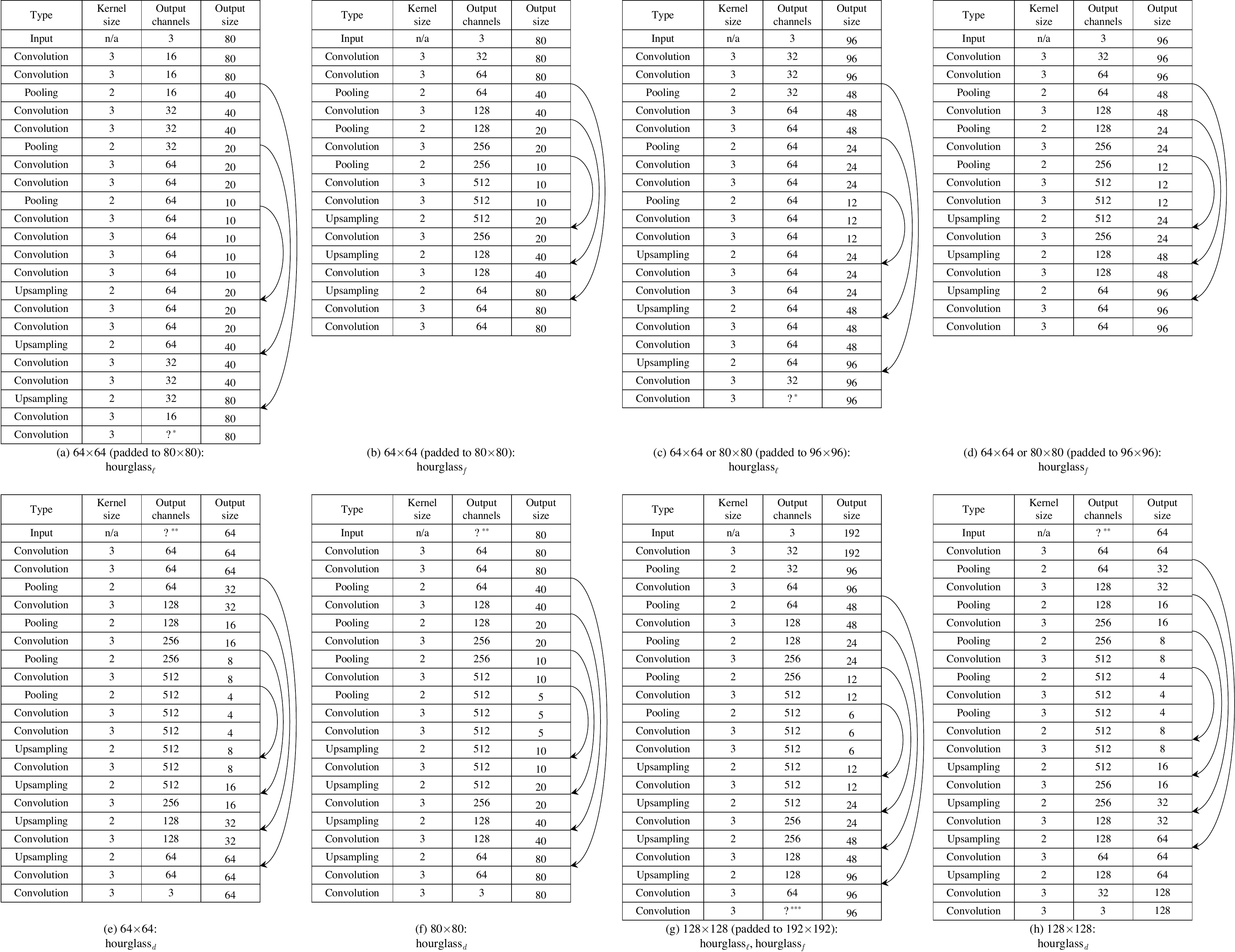}

\end{centering}

\begingroup

\small{}

?\textsuperscript{~*}: decided by the number of landmarks.
?\textsuperscript{~**}: decided by the number of landmarks and the dimension of the shared feature space.
?\textsuperscript{~***}: this layer is used only for $\op{hourglass}_{\ell}$, and the number of channels is decided by the number of landmarks.
?\textsuperscript{~****}: decided by the number of landmarks for $\op{hourglass}_{\ell}$, $=32$ for $\op{hourglass}_{f}$, and $=3$ for $\op{hourglass}_{d}$. 
Each skip-link in \textbf{(a)}, \textbf{(b)}, \textbf{(c)}, \textbf{(d)}, and \textbf{(f)} consists of three convolutional layers of 3$\times$3 kernels. 
Each skip-link in \textbf{(e)}, \textbf{(g)}, and \textbf{(h)} consists of two convolutional layers of 3$\times$3 kernels.

\endgroup

\caption{\label{fig:hourglass}
Architectures of our hourglass-style networks. 
\textbf{(a)},\textbf{(b)}, \textbf{(c)}, \textbf{(d)}:~For 80$\times$80 and 64$\times$64 images, the outputs of  $\op{hourglass}_{\ell}$ and $\op{hourglass}_{f}$ have the same size as the image. \textbf{(g)}: For 128$\times$128 images, the outputs of  $\op{hourglass}_{\ell}$ and $\op{hourglass}_{f}$ are 64$\times$64. 
}

\end{figure}

\end{landscape}

\subsection{Network architectures}
\label{supp:arch-details}

In Section~\ref{sec:method} of the main paper, we describe the key architectures of our model and leave some details unspecified. 
This section describes the detailed neural network architectures.  

Figure~\ref{fig:hourglass} summarizes the hourglass-like architectures that we used for images of different sizes. 
The image padding is explaned and specified in \supp{\ref{supp:data-preproc}}.
In general, an hourglass architecture has mirrored encoding (high-resolution to low-resolution) and decoding (low-resolution and high-resolution) architectures. 
Skip-links, made up of convolutional layers, create shortcuts from the encoding feature maps and decoding feature maps of the same resolution. 
The responses of the skip-links are fused with the main stream decoding responses using element-wise addition. 
We use the max-pooling to reduce the feature map size for encoding, and we upsample feature maps by the nearest interpolation for decoding. 
For each linear layer, a batch normalization layer is followed, and LeakyReLU~\citep{lrelu} is the default activation function. 

Based on the neural network architectures in Figure~\ref{fig:hourglass}, a convolutional layer of 1$\times$1 kernels calculates the raw detection score map of $K+1$ channels (recall that $K$ is the number of landmarks). 
For image output,  the last convolutional layer's responses ($\in\mathbb{R}$) are mapped to $(0,1)$ with a sigmoid function for the reconstructed color intensity and to $(0,+\infty)$ by  $\log(1 + 2\exp(z))/2$ for the pixel-wise standard deviation, respectively.

\subsection{Training strategy}

We use Adam with an initial learning rate of $0.001$  to optimize the neural network parameters. 
We set the training batch size to 16 or 32\footnote{There is no significant difference in the performance when using either 16 or 32 as the training batch size.} for 80$\times$80 and 60$\times$60 images and 8 for 128$\times$128 images. 
The learning rate starts from ${10}^{-3}$ and decreases to ${10}^{-4}$ and ${10}^{-5}$ later.
For color images, we do random brightness and contrast jittering. 

For batch normalization, the global mean and variance are computed using a random subset of the training set when the neural network training is done. 
Note that using the running average during training for the global mean and variance can hurt the performance. 

To implement the equivariance constraint in (\ref{eq:eqv-loss}), we use random TPS transformation to obtain warped input images (paired with the original images) in each training iteration. 
Taking the normalized image height and width as $1$, the random transformation parameters are:
\begin{itemize}
\item  \textbf{Global affine component.} 
Uniform random translation in $\pm0.15$;
Gaussian random rotation with the standard deviation of $10^{\circ}$;
Gaussian random scaling in the base-$2$ logarithm scale with the standard deviation of $1.25$. 
\item \textbf{Local TPS.} 
Gaussian random translation with the standard deviation of $0.1$ (regarding the regular-grid control points) or $0.05$ (regarding the landmark control points).
\end{itemize}

For the reconstruction loss weight $\lambda_{\mathrm{recon}}$ in \eqref{eq:ae-loss}, we find it helpful to start with a small value and increase it to its final value at a later stage.
At the early training stage, the discovered landmarks change significantly for each iteration, and the latent descriptor of landmarks inputted to the decoder also varies a lot. 
The model parameters of the decoder and the gradients from it can change too drastically if the reconstruction loss weights is large, which would harm the training of both the landmark detector and the landmark-based image decoder. 
As a particular training strategy, we increase the value of $\lambda_{\mathrm{recon}}$ by $\times 10$ twice during the training. 
Table~\ref{tab:hyperparam} in \supp{\ref{supp:hyperparam}} summarizes the detailed hyper-parameters.

For the L2 reconstruction loss $L_{\mathrm{recon}}$, we scale the image pixel values to $[0,1]$.
The loss is explicitly defined as the negative logarithm likelihood (NLL) to draw the input image $\mathbf{I}$ from the Gaussian distribution centered at the reconstructed image $\tilde{\mathbf{I}}$ the with a fix variation $\sigma_{\mathrm{color}}=0.05$. 
More specifically, 
\begin{equation}
L_{\mathrm{recon}}=\frac{1}{\sigma_{\mathrm{color}}^{2}}{\Vert\mathbf{I}-\tilde{\mathbf{I}}\Vert}_{F}^{2}+\ln{(2\pi\sigma_{\mathrm{color}}^{2})}.
\end{equation}
The value of the loss weight $\lambda_{\mathrm{recon}} $ is based on this definition.

\clearpage

\subsection{Hyper-parameters}
\label{supp:hyperparam}

Table~\ref{tab:hyperparam} summarizes the dataset-specific hyper-parameters.
Note that our model is not sensitive to minor changes of the hyper-parameters, but adjusting the hyper-parameters for each dataset can improve the performance slightly. 

\begin{table}[H]

\begingroup
\setlength\tabcolsep{4pt}
\begin{center}
\begin{tabular}{c|>{\centering}m{30pt}|>{\centering}m{45pt}>{\centering}m{45pt}>{\raggedleft}m{30pt}>{\centering}m{45pt}>{\centering}m{45pt}ccccc}
\hline 
Dataset & \# landmark & 1\textsuperscript{st} LR decay iter. & 2\textsuperscript{nd} LR decay iter. & Initial $\lambda_{\textrm{recon}}$ & 1\textsuperscript{st} $\lambda_{\textrm{recon}}$ increase iter. & 2\textsuperscript{nd}$\lambda_{\textrm{recon}}$ increase iter. & \multirow{1}{*}{$\lambda_{\textrm{conc}}$} & $\sigma_{\textrm{sep}}$ & \multirow{1}{*}{$\lambda_{\textrm{sep}}$} & \multirow{1}{*}{$\lambda_{\textrm{eqv}}$} & \multirow{1}{*}{$C$}\tabularnewline
\hline 
CelebA & 10 & 100K & 200K & 0.01 & 100K & 200K & 100 & 0.06 & 16 & $\text{10}^{\text{4}}$ & 8\tabularnewline
CelebA & 30 & 100K & 200K & 0.1 & 100K & 200K & 100 & 0.04 & 10 & $\text{10}^{\text{4}}$ & 8\tabularnewline
\hline 
AFLW & 10 & 100K & 200K & 0.1 & 100K & 200K & 100 & 0.06 & 16 & $\text{10}^{\text{4}}$ & 8\tabularnewline
AFLW & 30 & 100K & 200K & 0.0001 & 100K & 200K & 100 & 0.04 & 10 & $\text{10}^{\text{4}}$ & 8\tabularnewline
\hline 
Cat & 10 & 100K & 200K & 0.0001 & 100K & 200K & 100 & 0.08 & 20 & $\text{10}^{\text{4}}$ & 8\tabularnewline
Cat & 20 & 100K & 200K & 0.0001 & 100K & 200K & 100 & 0.05 & 10 & $\text{10}^{\text{4}}$ & 8\tabularnewline
\hline 
Car & 10 & 40K & 80K & 0.001 & 40K & 50K & 100 & 0.08 & 200 & $\text{10}^{\text{4}}$ & 8\tabularnewline
Car & 24 & 40K & 80K & 0.001 & 40K & 50K & 100 & 0.05 & 200 & $\text{10}^{\text{4}}$ & 8\tabularnewline
\hline 
Animal & 10 & 20K & 50K & 0.001 & 40K & 50K & 100 & 0.08 & 20 & $\text{10}^{\text{4}}$ & 2\tabularnewline
Shoes & 8 & 100K & 20K & 0.01 & 100K & 200K & 100 & 0.05 & 20 & $\text{10}^{\text{4}}$ & 8\tabularnewline
\hline 
Human & 16 & 100K & 200K & 0.1 & 100K & 200K & 100 & 0.06 & 20 & $\text{10}^{\text{4}}$ & 8\tabularnewline
\hline 
\end{tabular}
\par\end{center}

\endgroup

\caption{\label{tab:hyperparam}
The hyper-parameters for our models on different datasets. 
\textbf{
When computing loss $L_{\mathrm{conc}}, L_{\mathrm{sep}}, L_{\mathrm{eqv}}$, the coordinates is first \emph{normalized with respect to the image edge length} (i.e., the square root of the image area). 
All hyper-parameters (including, $\lambda_{\textrm{conc}}$, $\sigma_{\textrm{seq}}$, $\lambda_{\textrm{seq}}$, $\lambda_{\textrm{eqv}}$) are set according the normalized landmark coordinates.   
}
}
\end{table}

\subsection{Details about face generations using unsupervised landmarks}
\label{supp:face-generation-details}

Figure~\ref{fig:face-generation} in the main paper shows results of generating facial images conditioned on our discovered landmarks. 
In this experiment, we fix our landmark discovery module and use it to detect landmarks on training images. 
For the decoding module, we take the detected landmarks as a given input condition and map an isotropic Gaussian random variable to the latent part of the image representation. 
Inspired by \citep{wwgan}, we first use deconvolutional layers to get a feature map from the random variable and use convolutional layers to get another map of the same size from the reconstructed detection confidence map. 
We use element-wise multiplication and channel-wise concatenation to fuse the two into one feature map as the input of the decoding neural network. 
Thus, the way of calculating the input feature map of the decoding module is not the same as our landmark discovery model. 

We use the boundary equilibrium GAN (BEGAN)~\citep{began} framework to design the discriminator, which encourages the decoder to generate realistic images. 
More concretely, an autoencoder is trained as the energy function to distinguish the real and generated images. 
To make sure the generated images are consistent with the landmark condition, we first use our landmark discovery module to detect landmarks on them. 
We then take the L1 distance between them and those from the corresponding real images (i.e., input training images for getting the landmarks) as an extra training loss for the decoding module. 

This experiment is mainly to show that our discovered landmark is accurate and meaningful enough for controllable image generation.

\end{document}